\documentclass[runningheads]{llncs}
\usepackage[T1]{fontenc}
\usepackage[utf8]{inputenc}
\usepackage{graphicx}
\usepackage{booktabs}
\usepackage{multirow}
\usepackage{amsmath}
\usepackage{amssymb}
\usepackage{xcolor}
\usepackage[hyphens]{url}
\usepackage{hyperref}
\usepackage{microtype}         
\newcommand{\citep}[1]{\cite{#1}}
\newcommand{\citet}[1]{\cite{#1}}
\graphicspath{{figures/}}
\begin{document}
\title{Hallucination Is Linearly Decodable from Mid-Layer Hidden States in Quantized LLMs}
\titlerunning{Linearly Decodable Hallucination in Quantized LLMs}
\author{Aizierjiang Aiersilan}
\authorrunning{Aizierjiang Aiersilan}
\institute{University of Macau\\
\email{mc25101@um.edu.mo}}
\maketitle
\begin{abstract}
We investigate whether open-source LLMs encode a linearly separable truthfulness signal in their hidden states, and at which network depth this signal is strongest. Across three $7$B--$8$B instruction-tuned models (Llama-3.1-8B, Mistral-7B, Qwen2.5-7B) loaded in $4$-bit NF4 quantization, we extract per-layer hidden states on four hallucination benchmarks (TruthfulQA, HaluEval-QA, FEVER, and a controlled synthetic set) and compare four detection approaches: linear and MLP probes, INSIDE EigenScore, self-consistency, and attention entropy. A linear probe on a single mid-network layer achieves $0.904$--$1.000$ AUROC on held-out splits, while sampling-based detectors do not exceed $0.541$ AUROC under the same protocol. The truthfulness signal is approximately linear: MLP probes rarely surpass linear probes by more than $0.01$ AUROC. Peak probing layers fall in a consistent band across model families on natural-language benchmarks---blocks~$13$--$18$ of~$32$ for Llama and Mistral, and blocks~$19$--$25$ of~$28$ for Qwen. First-block attention entropy provides a complementary signal in knowledge-grounded settings ($0.866$--$0.941$ AUROC on HaluEval-QA) at no additional inference cost. The low discriminability of sampling methods under this protocol reflects a structural mismatch between paired-label evaluation and the information these methods access, rather than an inherent limitation of those methods. Code and data are released for full reproducibility on a single $8$\,GB GPU.

\end{abstract}
%
%
\section{Introduction}
\label{sec:intro}

Hallucinations, i.e., fluent but factually unsupported generations, remain a prevalent failure mode of large language models in real-world deployments \citep{ji2023survey,huang2025survey}. Detecting them at inference time is a key practical concern. A growing literature offers two complementary families of detectors. \emph{Sampling-based} methods draw several stochastic generations for the same prompt and flag disagreement at the lexical \citep{wang2022self,manakul2023selfcheckgpt}, semantic \citep{kuhn2023semantic,farquhar2024detecting}, or internal-state level \citep{chen2024inside}. \emph{Representation-based} methods train a small classifier (probe) to decode truthfulness directly from intermediate hidden states \citep{azaria2023internal,burns2022discovering,li2023inference,alain2016understanding}. These two families differ sharply in compute: sampling requires $K\!\geq\!5$ extra generations per prompt, while a probe requires only a single forward pass. Yet they are rarely compared head-to-head on identical models and datasets, especially in the small, open, quantized regime that practitioners deploy on consumer hardware.

We address this gap with a unified evaluation framework that runs end-to-end on a single $8$\,GB GPU. We study three $7$B--$8$B instruction-tuned models in $4$-bit NF4 quantization \citep{dettmers2023qlora} (Llama-3.1-8B-Instruct \citep{grattafiori2024llama}, Mistral-7B-Instruct-v0.3 \citep{jiang2023mistral}, and Qwen2.5-7B-Instruct \citep{qwen2024}) across four hallucination datasets: TruthfulQA \citep{lin2022truthfulqa}, HaluEval-QA \citep{li2023halueval}, FEVER \citep{thorne2018fever}, and a controlled synthetic benchmark. For every (model, dataset) combination we evaluate four detection methods: (i)~SAPLMA-style linear and MLP probes \citep{azaria2023internal} on per-layer hidden states; (ii)~INSIDE EigenScore \citep{chen2024inside} over multiple sampled completions; (iii)~self-consistency \citep{wang2022self,manakul2023selfcheckgpt} scored by exact-match plurality and sentence-embedding similarity \citep{reimers2019sentence}; and (iv)~attention-entropy summaries at three transformer blocks. Our main findings and contributions are as follows.

\paragraph{Contributions.}
(1)~A head-to-head, layer-resolved comparison of four hallucination-detection methods on three open $7$B--$8$B chat models across four datasets, fully reproducible on consumer hardware via $4$-bit quantization.
(2)~Empirical evidence that truthful and hallucinated hidden states are linearly separable in mid-to-late transformer blocks, with consistent peak-probing blocks across model families---blocks~$13$--$18$ of~$32$ for Llama and Mistral, blocks~$19$--$25$ of~$28$ for Qwen.
(3)~A negative result: INSIDE and self-consistency provide no discriminative signal under a paired-label protocol, which we trace to a structural mismatch between the evaluation setup and the information these methods access (\S\ref{sec:discussion-sampling}).
%
%
\section{Related Work}
\label{sec:related}
\noindent
\textbf{\textit{Hallucination taxonomy.}}
\citet{ji2023survey} and \citet{huang2025survey} provide comprehensive surveys distinguishing \emph{intrinsic} (contradicting the input) from \emph{extrinsic} (unsupported by the input) hallucinations. We focus on the extrinsic / factual-correctness dimension, since all four of our datasets supply binary truthful/hallucinated labels per (prompt, answer) pair.

\noindent
\textbf{\textit{Representation-based detection.}}
Linear classifier probes \citep{alain2016understanding} expose information that is present, but not necessarily used, in a network's internal states. SAPLMA \citep{azaria2023internal} applies this idea to true/false classification of LLM-generated statements; CCS \citep{burns2022discovering} extracts a truth direction without labels; ITI \citep{li2023inference} edits activations along that direction to improve factuality. Our framework re-implements the SAPLMA per-layer probing protocol end-to-end on three modern open chat models and contrasts it with sampling-based baselines under matched compute.

\noindent
\textbf{\textit{Sampling-based uncertainty.}}
Self-consistency \citep{wang2022self} uses generation agreement as a confidence signal; SelfCheckGPT \citep{manakul2023selfcheckgpt} extends this to black-box hallucination detection. Semantic entropy \citep{kuhn2023semantic,farquhar2024detecting} clusters samples by meaning to obtain a more robust uncertainty estimate. INSIDE \citep{chen2024inside} replaces semantic clustering with the log-determinant of the regularized covariance of internal states across sampled completions (the \emph{EigenScore}); this is a recent and competitive sampling-and-internal-state hybrid, and we include it as our primary sampling baseline.

\noindent
\textbf{\textit{Cross-examination and external supervision.}}
Other lines of work query a second model to detect errors \citep{cohen2023lm} or supervise hallucination labels with external knowledge. We do not consider these here: our goal is white-box detection inside a single, deployed model.

\noindent
\textbf{\textit{Quantized inference.}}
$4$-bit NF4 \citep{dettmers2023qlora} and $8$-bit \citep{dettmers2022gpt3} quantization make $7$B--$8$B models accessible on commodity GPUs. Our results indicate that \emph{hidden-state probing remains effective under $4$-bit weight quantization}, with peak AUROC reaching $0.998$ on HaluEval-QA and $1.000$ on the synthetic benchmark.

\noindent
\textbf{\textit{Positioning.}}
Each of the above methods was typically evaluated in isolation or on a different model family, and we are not aware of a prior head-to-head, layer-resolved comparison of probe-based and sampling-based detectors on the same models under the same compute budget. Our study brings all four methods into a single framework and provides evidence that the evaluation protocol (paired-label vs.\ generate-then-judge) is a key factor governing their relative performance (\S\ref{sec:method-paired}).
%
%
\section{Methodology}
\label{sec:method}

\noindent
\textbf{\textit{Models.}}
We study three publicly accessible open-source chat models loaded in $4$-bit NF4 quantization \citep{dettmers2023qlora} with double quantization and \texttt{bfloat16} compute dtype (see Appendix~\ref{sec:appendix-hyperparams} for full configuration details). The three models, summarized in Table~\ref{tab:models}, span three widely deployed open-source families (Llama, Mistral, Qwen) and two architectural sizes ($28$- and $32$-block transformers).

\begin{table}[t!]
  \centering
  \small
\caption{Models used in this study, all loaded in $4$-bit NF4 quantization with \texttt{bfloat16} compute dtype. VRAM denotes the quantized weight footprint, exclusive of activations.}
  \label{tab:models}
  \begin{tabular}{lccc}
    \toprule
\textbf{Model} & \textbf{Blocks} & \textbf{Hidden dim} & \textbf{VRAM} \\
    \midrule
Llama-3.1-8B-Instruct    & $32$ & $4096$ & ${\sim}5.5$\,GB \\
Mistral-7B-Instruct-v0.3 & $32$ & $4096$ & ${\sim}4.8$\,GB \\
Qwen2.5-7B-Instruct      & $28$ & $3584$ & ${\sim}4.5$\,GB \\
    \bottomrule
  \end{tabular}
\end{table}
\noindent
\textbf{\textit{Datasets.}}
We cap each dataset at $N\!=\!400$ items. For every dataset we pair a truthful ($\text{label}{=}1$) and a hallucinated ($\text{label}{=}0$) answer for the same prompt, yielding near-perfect $50{:}50$ class balance (positive-class ratios range from $0.500$ to $0.513$).

\textit{TruthfulQA.}
We use the multiple-choice configuration of TruthfulQA \citep{lin2022truthfulqa} and, for each question, pair the canonical correct choice with one randomly sampled incorrect choice. This produces controlled adversarial pairs of plausible but wrong continuations and is the most difficult of our four benchmarks because the wrong answers are explicitly chosen to mimic plausible human errors.

\textit{HaluEval-QA.}
HaluEval \citep{li2023halueval} provides a knowledge-conditioned QA split with paired correct (label~$1$) and hallucinated (label~$0$) answers generated by GPT-style models. We prepend the supplied knowledge to the question, giving the model an explicit evidence channel that the other datasets lack.

\textit{FEVER.}
FEVER \citep{thorne2018fever} provides Wikipedia-supported claims labelled as \emph{supported} ($1$) or \emph{refuted} ($0$); claims with insufficient evidence are excluded to preserve binary clarity.

\textit{Synthetic.}
A controlled benchmark of paired true/false statements about world capitals, chemical symbols, literary authorship, and planetary orbits, generated from four knowledge banks (see Appendix~\ref{sec:appendix-synthetic}). Both items in a pair share the same prompt, differing only in a single substantive token, giving a clean class signal.

\subsection{Paired-Label Protocol}
\label{sec:method-paired}

A design choice central to interpreting our results is the \emph{paired-label protocol}. In each dataset, the candidate answer is supplied by the benchmark rather than generated by the model. We concatenate prompt and candidate answer, run a single forward pass, and extract hidden states following the SAPLMA convention \citep{azaria2023internal}. Probe-based detectors thus observe the model's internal representation of the supplied answer, whereas sampling-based detectors (INSIDE, self-consistency) draw $K$ free continuations from the prompt alone and measure their diversity \emph{without conditioning on the supplied candidate}.

This asymmetry has two consequences. First, probes receive a clean binary signal with near-perfect $50{:}50$ class balance. Second, the $K$ stochastic completions are unconstrained by the supplied candidate, so the EigenScore and self-consistency scores measure \emph{generative diversity}, which is only weakly correlated with the truthfulness of the supplied answer. The two detector families thus answer \emph{different questions} under this protocol (\S\ref{sec:discussion-sampling}).

\subsection{Hidden-state and attention extraction}
\label{sec:method-extraction}

For every $\langle\text{prompt},\text{answer}\rangle$ pair we run a single forward pass and extract all $L\!+\!1$ hidden-state tensors (embedding output plus $L$ transformer-block outputs). Each tensor is reduced to a single vector via \emph{last-token} pooling over the answer span, following the SAPLMA convention \citep{azaria2023internal}, yielding a representation of shape $[N, L\!+\!1, D]$ per (model, dataset) pair ($N\!=\!400$, $D$ = hidden dimensionality).

For attention analysis we capture attention weight matrices at the first, middle, and last transformer blocks (indices $0$, $\lfloor L/2\rfloor$, $L\!-\!1$). From each block we compute the Shannon entropy of the last-answer-token attention distribution per head, yielding a compact per-head entropy vector $\mathbf{e}\in\mathbb{R}^{H}$.

\subsection{Detection methods}
\label{sec:method-detection}

\textit{SAPLMA-style probes.}
For each block index $\ell\in\{0,\dots,L\}$ we train two probes on the $D$-dimensional hidden vector: (a)~a \emph{Linear} probe -- a single affine logistic head; (b)~an \emph{MLP} probe -- two hidden layers ($256\!\to\!64$, GELU, $0.2$ dropout) followed by a softmax head. Probes are trained with AdamW ($\text{lr}\!=\!10^{-3}$, weight decay $10^{-4}$) for $30$ epochs with batch size $128$, on a stratified $70\!:\!10\!:\!20$ train/val/test split selected by class. We run each probe with $3$ seeds and report the mean and standard deviation of AUROC, AUPRC, accuracy and $F_1$. The peak block and peak AUROC summarize the model's probing score.

\textit{INSIDE EigenScore.}
Following \citet{chen2024inside}, for each prompt we sample $K{=}10$ stochastic continuations (temperature $0.7$, top-$p$ $0.95$, $64$ tokens) and capture the last-token hidden state at the final block for each completion. Stacking these into $\mathbf{H}\in\mathbb{R}^{K\times D}$, the EigenScore is
\begin{equation}
\mathrm{EigenScore}(\mathbf{H}) \;=\; \log\det\!\bigl(\tfrac{1}{D}\,\mathbf{X}\mathbf{X}^{\!\top} \!+\alpha I_K\bigr),
\label{eq:eigenscore}
\end{equation}
where $\mathbf{X}=\mathbf{H}-\bar{\mathbf{H}}$ centers the rows and $\alpha{=}10^{-3}$. We compute in the $K\!\times\!K$ Gram space for efficiency. A higher EigenScore indicates greater dispersion across completions and is interpreted as evidence of hallucination.

\textit{Self-consistency.}
For each prompt we draw $K{=}5$ stochastic completions with the same generation parameters and measure their agreement via (a)~exact-match plurality (largest equivalence class of normalized strings divided by $K$) and (b)~the mean pairwise cosine similarity of their sentence embeddings \citep{reimers2019sentence}. The hallucination score is $1 - \text{consistency}$.

\textit{Attention entropy.} 
For each kept attention block $\ell\in\{0,\lfloor L/2\rfloor,L\!-\!1\}$ we summarize the last-answer-token attention distribution by its head-averaged Shannon entropy, and treat that scalar as a continuous hallucination signal (high entropy $\Rightarrow$ diffuse attention $\Rightarrow$ unclear evidence).

\textit{Metrics and layer geometry.}
\label{sec:method-metrics}
Probes are evaluated on the held-out test split with AUROC, AUPRC, accuracy, and $F_1$ (threshold $0.5$). The remaining signals are evaluated with AUROC and AUPRC directly, plus accuracy and $F_1$ at the Youden-$J$ optimal threshold. We additionally report classifier-free geometric statistics per block: centroid distance, within-class spread, and their ratio (the \emph{separation ratio}), which tracks whether class separability develops as a geometric property of the representations independently of any probe.

\paragraph{Reproducibility and compute.}
\label{sec:method-reproducibility}
All random seeds and determinism flags are fixed (see Appendix~\ref{sec:appendix-hyperparams} for the complete list). The full experiment runs end-to-end on a single RTX~5060 ($8$\,GB) in roughly $4$--$8$~hours; the bulk of wall-clock time is consumed by the INSIDE and self-consistency stages, each of which requires $K$ stochastic generations per prompt across all $12$ model--dataset configurations.
%
%
\section{Experiments}
\label{sec:experiments}

For each of the $12$ model--dataset combinations we extract hidden states and attention summaries in a single forward pass, then run all four detection methods. Results are organized into four groups: best-block detection performance, layer-wise trajectories, class geometry, and attention entropy.

\subsection{Best-block detection performance}
\label{sec:experiments-best-layer}

Table~\ref{tab:headline} reports the best-block AUROC of each method for every model--dataset pair. For probes, ``best block'' is the layer $\mathrm{argmax}_{\ell}\,\overline{\mathrm{AUROC}}_{\ell}$ across three seeds ($\ell\in\{0,\dots,L\}$, where $0$ is the embedding output and $L$ is the final transformer block); for INSIDE and self-consistency, a single score per prompt is produced; for attention entropy we report the best of the three captured blocks (first, middle, last).

\begin{table}[t!]
\centering
\scriptsize
\setlength{\tabcolsep}{3pt}
\caption{Best-block AUROC per (model, dataset) cell. \textbf{Linear}/\textbf{MLP}: SAPLMA-style probes; parenthetical numbers are the best block index ($0$\,=\,embedding, $L$\,=\,final block). \textbf{INSIDE}: EigenScore at the final block ($K{=}10$). \textbf{SC}: self-consistency ($K{=}5$), exact-match or embedding. \textbf{Attn}: head-averaged last-token attention entropy at the best of three blocks. Bold marks the best probe per row. Probe values are means over three seeds (std~$\leq\!0.025$). Llama\,=\,Llama-3.1-8B, Mistral\,=\,Mistral-7B, Qwen\,=\,Qwen2.5-7B.}
\label{tab:headline}
\begin{tabular}{llcccccc}
\toprule
\textbf{Dataset} & \textbf{Model} & \textbf{Linear} & \textbf{MLP} & \textbf{INSIDE} & \textbf{SC\textsubscript{ex}} & \textbf{SC\textsubscript{sem}} & \textbf{Attn} \\
\midrule
\multirow{3}{*}{TruthfulQA}
 & Llama  & \textbf{.914}\,(13) & \textbf{.914}\,(14) & .433 & .500 & .511 & .603\,(31) \\
 & Mistral & .904\,(16) & \textbf{.908}\,(16) & .525 & .502 & .541 & .581\,(31) \\
 & Qwen   & .915\,(19) & \textbf{.925}\,(19) & .506 & .502 & .478 & .489\,(14) \\
\midrule
\multirow{3}{*}{HaluEval}
 & Llama  & \textbf{.998}\,(15) & .998\,(15) & .529 & .502 & .489 & .941\,(0)  \\
 & Mistral & \textbf{.998}\,(18) & .998\,(18) & .439 & .467 & .465 & .902\,(0)  \\
 & Qwen   & .997\,(19) & \textbf{.998}\,(25) & .491 & .505 & .522 & .866\,(0)  \\
\midrule
\multirow{3}{*}{FEVER}
 & Llama  & \textbf{.955}\,(17) & .953\,(17) & .518 & .500 & .468 & .686\,(16) \\
 & Mistral & \textbf{.960}\,(14) & .959\,(15) & .527 & .500 & .425 & .580\,(16) \\
 & Qwen   & .921\,(20) & \textbf{.927}\,(21) & .497 & .502 & .466 & .609\,(27) \\
\midrule
\multirow{3}{*}{Synthetic}
 & Llama  & \textbf{1.00}\,(32) & 1.00\,(28) & .518 & .500 & .502 & .639\,(31) \\
 & Mistral & .998\,(18) & \textbf{.999}\,(20) & .528 & .500 & .502 & .774\,(16) \\
 & Qwen   & \textbf{1.00}\,(17) & 1.00\,(18) & .489 & .500 & .502 & .683\,(27) \\
\bottomrule
\end{tabular}
\end{table}

Three findings emerge from Table~\ref{tab:headline}:

\noindent\textbf{(F1)~Probes yield higher AUROC than sampling-based detectors in this setting.} The best probe AUROC exceeds the best sampling-based AUROC by at least $0.367$ in every cell (median gap $0.453$). On HaluEval-QA all probes reach $\geq\!0.997$; on TruthfulQA, the most challenging of our four benchmarks, probes attain $0.904$--$0.925$.

\noindent\textbf{(F2)~MLP $\approx$ Linear.} The MLP probe rarely exceeds the linear probe by more than $0.01$ AUROC: the absolute MLP--Linear difference is ${\leq}\,0.005$ in $10$ of $12$ cells and ${\leq}\,0.01$ in all~$12$; the largest gap is $+0.010$ (TruthfulQA, Qwen2.5). This suggests that the truthfulness signal is approximately \emph{linear} in the hidden-state space, consistent with \citet{azaria2023internal,burns2022discovering}.

\begin{figure}[t!]
  \centering
  \includegraphics[width=0.65\textwidth]{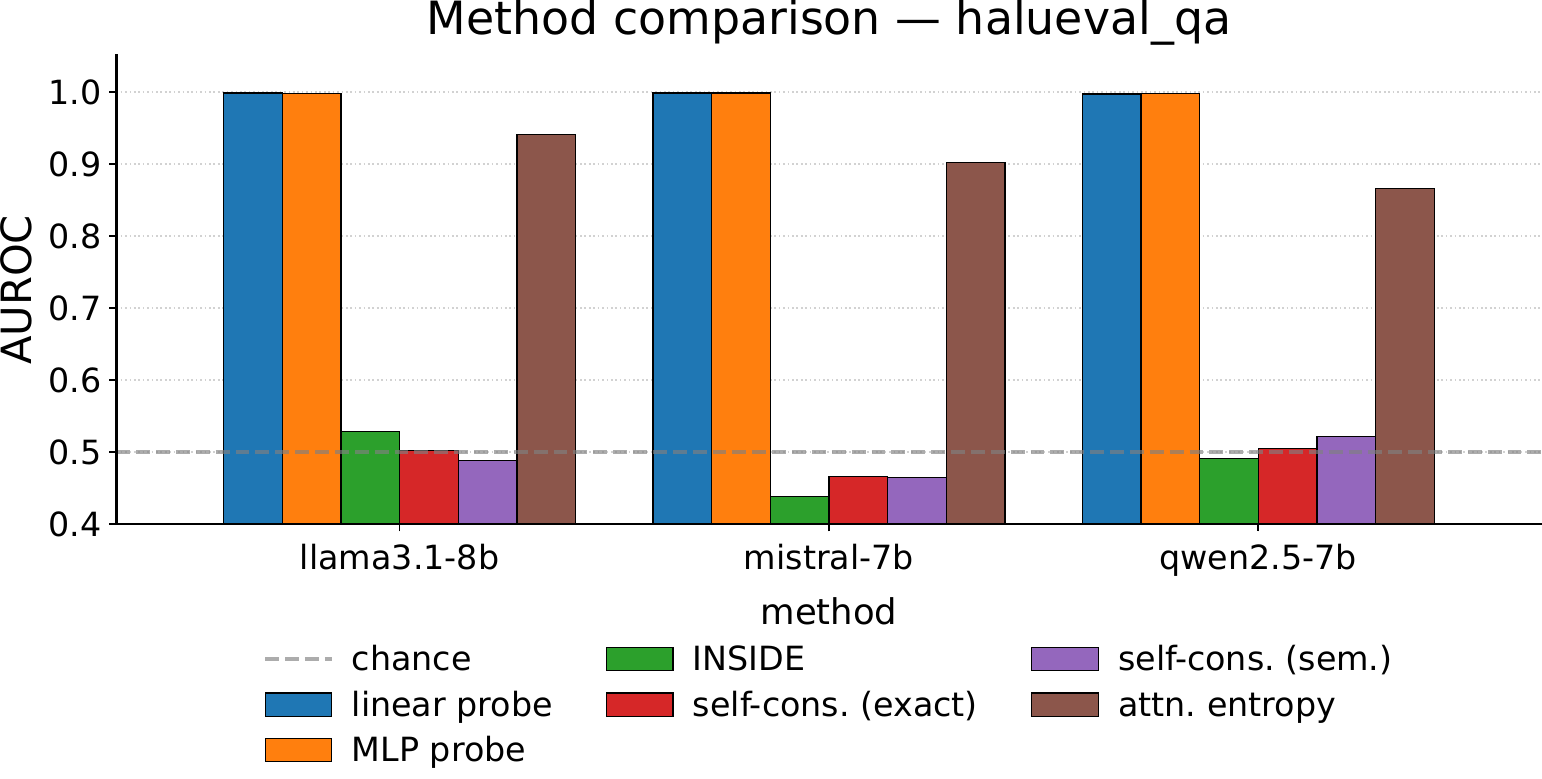}
  \vspace{-2mm}
  \caption{Method comparison on HaluEval-QA. Probes reach $\geq\!0.997$ AUROC on all three models; attention entropy at the first block yields $0.866$--$0.941$ AUROC at no extra inference cost; INSIDE and self-consistency are at chance.}
  \label{fig:method-comparison}
\end{figure}

\noindent\textbf{(F3)~INSIDE and self-consistency are near chance under this protocol.} The INSIDE EigenScore lies in $[0.433,\,0.529]$ across all~$12$ cells; exact-match self-consistency in $[0.467,\,0.505]$; and embedding self-consistency in $[0.425,\,0.541]$. No sampling method exceeds $0.55$ AUROC on any cell. We attribute this to our evaluation protocol (\S\ref{sec:method-paired}), discussed further in \S\ref{sec:experiments-sampling} and \S\ref{sec:discussion-sampling}. Figure~\ref{fig:method-comparison} visualizes the pattern on HaluEval-QA: probes dominate, attention entropy is second, and sampling variants cluster at chance.

Tables~\ref{tab:accuracy} and \ref{tab:f1} confirm the same pattern with accuracy and $F_1$: probes achieve $\geq\!0.800$ accuracy on every natural-language benchmark; sampling methods remain near chance; attention entropy is again the exception on HaluEval-QA ($0.803$--$0.863$ accuracy).

\begin{table}[t!]
\centering
\scriptsize
\setlength{\tabcolsep}{3pt}
\caption{Accuracy at the best block. Probe threshold is $0.5$; other methods use the Youden-$J$ optimal threshold. Bold marks the best probe variant per row.}
\label{tab:accuracy}
\begin{tabular}{llcccccc}
\toprule
\textbf{Dataset} & \textbf{Model} & \textbf{Linear} & \textbf{MLP} & \textbf{INSIDE} & \textbf{SC\textsubscript{ex}} & \textbf{SC\textsubscript{sem}} & \textbf{Attn} \\
\midrule
\multirow{3}{*}{TruthfulQA}
 & Llama  & \textbf{.825} & .813 & .528 & .510 & .533 & .598 \\
 & Mistral & .800 & \textbf{.821} & .558 & .498 & .553 & .590 \\
 & Qwen   & .808 & \textbf{.833} & .540 & .498 & .515 & .518 \\
\midrule
\multirow{3}{*}{HaluEval}
 & Llama  & \textbf{.979} & .971 & .553 & .500 & .520 & .863 \\
 & Mistral & .967 & \textbf{.971} & .503 & .500 & .505 & .850 \\
 & Qwen   & \textbf{.971} & .967 & .515 & .505 & .540 & .803 \\
\midrule
\multirow{3}{*}{FEVER}
 & Llama  & .858 & \textbf{.863} & .538 & .513 & .510 & .683 \\
 & Mistral & \textbf{.900} & .900 & .560 & .513 & .503 & .578 \\
 & Qwen   & .821 & \textbf{.854} & .520 & .490 & .528 & .613 \\
\midrule
\multirow{3}{*}{Synthetic}
 & Llama  & .933 & \textbf{.988} & .540 & .500 & .525 & .638 \\
 & Mistral & .913 & \textbf{.975} & .540 & .500 & .528 & .735 \\
 & Qwen   & .963 & \textbf{.992} & .518 & .500 & .523 & .663 \\
\bottomrule
\end{tabular}
\end{table}

\begin{table}[t!]
\centering
\scriptsize
\setlength{\tabcolsep}{3pt}
\caption{$F_1$ at the best block. Bold marks the best probe variant per row. SC\textsubscript{ex} values are unstable because exact-match self-consistency often assigns identical scores to all samples, yielding degenerate $F_1$.}
\label{tab:f1}
\begin{tabular}{llcccccc}
\toprule
\textbf{Dataset} & \textbf{Model} & \textbf{Linear} & \textbf{MLP} & \textbf{INSIDE} & \textbf{SC\textsubscript{ex}} & \textbf{SC\textsubscript{sem}} & \textbf{Attn} \\
\midrule
\multirow{3}{*}{TruthfulQA}
 & Llama  & \textbf{.833} & .818 & .681 & .675 & .490 & .571 \\
 & Mistral & .807 & \textbf{.829} & .645 & .038 & .647 & .531 \\
 & Qwen   & .819 & \textbf{.842} & .621 & .038 & .606 & .499 \\
\midrule
\multirow{3}{*}{HaluEval}
 & Llama  & \textbf{.980} & .971 & .589 & .074 & .625 & .858 \\
 & Mistral & .967 & \textbf{.971} & .668 & .010 & .124 & .840 \\
 & Qwen   & \textbf{.971} & .967 & .378 & .039 & .665 & .806 \\
\midrule
\multirow{3}{*}{FEVER}
 & Llama  & .847 & \textbf{.870} & .575 & .678 & .581 & .654 \\
 & Mistral & \textbf{.904} & .904 & .553 & .678 & .167 & .663 \\
 & Qwen   & .811 & \textbf{.864} & .319 & .010 & .677 & .498 \\
\midrule
\multirow{3}{*}{Synthetic}
 & Llama  & .938 & \textbf{.988} & .324 & .667 & .379 & .640 \\
 & Mistral & .922 & \textbf{.974} & .523 & .667 & .404 & .725 \\
 & Qwen   & .964 & \textbf{.992} & .499 & .667 & .565 & .536 \\
\bottomrule
\end{tabular}
\end{table}

\subsection{Layer-wise probing trajectories}
\label{sec:experiments-layerwise}

\begin{figure}[t!]
  \centering
  \includegraphics[width=0.48\textwidth]{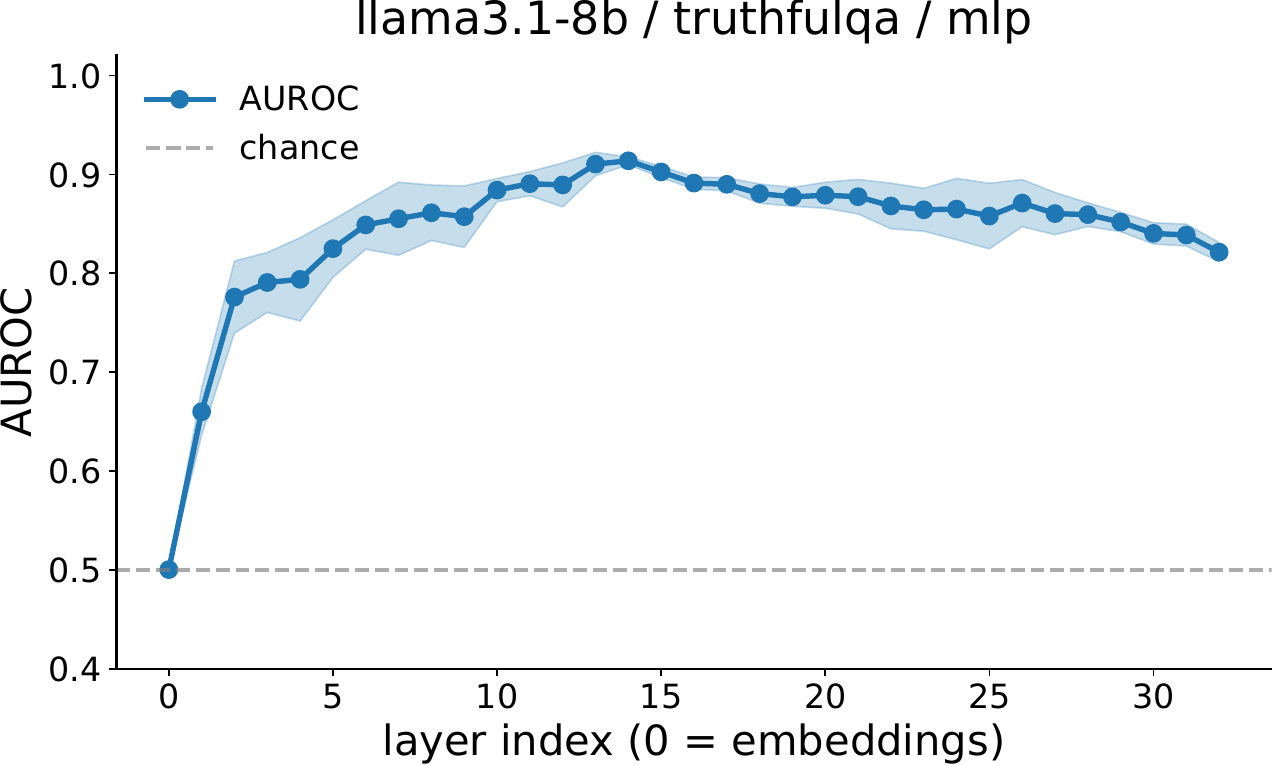}\hfill
  \includegraphics[width=0.48\textwidth]{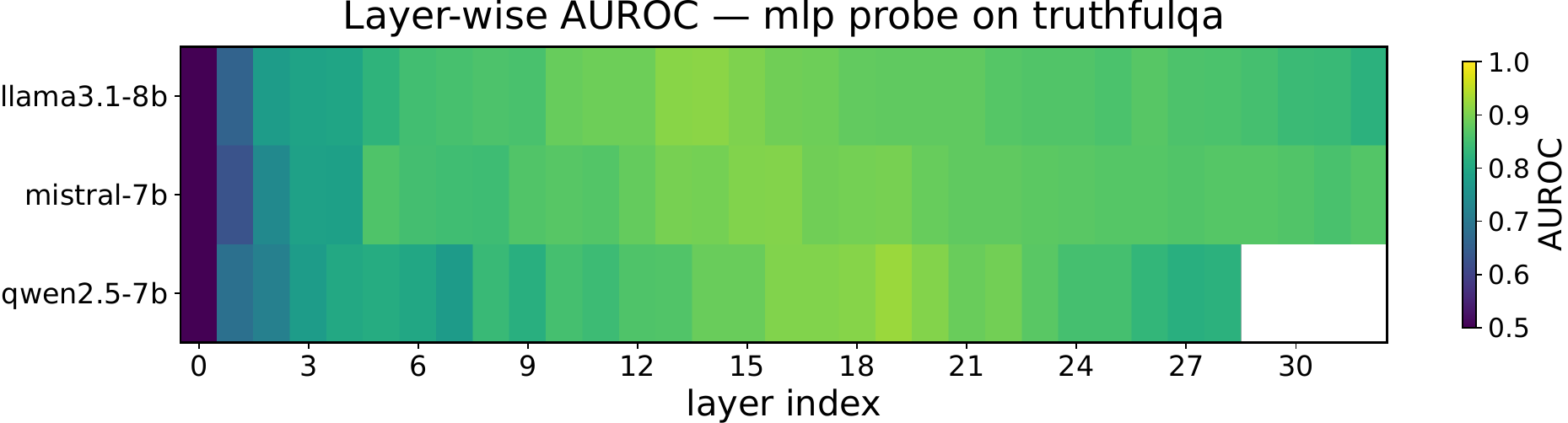}
  \vspace{-2mm}
\caption{\emph{Left}: Layer-wise MLP-probe AUROC for Llama-3.1-8B on TruthfulQA; the signal peaks near block~$14$ (shaded band: std over three seeds). \emph{Right}: Per-block MLP-probe AUROC heatmap across all three models on TruthfulQA; the peak band is consistently in the second half of the network.}
  \label{fig:auroc-layer}
  \label{fig:heatmap-truthful}
\end{figure}

Probing performance varies smoothly with depth (Figure~\ref{fig:auroc-layer}). For Llama-3.1-8B on TruthfulQA, the MLP-probe AUROC peaks near block~$14$ ($\approx\!0.914$, std $<\!0.005$). The heatmap (right panel) shows the peak band lies in $40\%$--$90\%$ of network depth across models; all $12$ cells exhibit qualitatively similar trajectories (Appendix~\ref{sec:appendix-auroc-all}).

This pattern is consistent across all four datasets (Appendix~\ref{sec:appendix-heatmaps}). On natural-language benchmarks the MLP-probe peak blocks for Llama, Mistral, and Qwen are $\{14,16,19\}$ on TruthfulQA, $\{15,18,25\}$ on HaluEval-QA, and $\{17,15,21\}$ on FEVER (Table~\ref{tab:headline}). On the synthetic benchmark the peak shifts later ($\{28,20,18\}$) because the AUROC plateau reaches~$1.000$ across several late blocks, making the argmax arbitrary.

\subsection{Class geometry through depth}
\label{sec:experiments-geometry}

\begin{figure}[t!]
  \centering
  \includegraphics[width=0.65\textwidth]{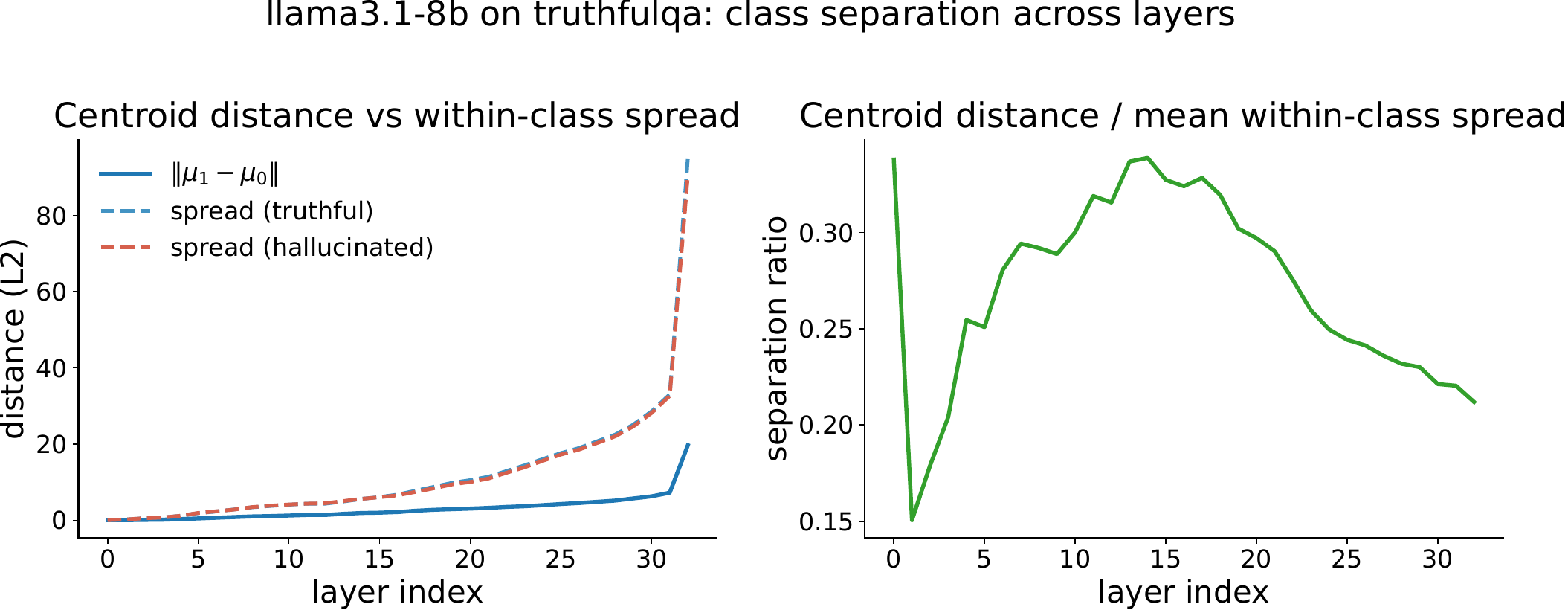}\hfill
  \includegraphics[width=0.35\textwidth]{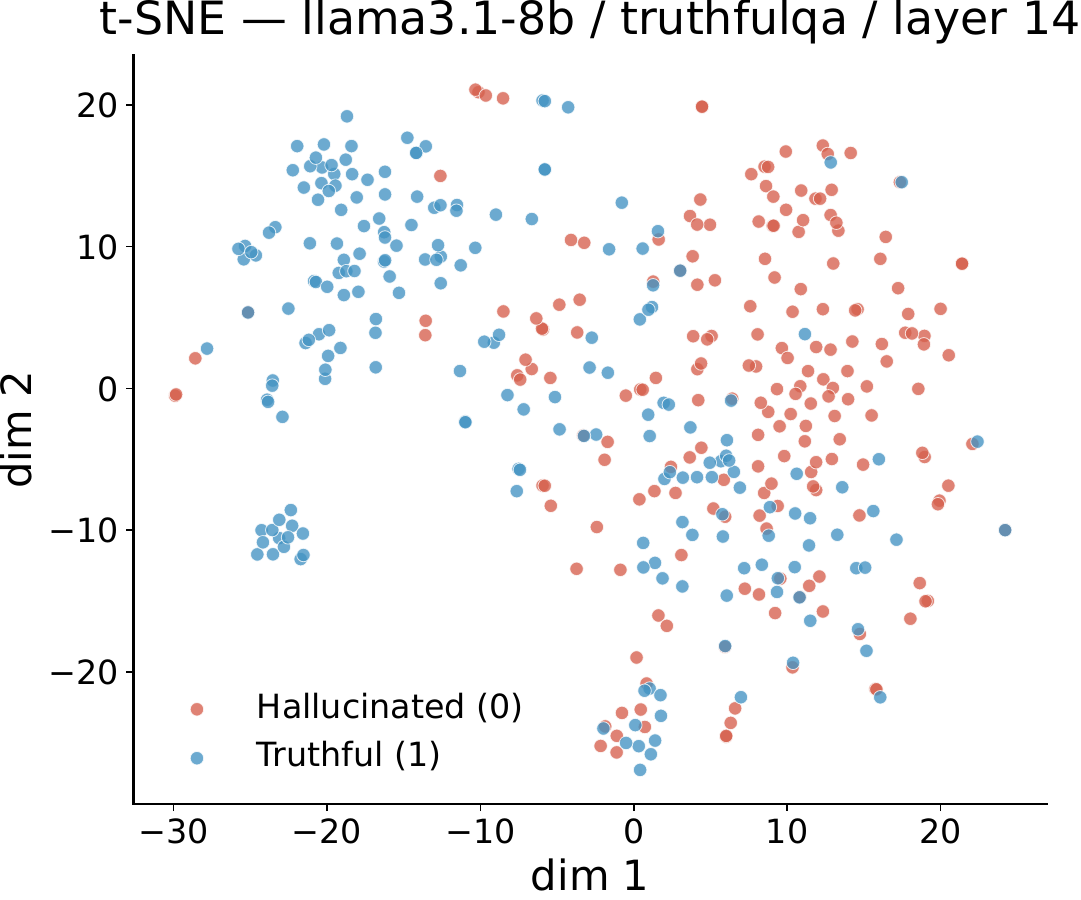}
  \vspace{-2mm}
\caption{\emph{Left}: Layer-wise class geometry for Llama-3.1-8B on TruthfulQA showing centroid distance, within-class spread, and their ratio; the separation ratio plateaus in the second half of the network, mirroring probe AUROC (Figure~\ref{fig:auroc-layer}). \emph{Right}: t-SNE projection of layer-$14$ hidden states; truthful (blue) and hallucinated (red) samples form distinct clusters, consistent with $0.914$ AUROC \citep{van2008visualizing}.}
  \label{fig:separation}
  \label{fig:tsne}
\end{figure}

Probing performance mirrors the underlying hidden-state geometry. Figure~\ref{fig:separation} (left) shows the separation ratio for Llama-3.1-8B on TruthfulQA: it increases through early blocks and plateaus where probes peak, confirming that class separability is a geometric property of the representations. The t-SNE projection at the probing peak (right) shows visually distinct clusters. This behavior is reproduced across all models and datasets (Appendix~\ref{sec:appendix-separation},~\ref{sec:appendix-sep-all}).

\subsection{Attention entropy as an auxiliary signal}
\label{sec:experiments-attention}

\begin{figure}[t!]
  \centering
  \includegraphics[width=0.68\textwidth]{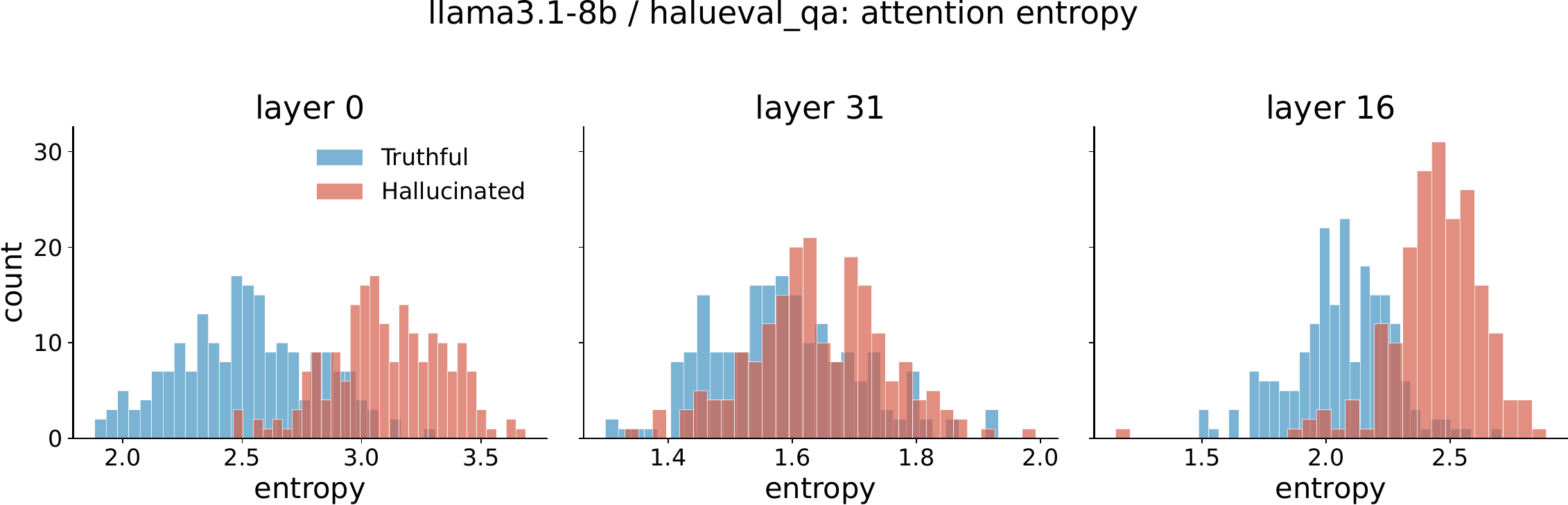}\hfill
  \includegraphics[width=0.32\textwidth]{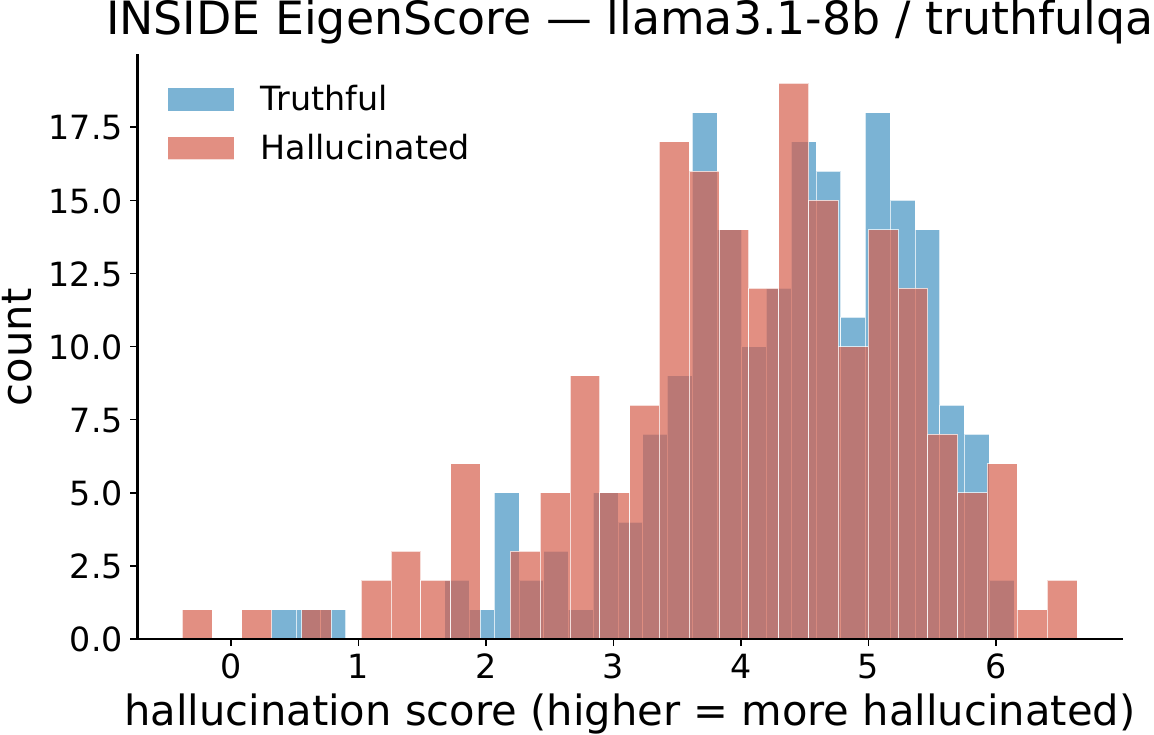}
  \vspace{-2mm}
\caption{\emph{Left}: Last-token attention entropy on Llama-3.1-8B / HaluEval-QA at three transformer blocks; the first block produces the best class separation (AUROC $0.941$). \emph{Right}: INSIDE EigenScore distribution on Llama-3.1-8B / TruthfulQA; the truthful and hallucinated histograms overlap almost completely (AUROC $0.433$).}
  \label{fig:attention}
  \label{fig:inside-dist}
\end{figure}

Figure~\ref{fig:attention} (left) shows the head-averaged last-token attention entropy on HaluEval-QA for Llama-3.1-8B. At the first block, hallucinated items produce higher entropy (AUROC $0.941$): when the supplied knowledge supports the answer, attention concentrates on the relevant tokens.

This effect is robust across models on HaluEval-QA (AUROC $0.866$--$0.941$; Appendix~\ref{sec:appendix-attn-cross}) but does not generalize to datasets lacking an explicit evidence passage (Appendix~\ref{sec:appendix-attn-all}). For retrieval-augmented or knowledge-conditioned settings, first-block attention entropy requires no additional forward pass and complements a mid-network probe at no extra cost.

\subsection{Why sampling-based methods underperform}
\label{sec:experiments-sampling}

As shown in Figure~\ref{fig:inside-dist} (right), the per-class INSIDE EigenScore distributions are nearly indistinguishable on TruthfulQA; the same holds across all configurations (Appendix~\ref{sec:appendix-inside-dist},~\ref{sec:appendix-inside-all}). Exact-match self-consistency is effectively $0.50$ on $11/12$ cells, as stochastic completions rarely produce identical strings. Embedding self-consistency never exceeds $0.541$ AUROC, consistent with prior findings that simple agreement metrics are insufficient without semantic-entropy clustering \citep{kuhn2023semantic,farquhar2024detecting}. This outcome is inherent to our paired-label protocol (\S\ref{sec:method-paired},~\S\ref{sec:discussion-sampling}).
%
%
\section{Discussion}
\label{sec:discussion}

\paragraph{Why probes are strong.}
\label{sec:discussion-probes}
Three properties of our experimental setting contribute to the high probe AUROC observed. First, our evaluation protocol (\S\ref{sec:method-paired}) passes each candidate answer through the network, giving the probe access to the model's internal representation of that specific answer. Second, balanced pairing of truthful and hallucinated answers for every prompt yields clean binary supervision with low variance ($\leq\!0.025$ AUROC across seeds). Third, instruction-tuned chat models \citep{ouyang2022training,grattafiori2024llama,jiang2023mistral,qwen2024} may develop more linearly decodable truthfulness directions than base models, though we leave this comparison to future work.

\paragraph{Why sampling methods are weak here.}
\label{sec:discussion-sampling}
As detailed in \S\ref{sec:method-paired} and confirmed empirically in \S\ref{sec:experiments-sampling}, sampling-based and probe-based detectors answer \emph{different questions}: probes measure whether the network internally distinguishes a given answer as truthful, while sampling methods measure the model's generative uncertainty about the prompt. Under paired-label evaluation, probes have a structural advantage; in generate-then-judge settings, sampling approaches would be expected to recover more of their reported effectiveness. This distinction is relevant for future benchmark design.

\paragraph{Probing remains effective under quantization.}
\label{sec:discussion-quantization}
The peak AUROC values ($0.998$ on HaluEval-QA, $1.000$ on the synthetic benchmark) are obtained entirely under $4$-bit quantization and are comparable to full-precision results reported on similar models \citep{azaria2023internal}. While we do not include a direct full-precision baseline, this comparison suggests that hidden-state detectors can be deployed at the same inference cost as the chat model itself, without requiring de-quantization.

\paragraph{Where to attach the probe.}
\label{sec:discussion-where}
The MLP-probe peak falls in a consistent band: blocks~$14$--$18$ for Llama and Mistral ($44$--$56\%$ of depth) and blocks~$19$--$25$ for Qwen ($68$--$89\%$). In practice, a single fixed block (e.g.,~${\approx}15$ for Llama,~${\approx}16$ for Mistral,~${\approx}20$ for Qwen) would incur at most $0.02$ AUROC loss relative to the per-dataset optimum. On the synthetic benchmark the AUROC plateau is close enough to $1.000$ that any block in the second half of the network suffices.
%
%
\section{Limitations}
\label{sec:limitations}

Our study is restricted to $7$B--$8$B chat models under $4$-bit quantization, evaluated on four English-language benchmarks of $400$ items each; results may differ for larger or base models, non-English languages, and free-form generations without benchmark-supplied candidate answers. INSIDE uses final-block hidden states with $K{=}10$ samples and self-consistency uses $K{=}5$; a budget sweep, intermediate-layer EigenScores, or semantic-entropy clustering might recover some discriminative power. The fixed sampling temperature ($0.7$) and top-$p$ ($0.95$) may underestimate the dispersion these methods rely on. Attention entropy is captured at only three blocks; a denser sweep might reveal stronger intermediate-block signals. Finally, all probes are trained and tested on the same dataset; cross-dataset transfer is an important next step.
%
%
\section{Conclusion}
\label{sec:conclusion}

We presented a unified evaluation of four hallucination-detection methods on three $7$B--$8$B quantized chat models, reproducible on a single consumer GPU. Under the paired-label protocol, a linear probe on a single mid-network hidden state achieves $0.904$--$1.000$ AUROC across three model families and four datasets, while sampling-based detectors do not exceed $0.541$. The peak probing block is stable across architectures on natural-language benchmarks, and the separation it captures is corroborated by classifier-free geometric analysis. First-block attention entropy provides a complementary signal at no additional inference cost in knowledge-grounded settings. These results suggest that white-box probing is both more effective and less expensive than sampling-based detection in this regime. The observed weakness of sampling methods reflects the structural mismatch inherent to paired-label evaluation rather than a limitation of those methods in general.

\paragraph{Data, code, and supplementary material.}
All source code, trained probe weights, and evaluation scripts are publicly available at \url{https://github.com/Ezharjan/HallucinationPatternDetection}. The repository includes the complete pipeline, pre-computed results, and the eighteen supplementary appendices referenced above.
\begin{credits}
\subsubsection{\discintname}
The authors have no competing interests to declare that are relevant to the content of this article.
\end{credits}
%
%
\bibliographystyle{splncs04}
\bibliography{ref}

\appendix


\section*{Appendix Roadmap}
\label{sec:appendix-roadmap}

The eighteen appendices that follow provide exhaustive supporting evidence for every main-text claim. The roadmap below maps each appendix to its primary function and lists the figures it contains, so that every one of the $28$ appendix figures and the appendix hyperparameter table (Table~\ref{tab:hyper}) can be located by cross-reference.

\begin{itemize}
\item \textbf{Appendix~\ref{sec:appendix-method-comparison}} extends the method-comparison bar plot (main-text Figure~\ref{fig:method-comparison}) to the remaining three datasets (Figure~\ref{fig:method-all}: TruthfulQA, FEVER, Synthetic), supporting Finding~F1 (\S\ref{sec:experiments-best-layer}).

\item \textbf{Appendix~\ref{sec:appendix-heatmaps}} provides MLP-probe per-block AUROC heatmaps on HaluEval-QA, FEVER and the synthetic benchmark (Figure~\ref{fig:heatmap-all}), complementing Figure~\ref{fig:heatmap-truthful} and supporting the peak-band consistency claim of \S\ref{sec:experiments-layerwise}.

\item \textbf{Appendix~\ref{sec:appendix-trajectories}} shows layer-AUROC trajectories for Mistral and Qwen on TruthfulQA (Figure~\ref{fig:auroc-cross}) and Llama on HaluEval-QA (Figure~\ref{fig:auroc-easy}), extending Figure~\ref{fig:auroc-layer}.

\item \textbf{Appendix~\ref{sec:appendix-separation}} shows class-separation diagnostics for Mistral and Qwen on TruthfulQA (Figure~\ref{fig:sep-cross}), extending Figure~\ref{fig:separation}.

\item \textbf{Appendix~\ref{sec:appendix-inside-dist}} shows INSIDE EigenScore distributions for Mistral and Qwen on TruthfulQA (Figure~\ref{fig:inside-cross}), extending Figure~\ref{fig:inside-dist} and supporting Finding~F3.

\item \textbf{Appendix~\ref{sec:appendix-attn-cross}} shows attention-entropy distributions for Mistral and Qwen on HaluEval-QA (Figure~\ref{fig:attn-cross}), extending Figure~\ref{fig:attention} and supporting the attention-entropy result of \S\ref{sec:experiments-attention}.

\item \textbf{Appendix~\ref{sec:appendix-pca}} provides side-by-side PCA projections at the probing peak for Llama and Qwen (Figure~\ref{fig:pca}), complementing the t-SNE visualization in Figure~\ref{fig:tsne}.

\item \textbf{Appendix~\ref{sec:appendix-tsne-qwen}} provides the Qwen t-SNE projection at its probing peak (Figure~\ref{fig:tsne-qwen}), extending Figure~\ref{fig:tsne}.

\item \textbf{Appendix~\ref{sec:appendix-synthetic}} describes the synthetic benchmark construction (no figures).

\item \textbf{Appendix~\ref{sec:appendix-heatmaps-linear}} provides per-block AUROC heatmaps for the \emph{linear} probe across all four datasets (Figure~\ref{fig:heatmap-linear-all}), supporting Finding~F2 (\S\ref{sec:experiments-best-layer}).

\item \textbf{Appendix~\ref{sec:appendix-auroc-all}} reports layer-AUROC trajectories for all $12$ configurations: Llama (Figure~\ref{fig:auroc-llama-all}), Mistral (Figure~\ref{fig:auroc-mistral-all}), Qwen (Figure~\ref{fig:auroc-qwen-all}).

\item \textbf{Appendix~\ref{sec:appendix-sep-all}} reports class-geometry diagnostics for all $12$ configurations: Llama (Figure~\ref{fig:sep-llama-all}), Mistral (Figure~\ref{fig:sep-mistral-all}), Qwen (Figure~\ref{fig:sep-qwen-all}).

\item \textbf{Appendix~\ref{sec:appendix-inside-all}} reports INSIDE EigenScore histograms for all $12$ configurations: Llama (Figure~\ref{fig:inside-llama-all}), Mistral (Figure~\ref{fig:inside-mistral-all}), Qwen (Figure~\ref{fig:inside-qwen-all}).

\item \textbf{Appendix~\ref{sec:appendix-attn-all}} reports attention-entropy distributions on TruthfulQA (Figure~\ref{fig:attn-truthfulqa-all}), FEVER (Figure~\ref{fig:attn-fever-all}), and the synthetic benchmark (Figure~\ref{fig:attn-synthetic-all}) for all three models, substantiating the claim that the first-block signal does not generalize beyond knowledge-grounded settings.

\item \textbf{Appendix~\ref{sec:appendix-proj-all}} provides 2-D PCA and t-SNE projections for all $12$ configurations: Llama (Figure~\ref{fig:proj-llama-all}), Mistral (Figure~\ref{fig:proj-mistral-all}), Qwen (Figure~\ref{fig:proj-qwen-all}).

\item \textbf{Appendix~\ref{sec:appendix-acc-f1}} provides thresholding details for the accuracy and $F_1$ tables (Tables~\ref{tab:accuracy} and \ref{tab:f1}) presented in the main text (\S\ref{sec:experiments-best-layer}).

\item \textbf{Appendix~\ref{sec:appendix-auroc-linear}} reports per-block \emph{linear}-probe AUROC trajectories for all $12$ configurations: Llama (Figure~\ref{fig:auroc-llama-linear}), Mistral (Figure~\ref{fig:auroc-mistral-linear}), Qwen (Figure~\ref{fig:auroc-qwen-linear}), supporting Finding~F2.

\item \textbf{Appendix~\ref{sec:appendix-hyperparams}} consolidates all hyperparameters in Table~\ref{tab:hyper}.
\end{itemize}


\section{Per-Dataset Method Comparisons}
\label{sec:appendix-method-comparison}

\begin{figure}[htbp]
  \centering
  \includegraphics[width=0.75\textwidth]{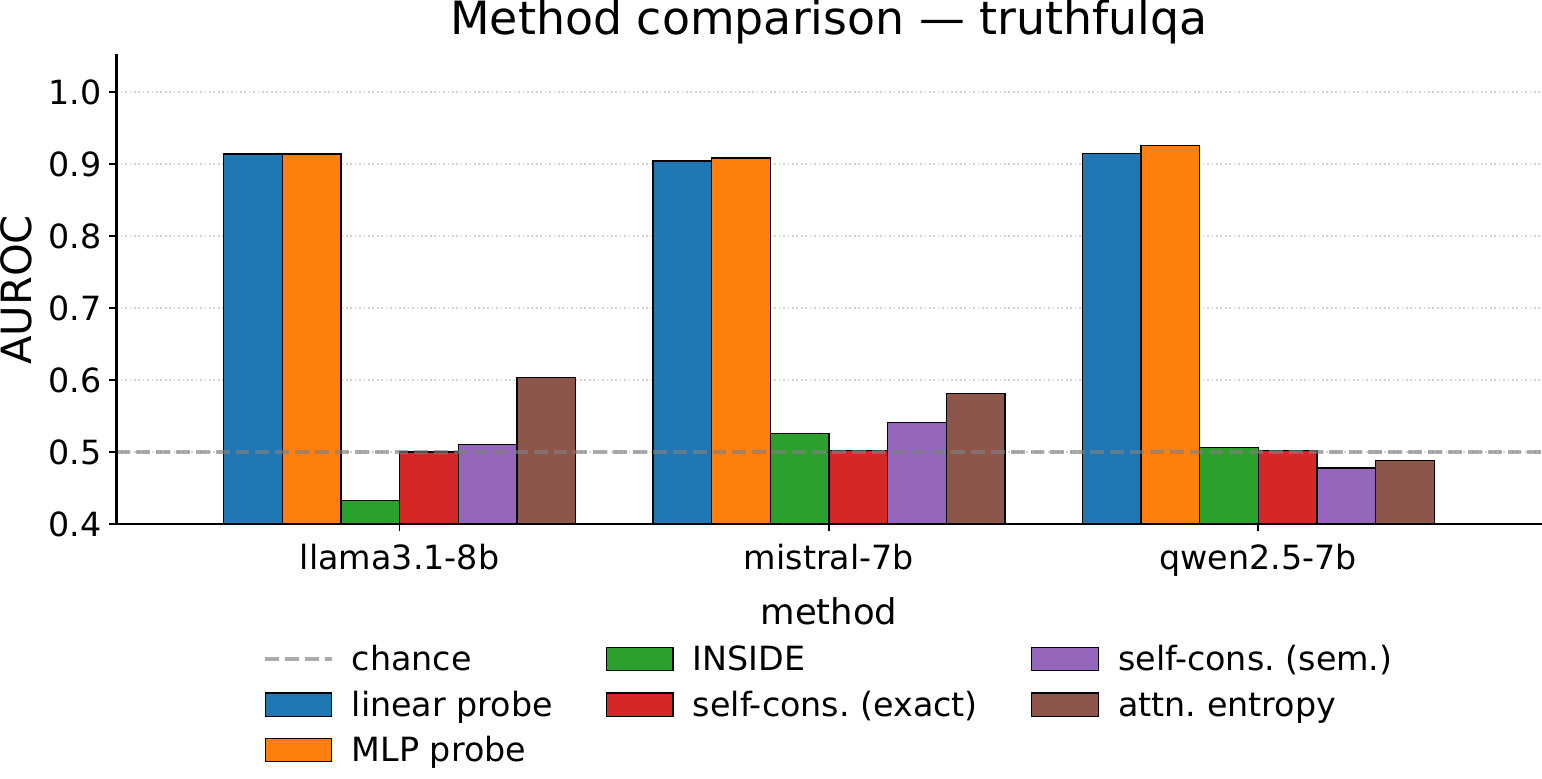}\\[3pt]
  \includegraphics[width=0.75\textwidth]{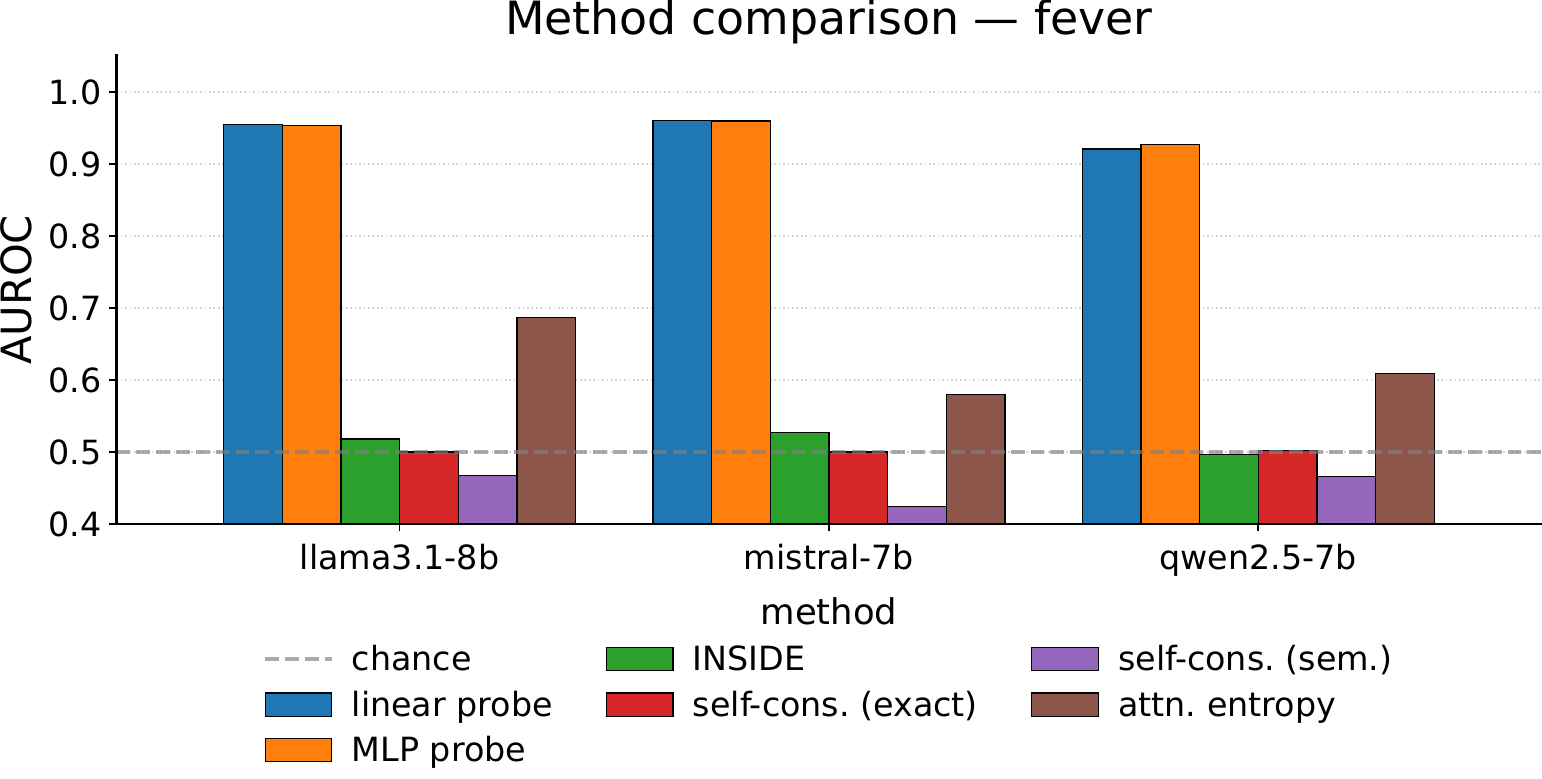}\\[3pt]
  \includegraphics[width=0.75\textwidth]{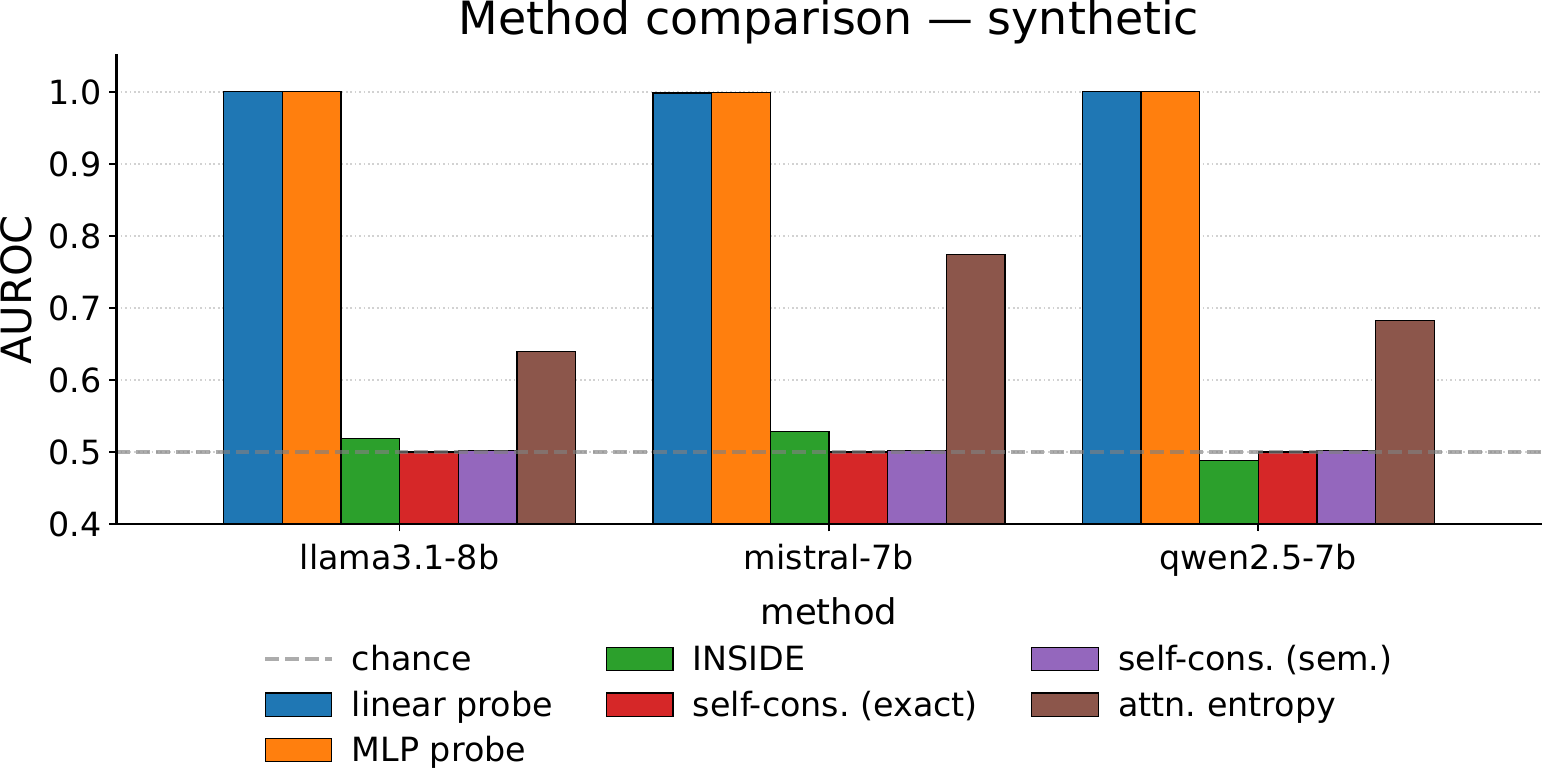}
\caption{Per-dataset method-comparison bar plots for TruthfulQA (top), FEVER (middle), and the synthetic benchmark (bottom). Probes are uniformly at the top, attention entropy is competitive on knowledge-grounded settings, and sampling-based detectors cluster at chance. Cf.\ Figure~\ref{fig:method-comparison} for HaluEval-QA.}
  \label{fig:method-all}
\end{figure}

For completeness, Figure~\ref{fig:method-all} reproduces the method-comparison bar plot for all four datasets. The pattern visible on HaluEval-QA (Figure~\ref{fig:method-comparison}) holds qualitatively: probes are uniformly at the top, attention entropy is a competitive second on knowledge-grounded settings, and the sampling-based detectors cluster at chance.

\section{Cross-Dataset Heatmaps}
\label{sec:appendix-heatmaps}

\begin{figure}[htbp]
  \centering
  \includegraphics[width=0.78\textwidth]{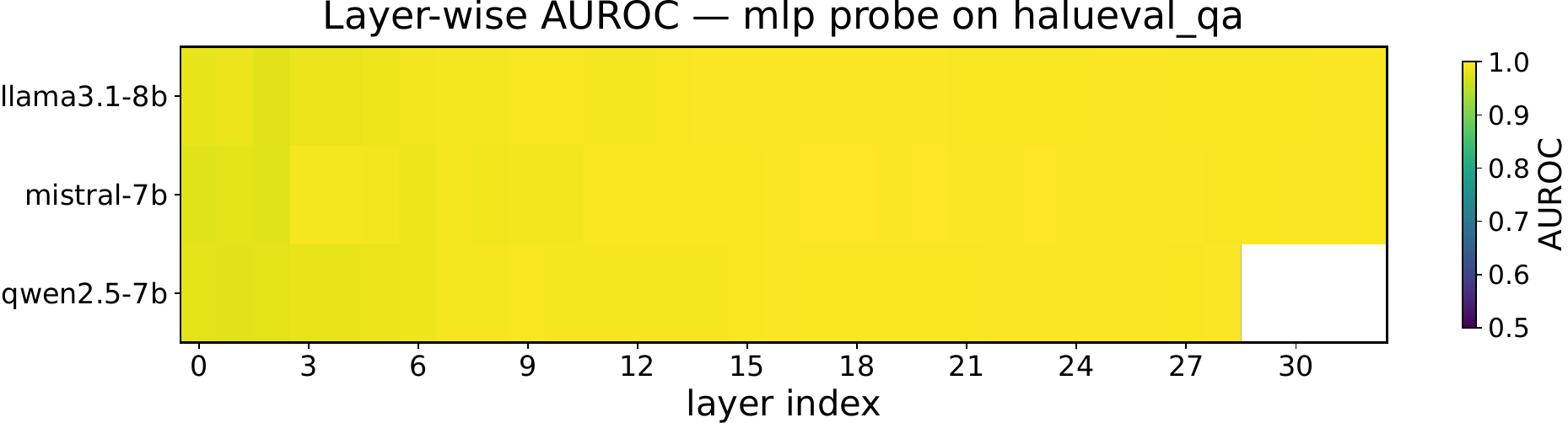}\\[3pt]
  \includegraphics[width=0.78\textwidth]{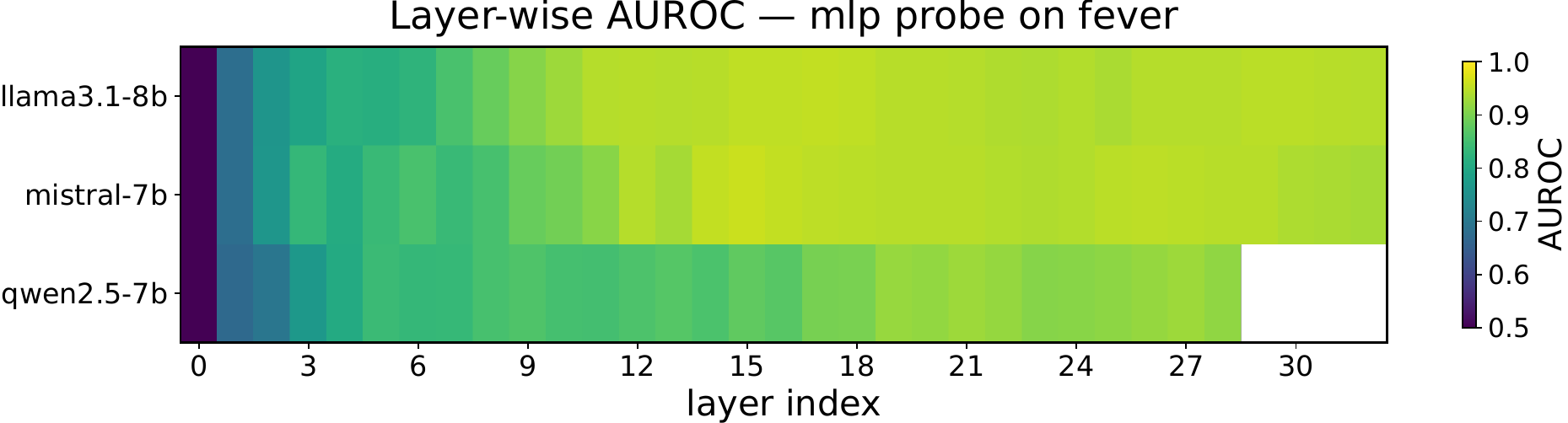}\\[3pt]
  \includegraphics[width=0.78\textwidth]{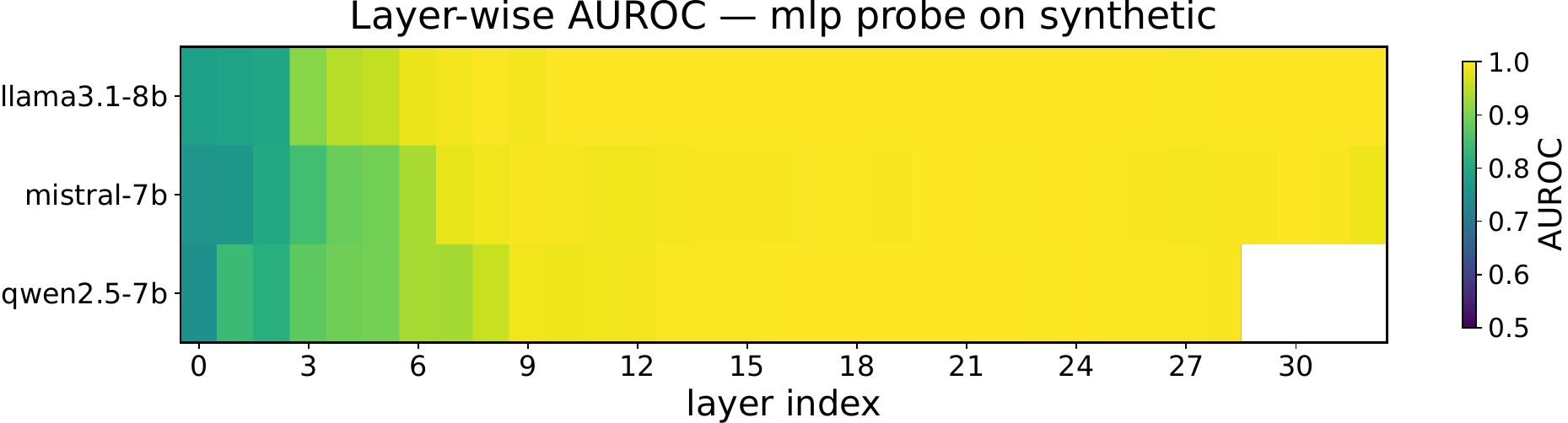}
\caption{MLP-probe per-block AUROC heatmaps on HaluEval-QA (top), FEVER (middle), and the synthetic benchmark (bottom). Rows within each heatmap correspond to the three models; columns are block indices; color encodes AUROC. The bright peak band is consistently in the second half of the network. Cf.\ Figure~\ref{fig:heatmap-truthful} for TruthfulQA.}
  \label{fig:heatmap-all}
\end{figure}

Figure~\ref{fig:heatmap-all} shows the MLP-probe per-block AUROC heatmaps on the remaining three datasets, complementing Figure~\ref{fig:heatmap-truthful} on TruthfulQA. The bright peak band is consistently in the second half of the network for every configuration. The HaluEval-QA peaks are the sharpest, saturating at AUROC $\approx\!0.998$ over a wide mid-to-late plateau, while the synthetic benchmark saturates at AUROC $\approx\!1.000$ in flat regions spanning several late blocks.

\clearpage

\section{Cross-Model Layer-AUROC Trajectories}
\label{sec:appendix-trajectories}

\begin{figure}[htbp]
  \centering
  \includegraphics[width=0.5\textwidth]{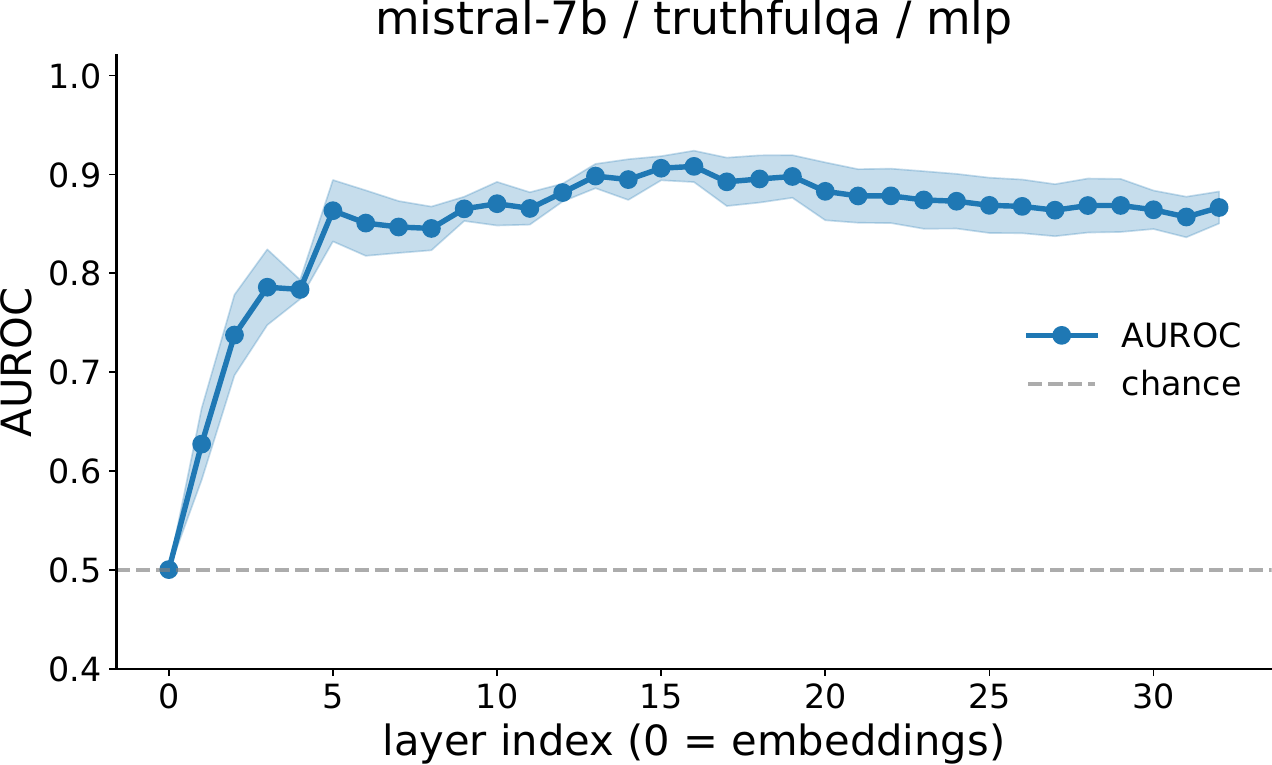}\hfill
  \includegraphics[width=0.5\textwidth]{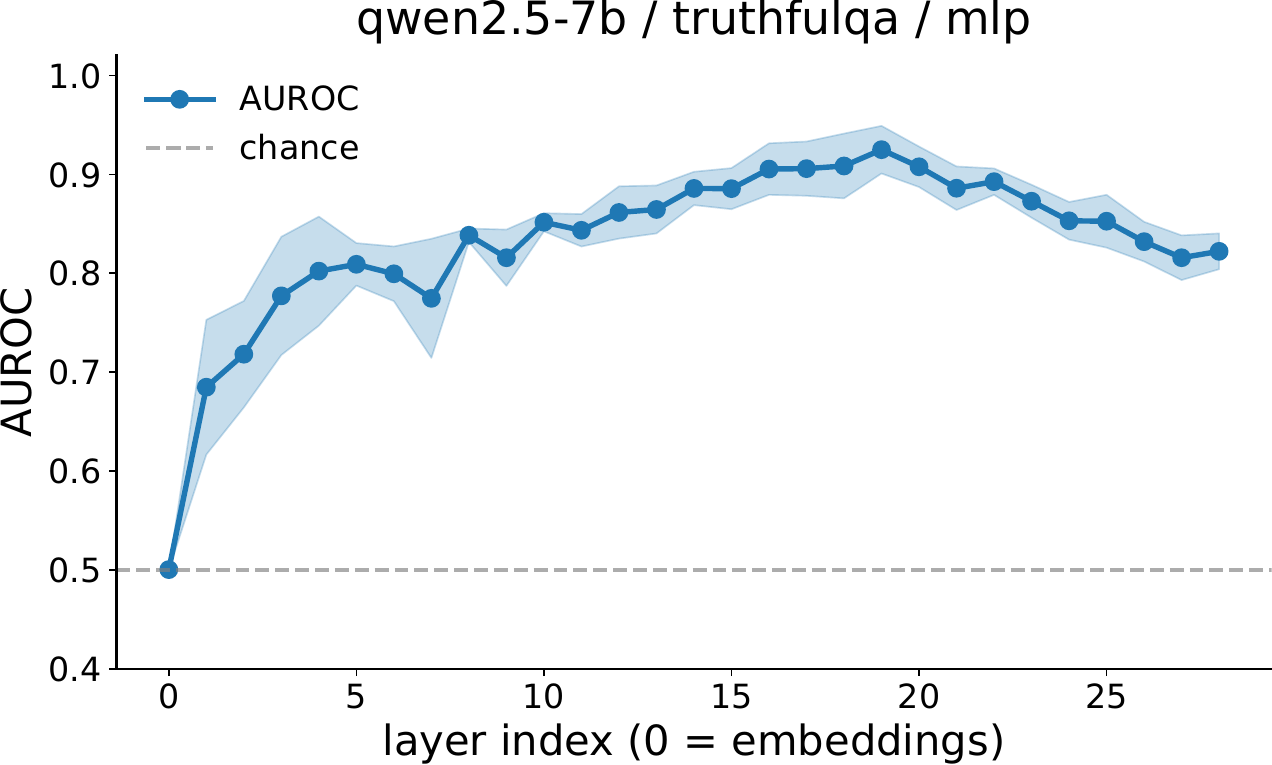}
\caption{Per-block MLP-probe AUROC on TruthfulQA for Mistral-7B (top) and Qwen2.5-7B (bottom). Shaded bands show standard deviation over three seeds. Both models exhibit the same qualitative rise-plateau-decay pattern as Llama-3.1-8B (Figure~\ref{fig:auroc-layer}).}
  \label{fig:auroc-cross}
\end{figure}

Figure~\ref{fig:auroc-cross} reproduces the per-block MLP-probe AUROC curves of Mistral-7B and Qwen2.5-7B on TruthfulQA, complementing Figure~\ref{fig:auroc-layer} for Llama-3.1-8B. The three curves share the same qualitative shape (a monotonic rise, a plateau in the second half of the network, and a small decay near the output), even though the absolute heights and the location of the plateau differ slightly across model families.

\begin{figure}[htbp]
  \centering
  \includegraphics[width=0.78\textwidth]{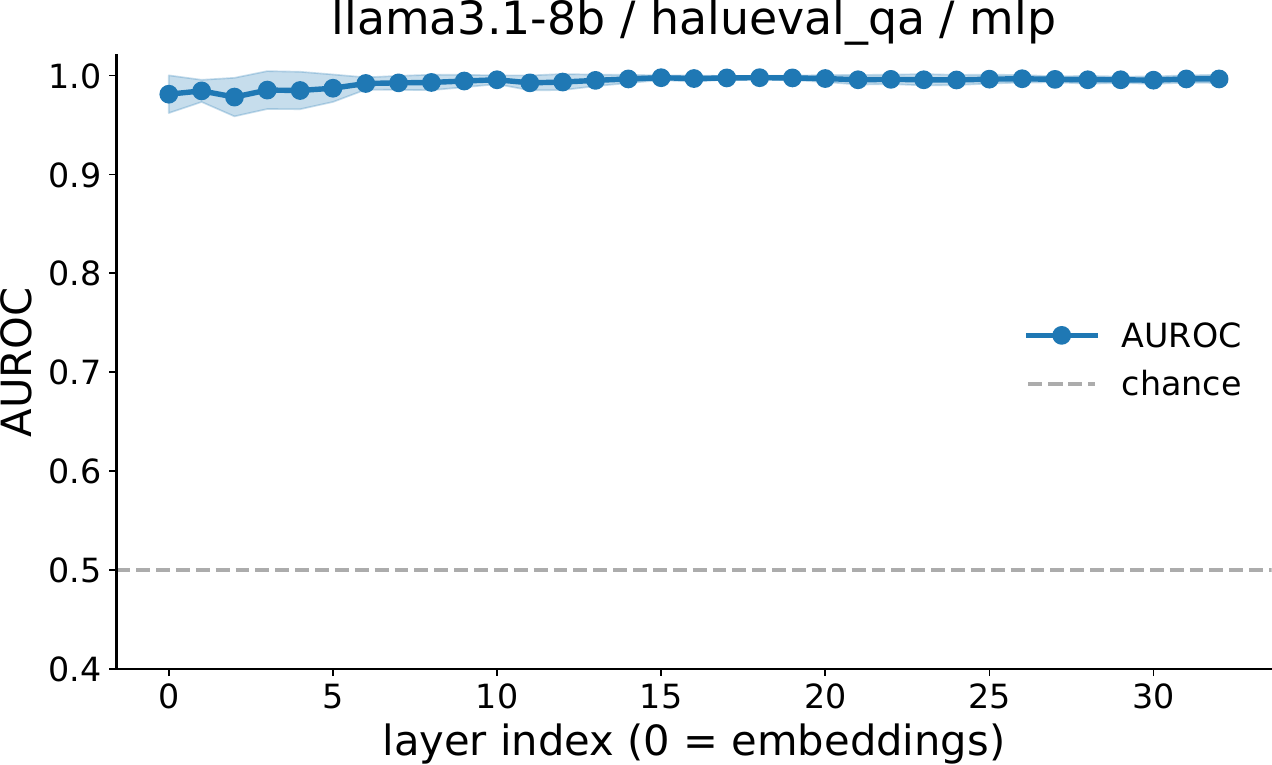}
\caption{Per-block MLP-probe AUROC for Llama-3.1-8B on HaluEval-QA. The plateau saturates at AUROC ${\approx}0.998$ across blocks ${\sim}12$--$24$, with seed standard deviation below $0.005$.}
  \label{fig:auroc-easy}
\end{figure}

For an even cleaner case, Figure~\ref{fig:auroc-easy} shows the Llama-3.1-8B per-block MLP-probe AUROC on HaluEval-QA: the signal saturates at $\approx\!0.998$ over the entire middle of the network, with seed standard deviation below $0.005$.

\section{Cross-Model Class Geometry}
\label{sec:appendix-separation}

\begin{figure}[htbp]
  \centering
  \includegraphics[width=0.78\textwidth]{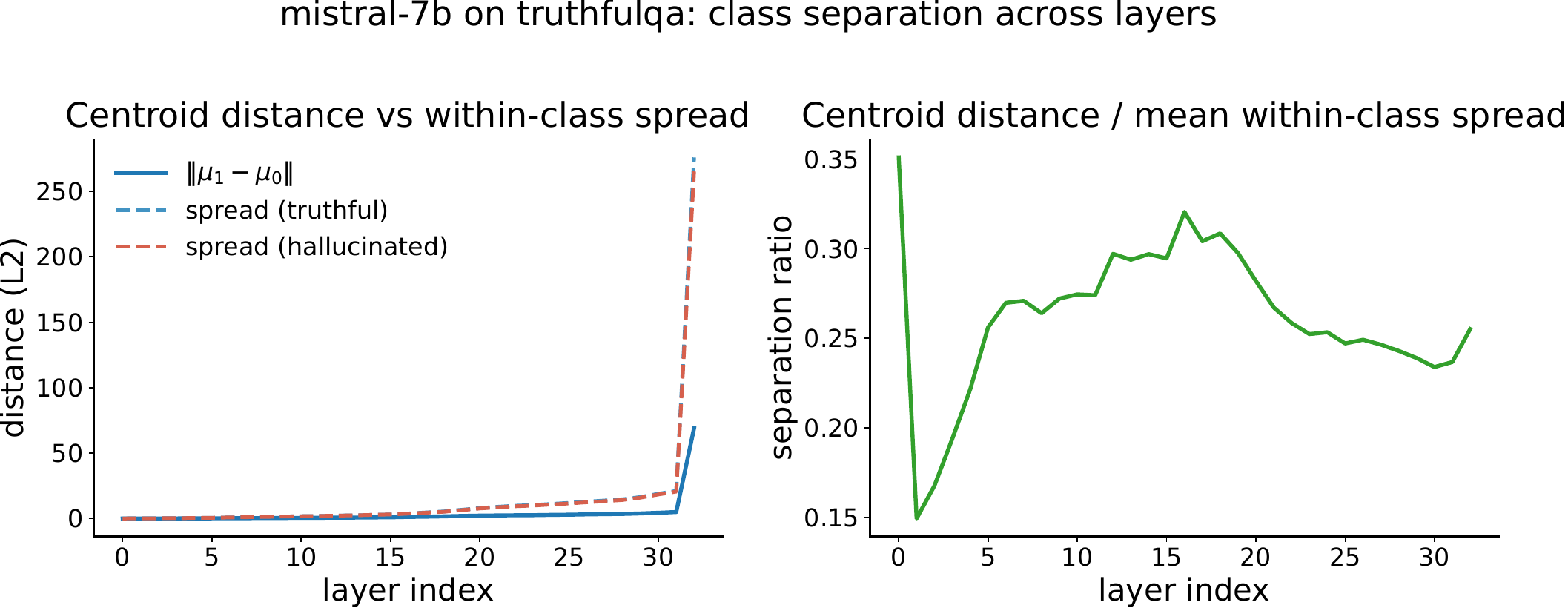}\\[4pt]
  \includegraphics[width=0.78\textwidth]{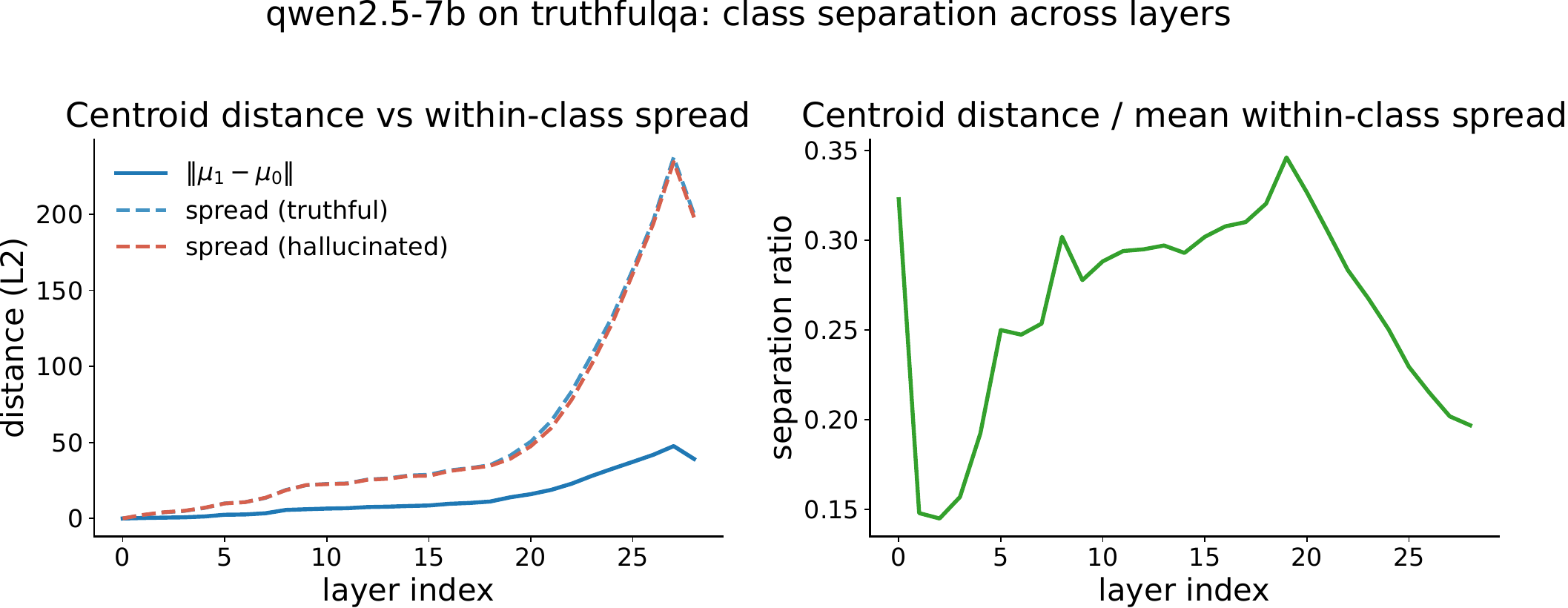}
\caption{Layer-wise class geometry on TruthfulQA for Mistral-7B (top) and Qwen2.5-7B (bottom), showing centroid distance, within-class spread, and separation ratio. The separation ratio rises through the early blocks and saturates in the mid-to-late blocks, mirroring the probe-AUROC trajectories (Figure~\ref{fig:auroc-cross}).}
  \label{fig:sep-cross}
\end{figure}

Figure~\ref{fig:sep-cross} reproduces the class-geometry diagnostic of Figure~\ref{fig:separation} for Mistral-7B and Qwen2.5-7B on TruthfulQA. The separation ratio increases monotonically through the early blocks and plateaus in the second half of the network for both models, consistent with the corresponding probe-AUROC trajectories.

\clearpage

\section{Cross-Model INSIDE Score Distributions}
\label{sec:appendix-inside-dist}

\begin{figure}[htbp]
  \centering
  \includegraphics[width=0.5\textwidth]{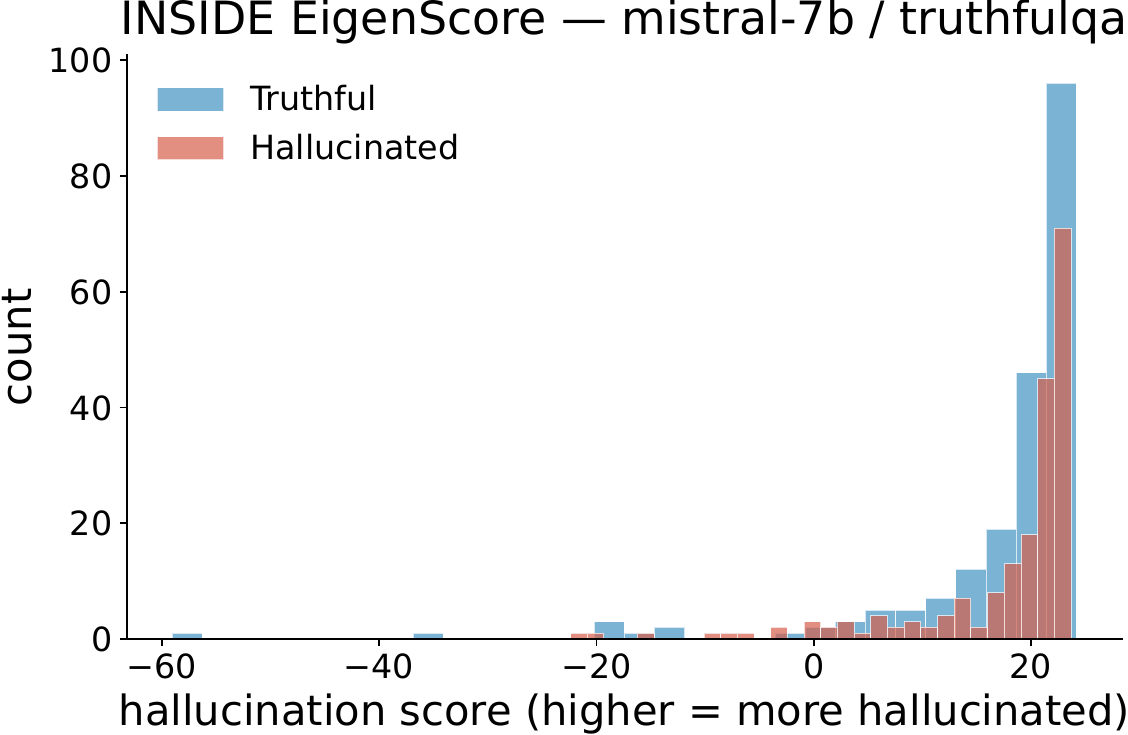}\hfill
  \includegraphics[width=0.5\textwidth]{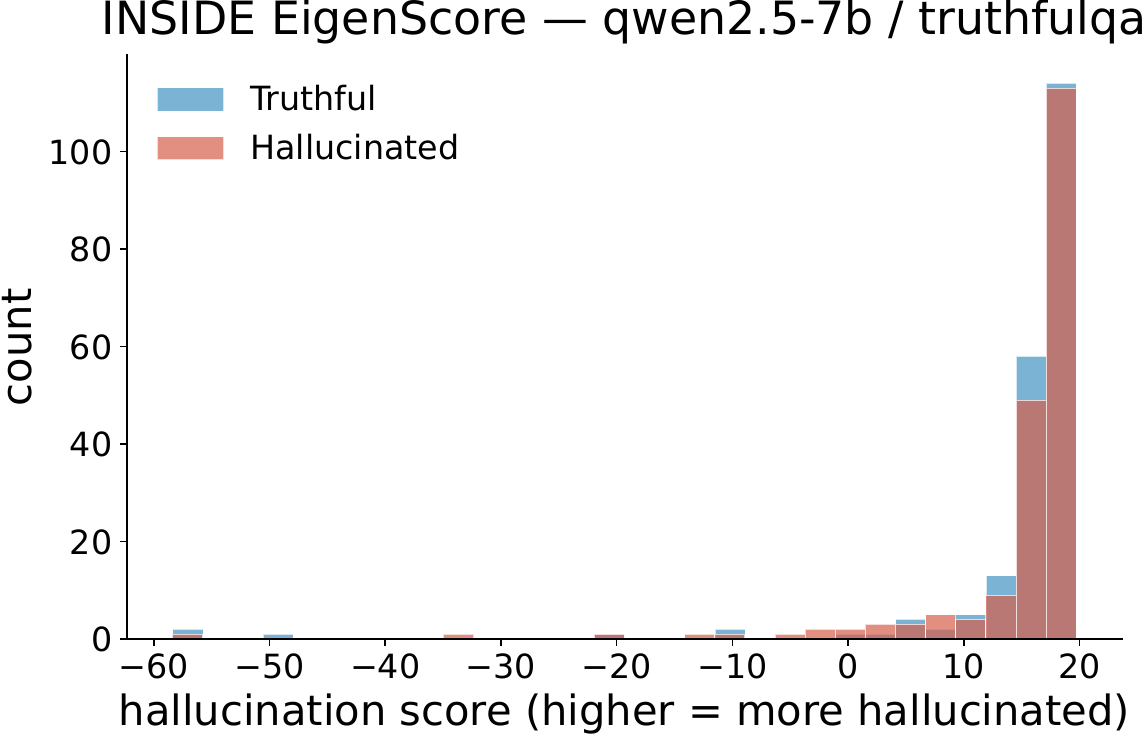}
\caption{INSIDE EigenScore distributions on TruthfulQA for Mistral-7B (top) and Qwen2.5-7B (bottom), separated by ground-truth label. The near-complete overlap of the truthful and hallucinated histograms explains the chance-level AUROC reported in Table~\ref{tab:headline}.}
  \label{fig:inside-cross}
\end{figure}

Figure~\ref{fig:inside-cross} reproduces the per-class INSIDE EigenScore distribution on TruthfulQA for Mistral-7B and Qwen2.5-7B, complementing Figure~\ref{fig:inside-dist}. The near-complete overlap of the truthful and hallucinated distributions persists across all three model families, consistent with the chance-level AUROC.

\section{Cross-Model Attention Entropy on HaluEval-QA}
\label{sec:appendix-attn-cross}

\begin{figure}[htbp]
  \centering
  \includegraphics[width=0.9\textwidth]{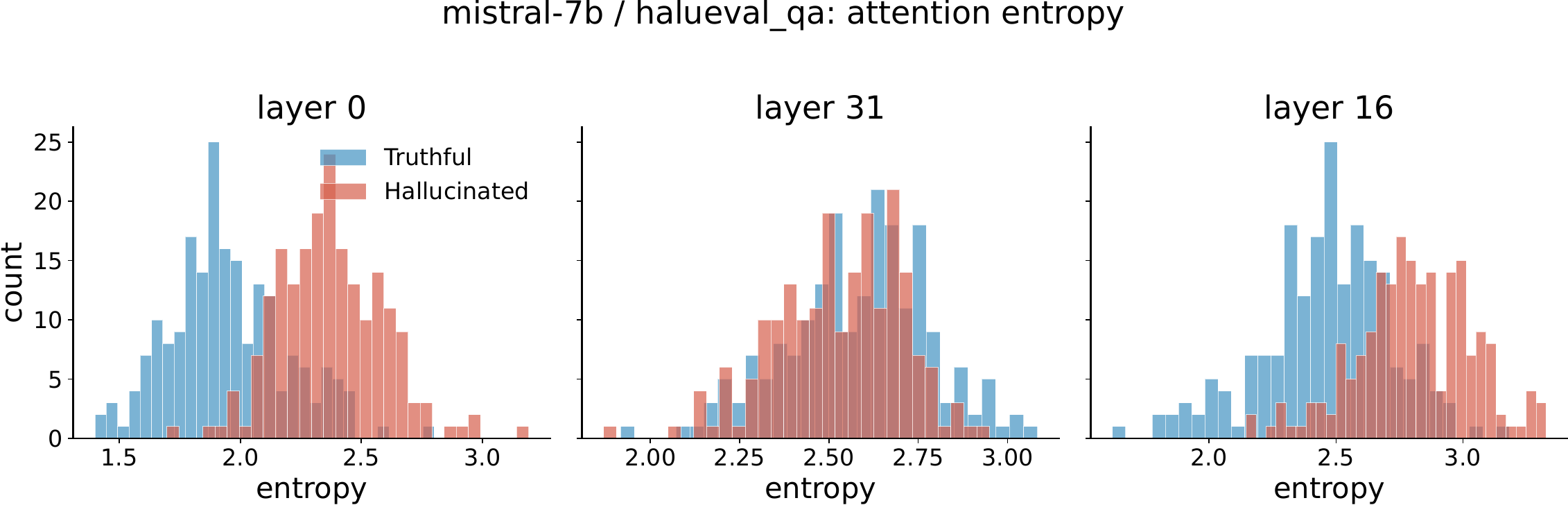}\\[4pt]
  \includegraphics[width=0.9\textwidth]{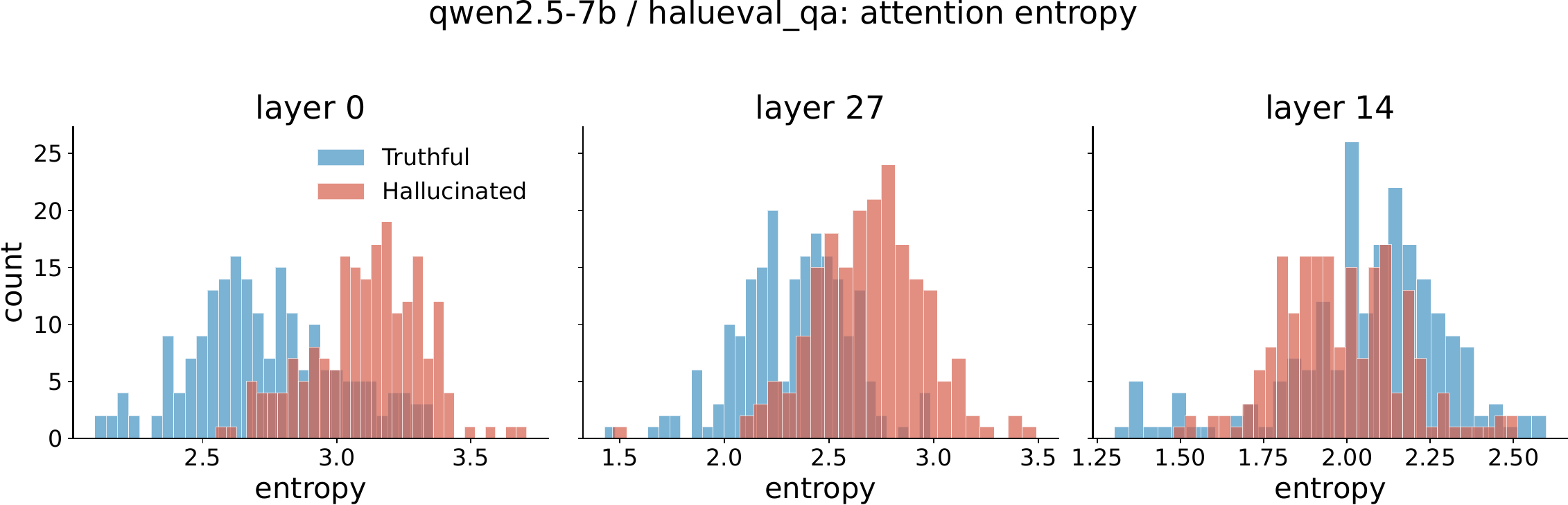}
\caption{Last-token attention entropy on HaluEval-QA for Mistral-7B (top) and Qwen2.5-7B (bottom) at the first, middle, and last captured transformer blocks. The first-block separation (AUROC $0.902$ and $0.866$ respectively) confirms the pattern observed for Llama-3.1-8B in Figure~\ref{fig:attention}.}
  \label{fig:attn-cross}
\end{figure}

Figure~\ref{fig:attn-cross} shows the head-averaged last-token attention-entropy distributions at three transformer blocks for Mistral-7B and Qwen2.5-7B on HaluEval-QA. The first-block separation observed for Llama (Figure~\ref{fig:attention}) is preserved in both models (AUROC $0.902$ and $0.866$ respectively), confirming that first-block attention entropy provides a useful signal at no additional inference cost in the knowledge-conditioned setting.

\section{2-D Projection (PCA) at the Probing Peak}
\label{sec:appendix-pca}

\begin{figure}[htbp]
  \centering
  \includegraphics[width=0.48\textwidth]{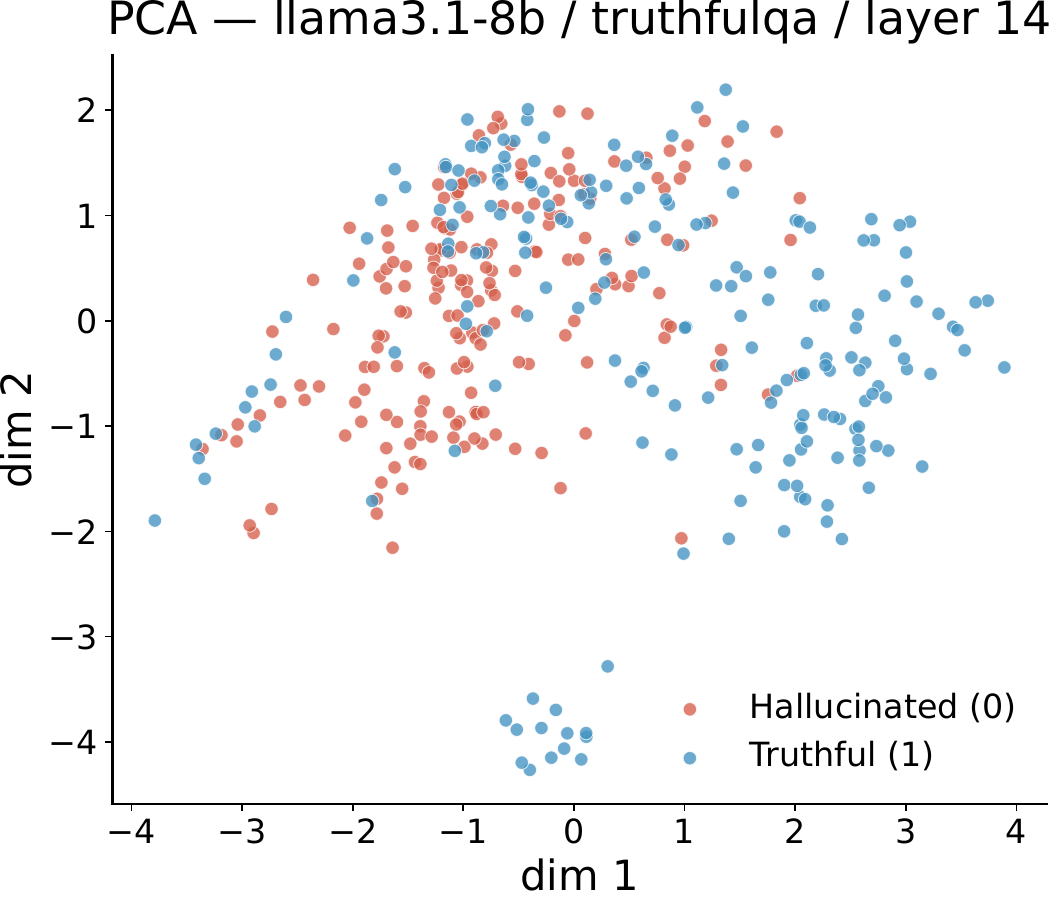}\hfill
  \includegraphics[width=0.48\textwidth]{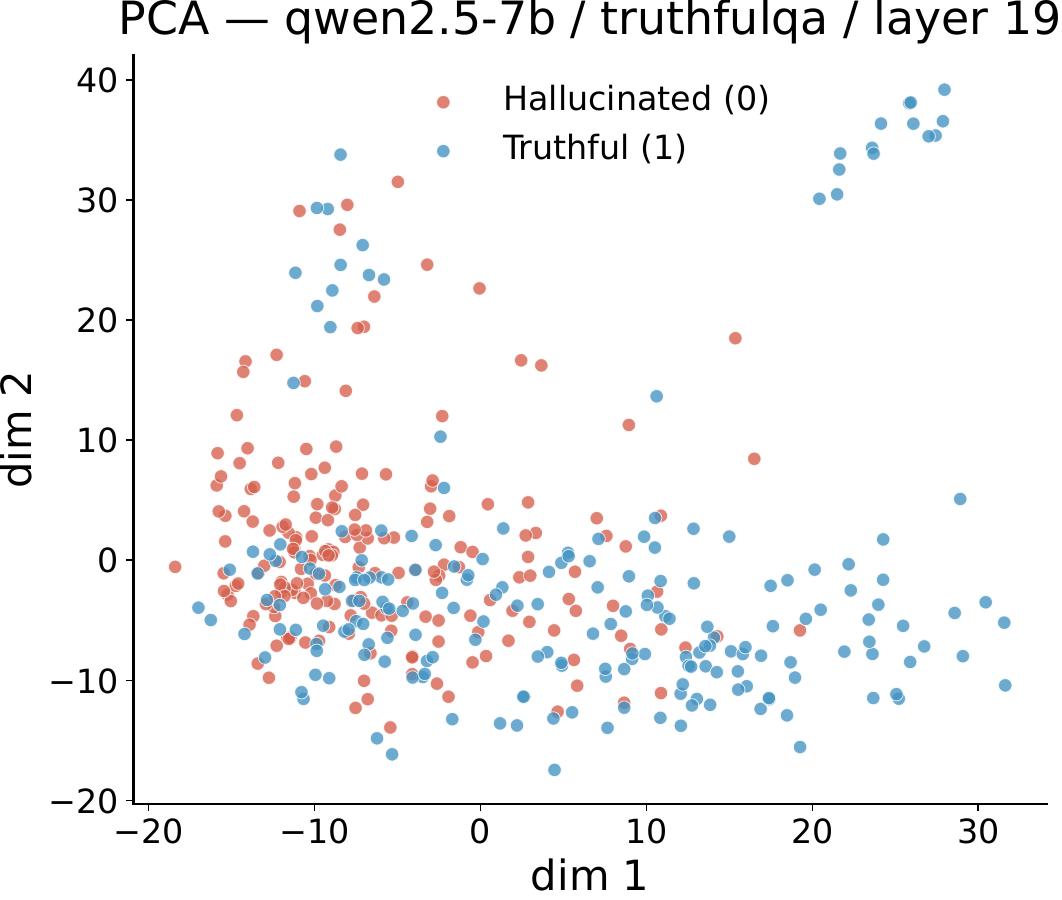}
\caption{2-D PCA projections on TruthfulQA at each model's probing peak. \emph{Left}: Llama-3.1-8B at block~$14$. \emph{Right}: Qwen2.5-7B at block~$19$. Both models show clear class separation along the leading principal components, consistent with the high linear-probe AUROC. Cf.\ Figure~\ref{fig:tsne} for the non-linear t-SNE counterpart.}
  \label{fig:pca}
  \label{fig:pca-qwen}
\end{figure}

Figure~\ref{fig:pca} shows side-by-side PCA projections of the hidden states at the probing peak for Llama-3.1-8B (block~$14$) and Qwen2.5-7B (block~$19$) on TruthfulQA. Unlike t-SNE, PCA is a linear projection and therefore mirrors what the linear probe sees. Both models show clear class separation along the leading principal directions, consistent with the $0.914$/$0.915$ linear-probe AUROC at these blocks.

\section{Cross-Model t-SNE at the Probing Peak (Qwen)}
\label{sec:appendix-tsne-qwen}

\begin{figure}[htbp]
  \centering
  \includegraphics[width=0.6\textwidth]{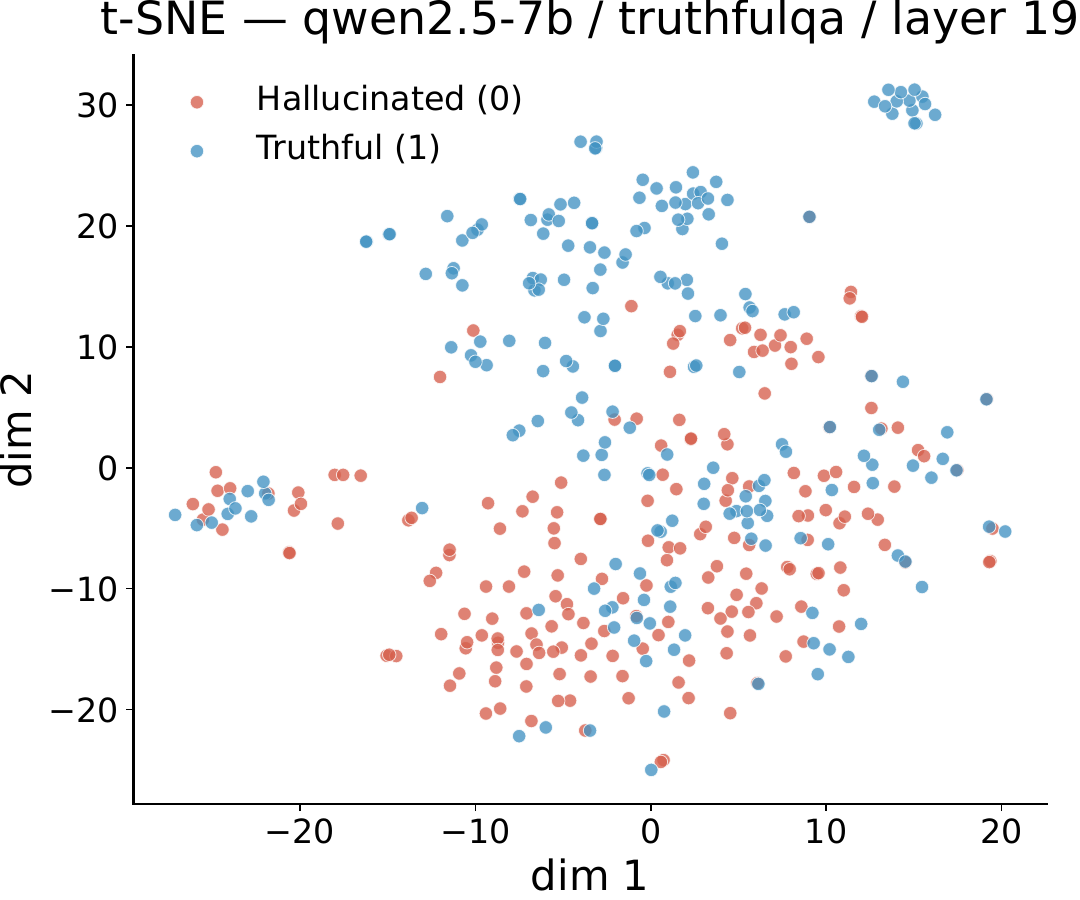}
\caption{t-SNE projection of the layer-$19$ hidden states of Qwen2.5-7B on TruthfulQA (Qwen probing peak, perplexity~$30$). The two classes form distinct clusters with a thin margin region, confirming the geometric separation that underlies the $0.925$ MLP-probe AUROC.}
  \label{fig:tsne-qwen}
\end{figure}

Figure~\ref{fig:tsne-qwen} shows the t-SNE projection of the layer-$19$ hidden states of Qwen2.5-7B on TruthfulQA, the block at which the Qwen MLP probe achieves its peak AUROC of $0.925$ (Table~\ref{tab:headline}). As for Llama-3.1 at its probing peak (Figure~\ref{fig:tsne}), the two classes form distinct clusters with a narrow margin region. The t-SNE visualization uses perplexity $30$ (see Appendix~\ref{sec:appendix-hyperparams}) and confirms that the linear separability observed by the probe corresponds to a genuine geometric separation in hidden-state space, not an artifact of the linear classifier.

\section{Synthetic Benchmark Construction}
\label{sec:appendix-synthetic}

The synthetic benchmark is generated locally from four small knowledge banks: world capitals ($30$ countries), chemical element symbols ($20$ elements), literary authorship ($20$ works), and planetary orbits ($8$ planets). For each underlying fact we emit two items, one truthful (label $1$) and one hallucinated (label $0$), sharing the same prompt and differing only in the substantive token of the answer (e.g.\ \texttt{The capital of France is}~$\to$~\texttt{Paris} vs.\ \texttt{Berlin}). Items are shuffled with a fixed seed before the $400$-item cap is applied; the resulting class balance is exactly $200{:}200$. This yields the simplest binary signal in our suite, corresponding to the $\geq\!0.998$ probing AUROC plateau in Table~\ref{tab:headline}.


\section{Linear-Probe Cross-Dataset Heatmaps}
\label{sec:appendix-heatmaps-linear}

\begin{figure}[htbp]
  \centering
  \includegraphics[width=0.49\textwidth]{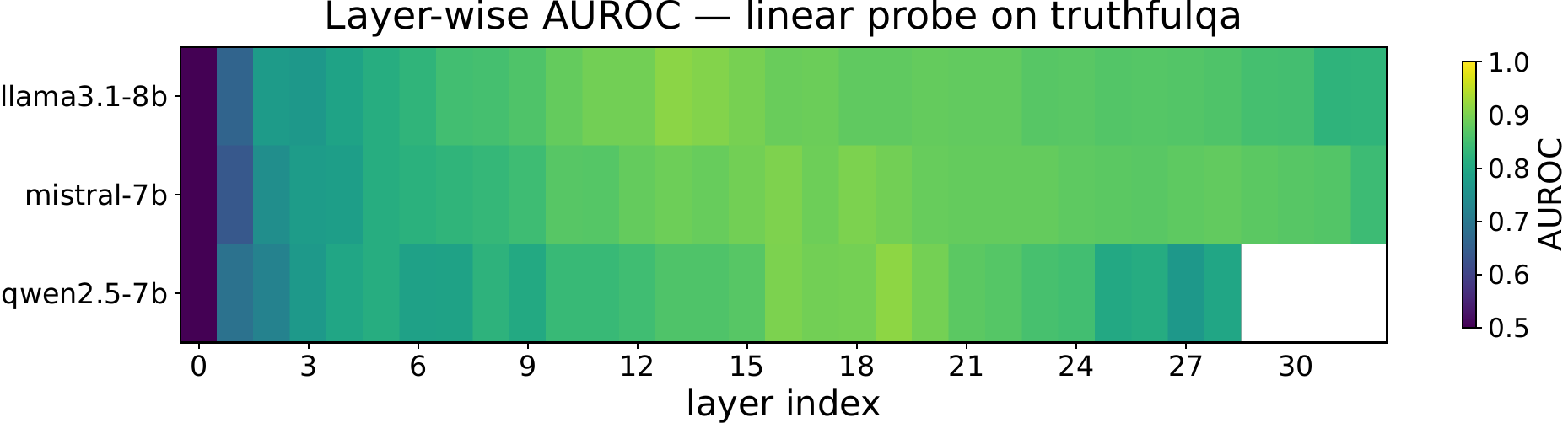}\hfill
  \includegraphics[width=0.49\textwidth]{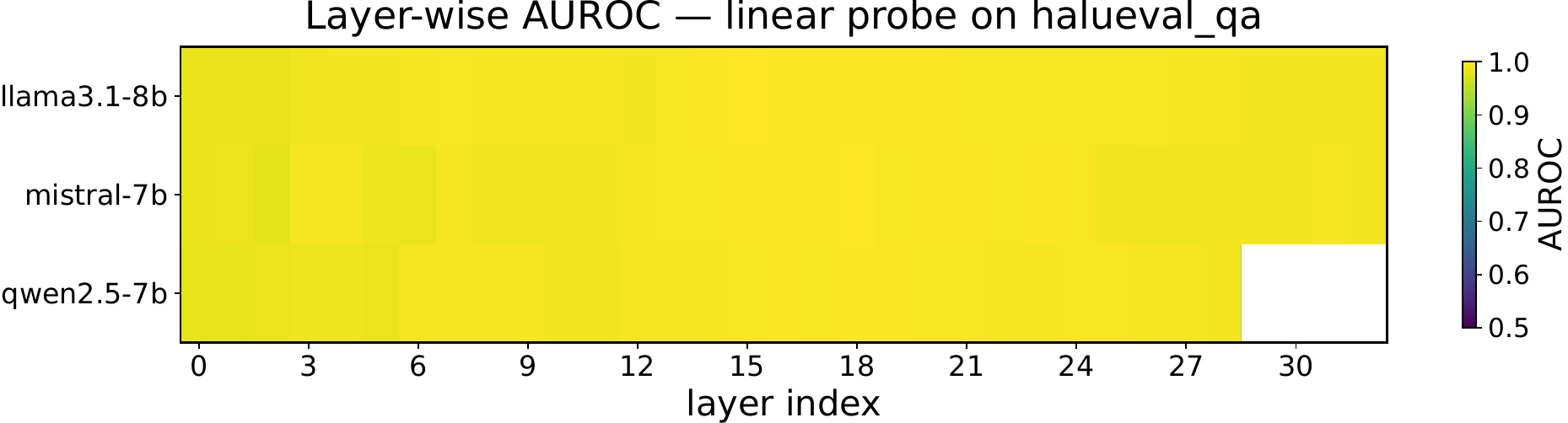}\\[2pt]
  \includegraphics[width=0.49\textwidth]{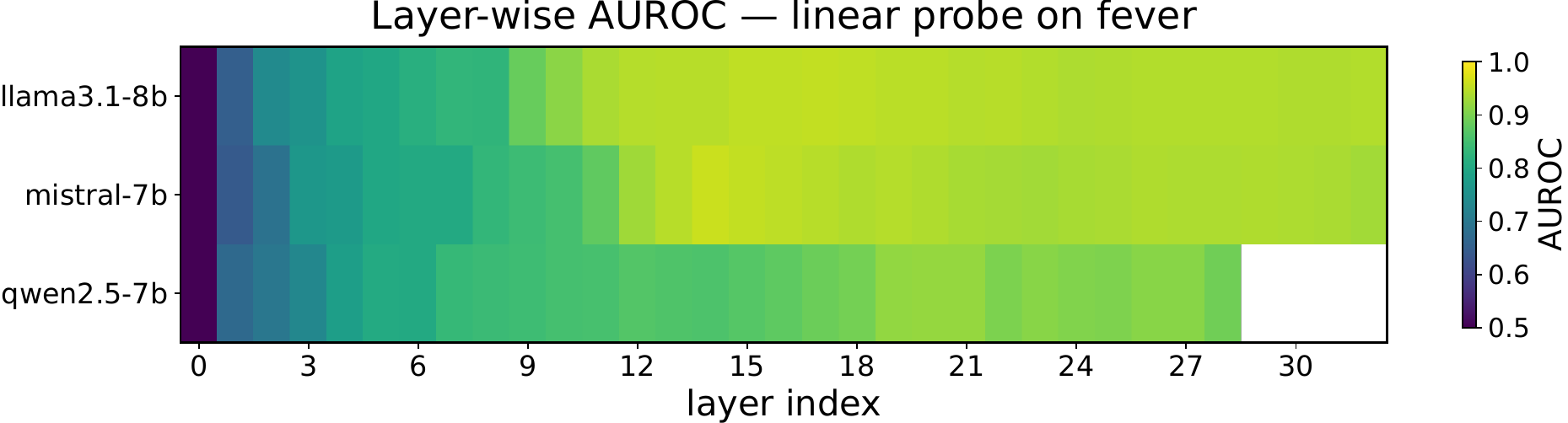}\hfill
  \includegraphics[width=0.49\textwidth]{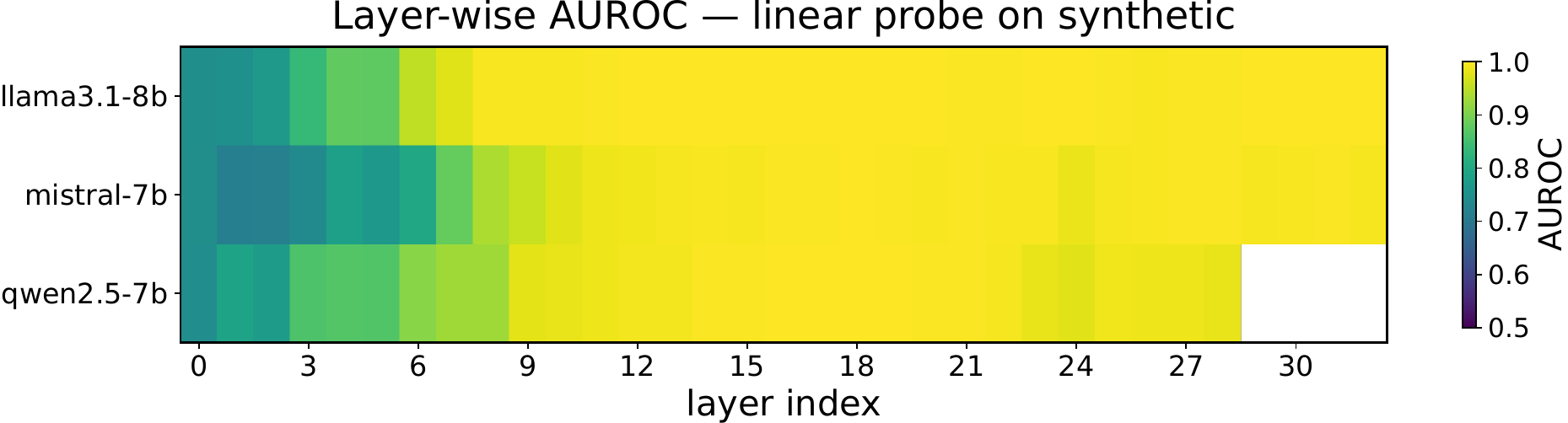}
\caption{Per-block AUROC heatmaps for the \emph{linear} SAPLMA probe across all four datasets. \emph{Top-left}: TruthfulQA. \emph{Top-right}: HaluEval-QA. \emph{Bottom-left}: FEVER. \emph{Bottom-right}: synthetic. Rows within each heatmap are the three models; columns are block indices; color is AUROC. The pattern is qualitatively identical to the MLP counterparts (Figures~\ref{fig:heatmap-truthful},~\ref{fig:heatmap-all}).}
  \label{fig:heatmap-linear-all}
\end{figure}

Figure~\ref{fig:heatmap-linear-all} reproduces the per-block AUROC heatmaps of the \textbf{linear} SAPLMA probe across all four datasets, complementing the MLP heatmaps in Figure~\ref{fig:heatmap-truthful} and Figure~\ref{fig:heatmap-all}. The linear and MLP heatmaps are visually almost indistinguishable on every dataset, reflecting the near-equality of the two probe variants (F2 in \S\ref{sec:experiments-best-layer}, $|\Delta\mathrm{AUROC}|\!\leq\!0.01$ in all $12$ cells). This provides further evidence that the truthfulness signal in hidden-state space is effectively linear.

\section{Layer-Wise AUROC Trajectories Across All Cells}
\label{sec:appendix-auroc-all}

\begin{figure}[htbp]
  \centering
  \includegraphics[width=0.49\textwidth]{auroc_llama3.1-8b__truthfulqa__mlp.pdf}\hfill
  \includegraphics[width=0.49\textwidth]{auroc_llama3.1-8b__halueval_qa__mlp.pdf}\\[2pt]
  \includegraphics[width=0.49\textwidth]{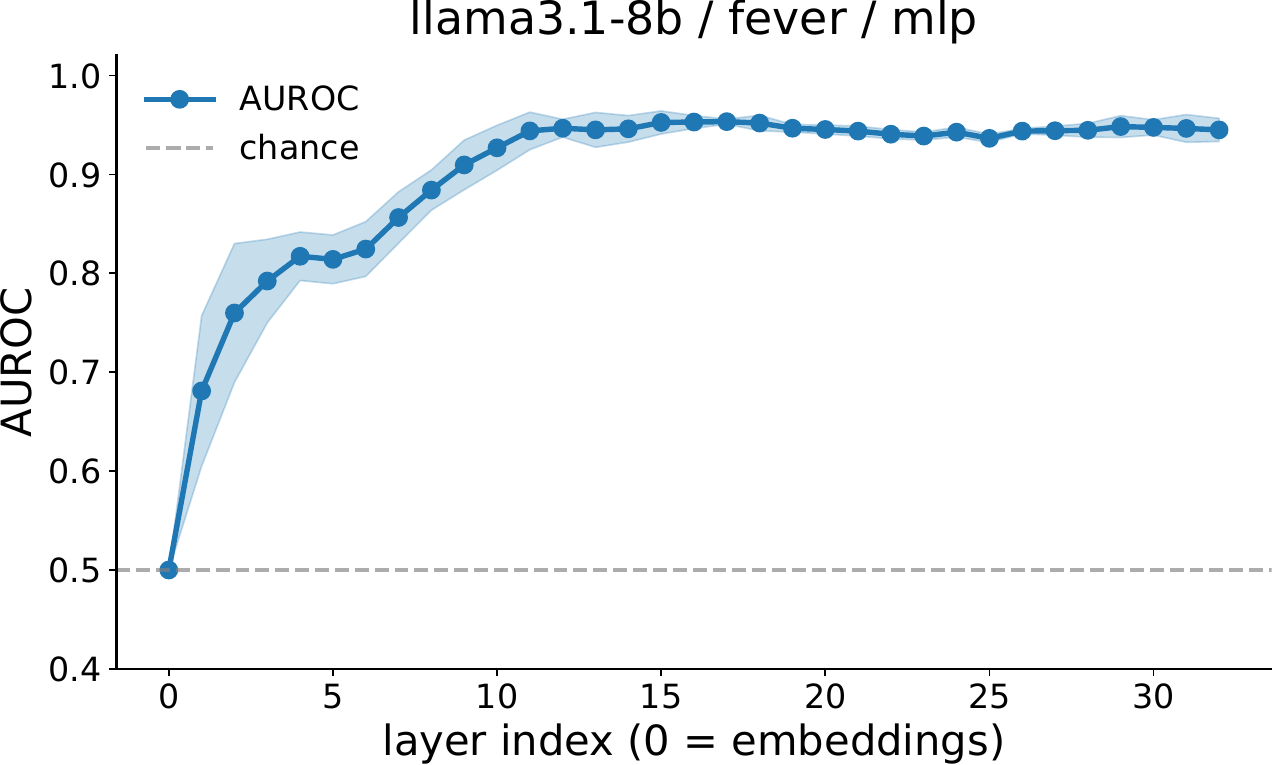}\hfill
  \includegraphics[width=0.49\textwidth]{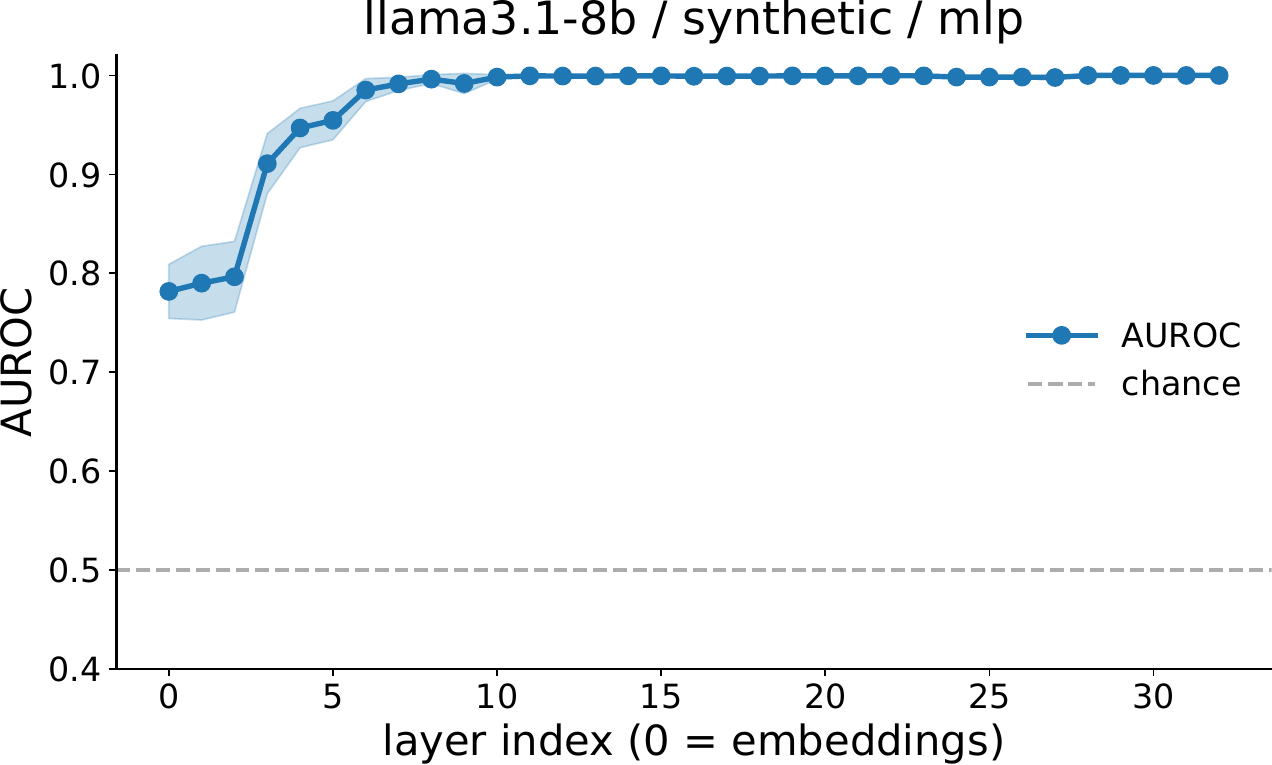}
\caption{Per-block MLP-probe AUROC for Llama-3.1-8B. \emph{Top-left}: TruthfulQA. \emph{Top-right}: HaluEval-QA. \emph{Bottom-left}: FEVER. \emph{Bottom-right}: synthetic. Shaded band: standard deviation over three seeds.}
  \label{fig:auroc-llama-all}
\end{figure}

\begin{figure}[htbp]
  \centering
  \includegraphics[width=0.49\textwidth]{auroc_mistral-7b__truthfulqa__mlp.pdf}\hfill
  \includegraphics[width=0.49\textwidth]{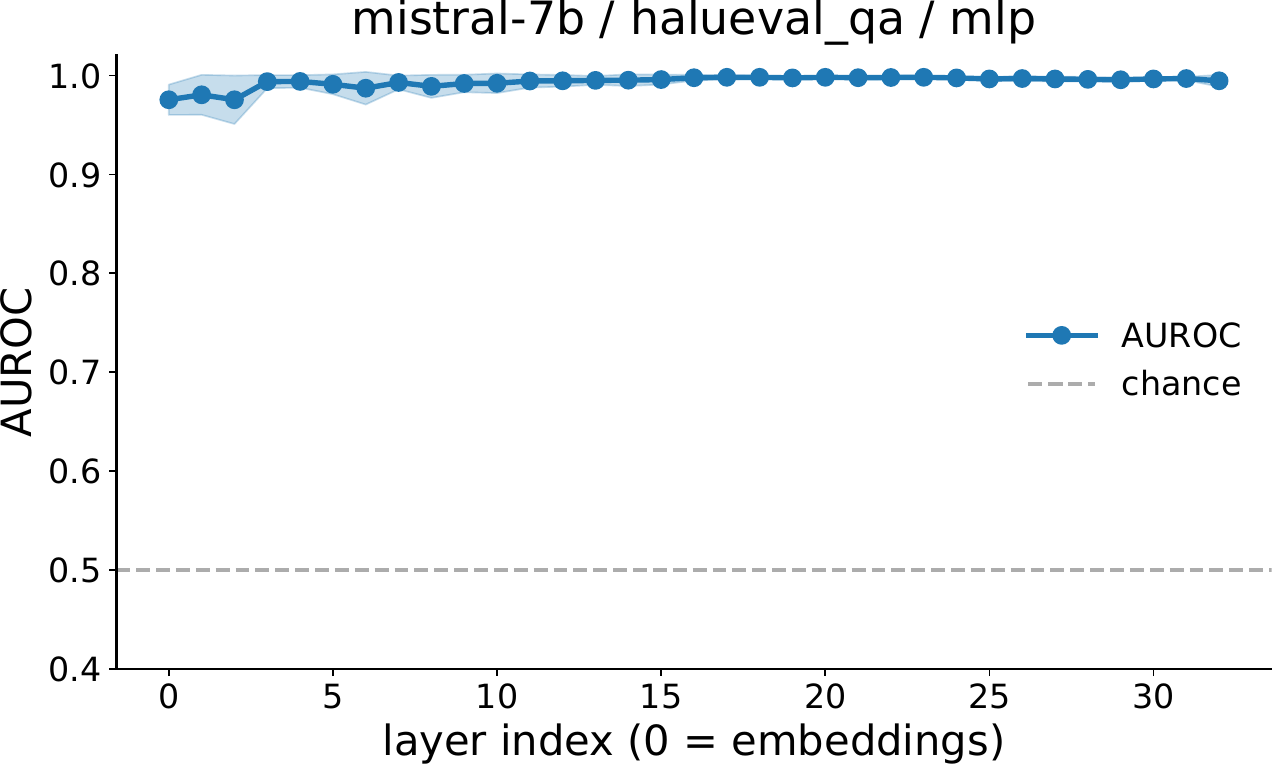}\\[2pt]
  \includegraphics[width=0.49\textwidth]{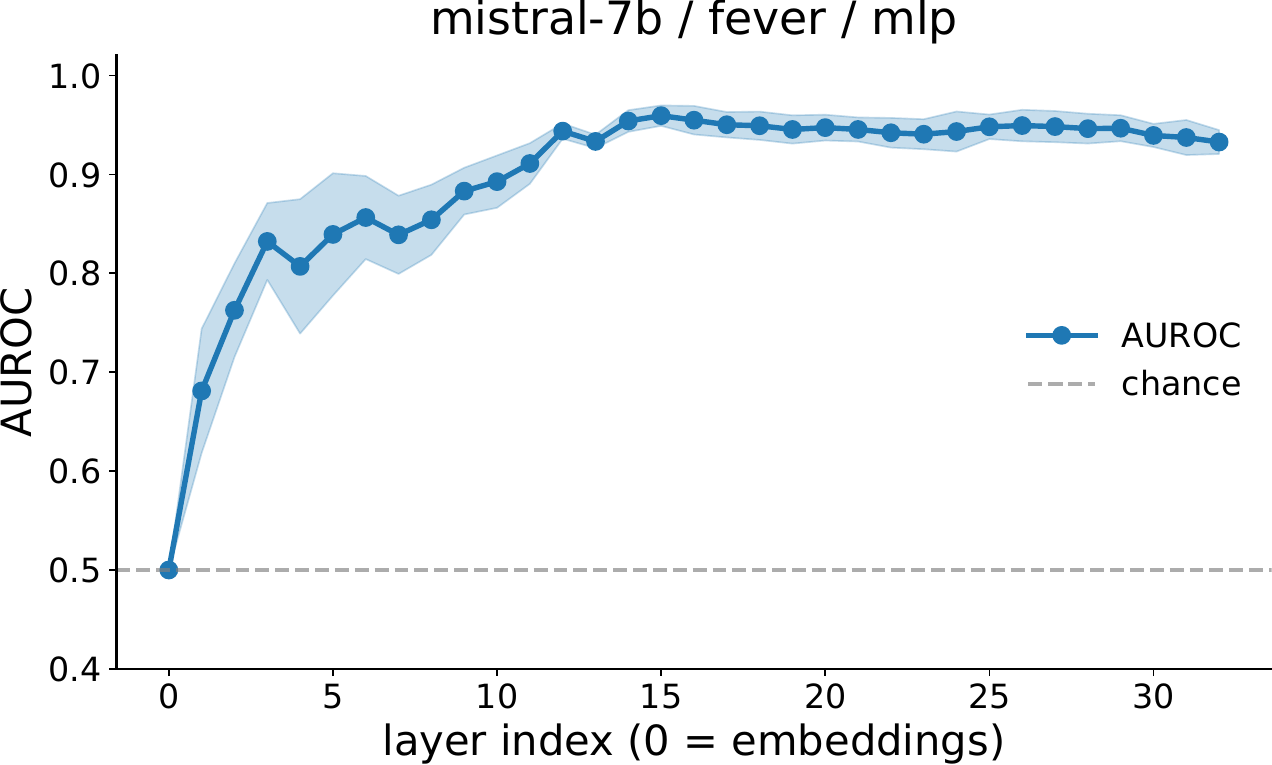}\hfill
  \includegraphics[width=0.49\textwidth]{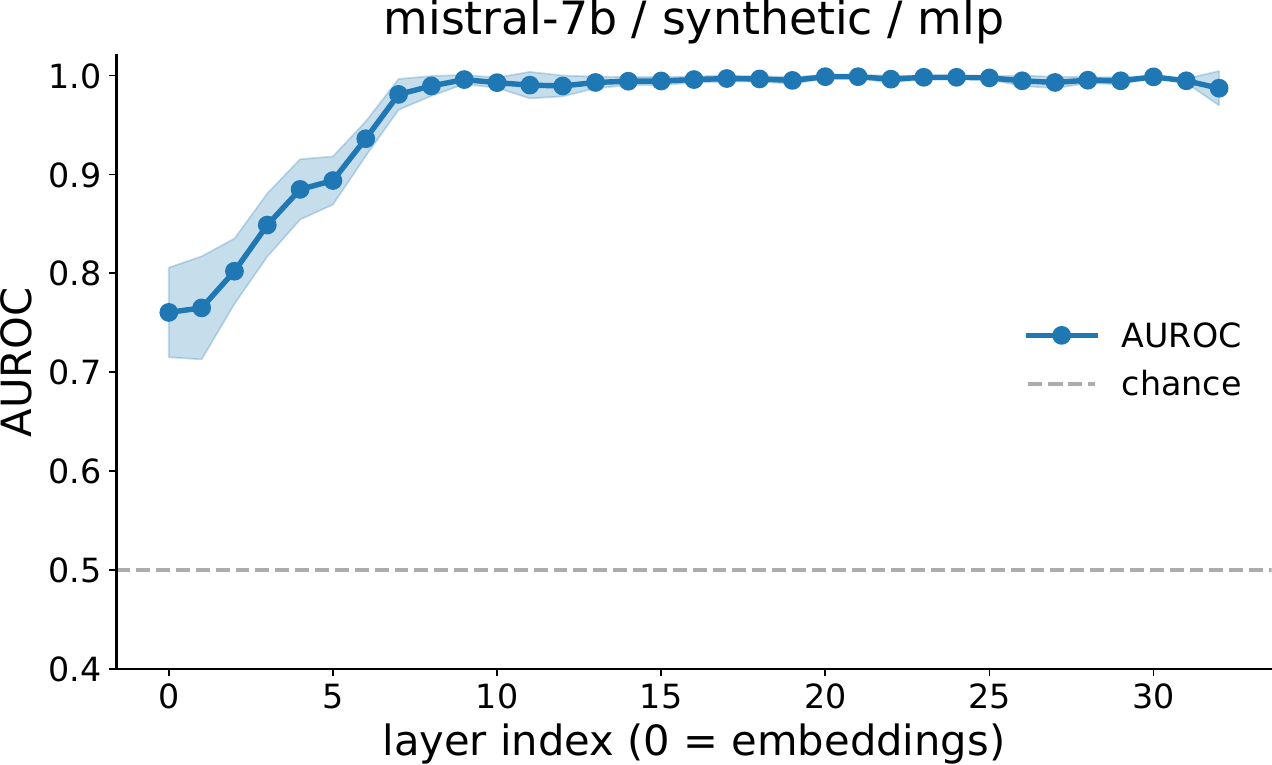}
\caption{Per-block MLP-probe AUROC for Mistral-7B-Instruct-v0.3. \emph{Top-left}: TruthfulQA. \emph{Top-right}: HaluEval-QA. \emph{Bottom-left}: FEVER. \emph{Bottom-right}: synthetic. Shaded band: standard deviation over three seeds.}
  \label{fig:auroc-mistral-all}
\end{figure}

\begin{figure}[htbp]
  \centering
  \includegraphics[width=0.49\textwidth]{auroc_qwen2.5-7b__truthfulqa__mlp.pdf}\hfill
  \includegraphics[width=0.49\textwidth]{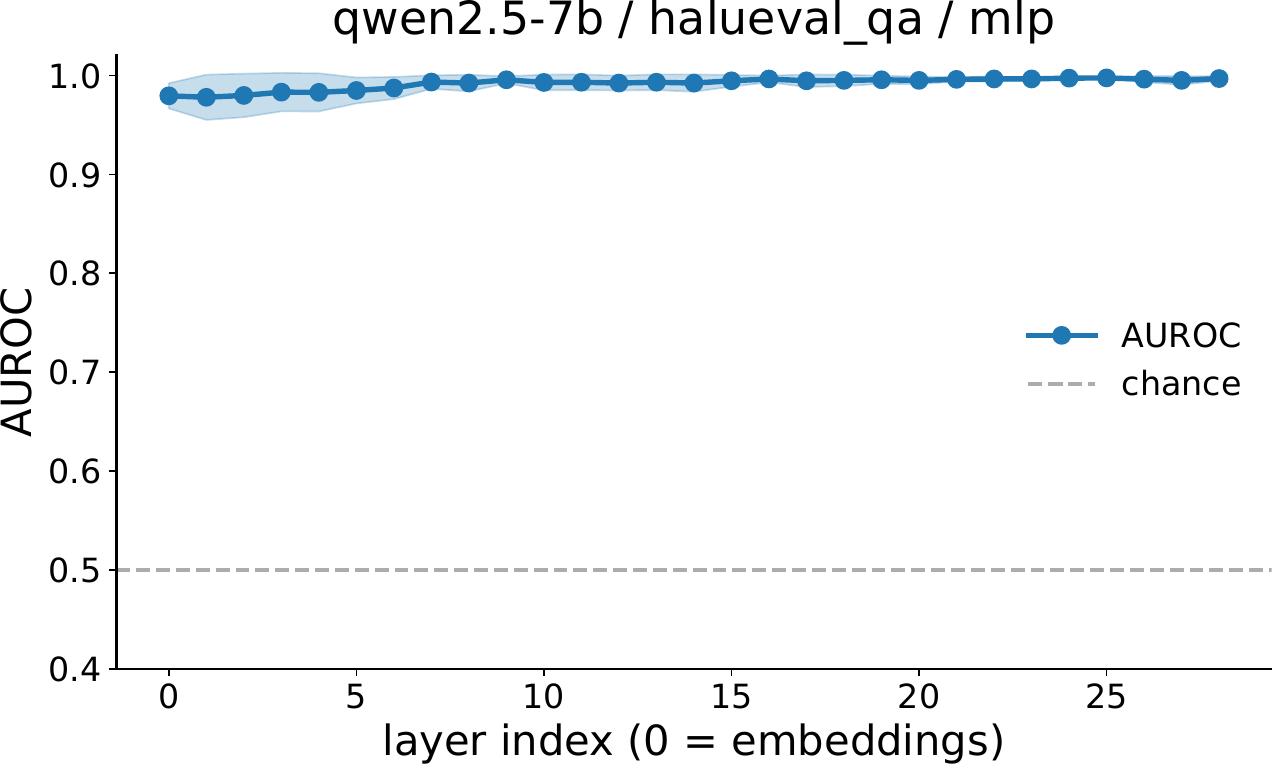}\\[2pt]
  \includegraphics[width=0.49\textwidth]{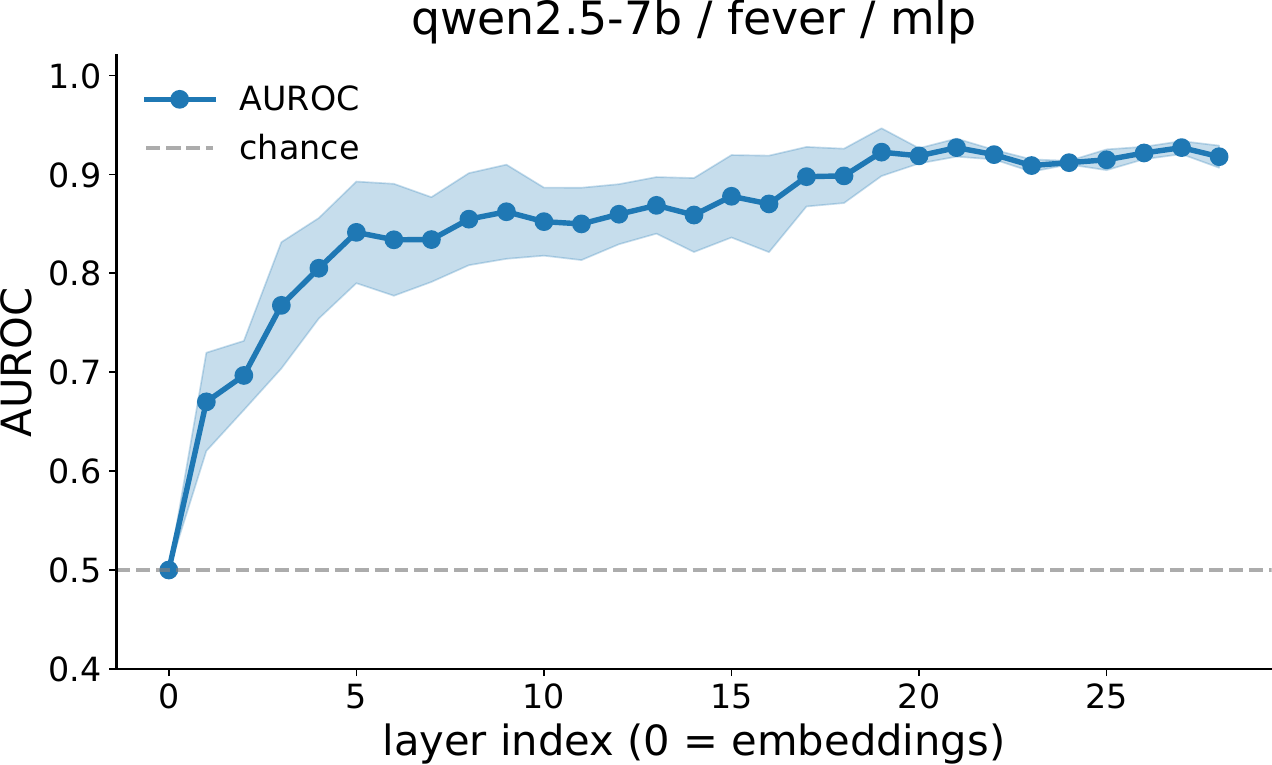}\hfill
  \includegraphics[width=0.49\textwidth]{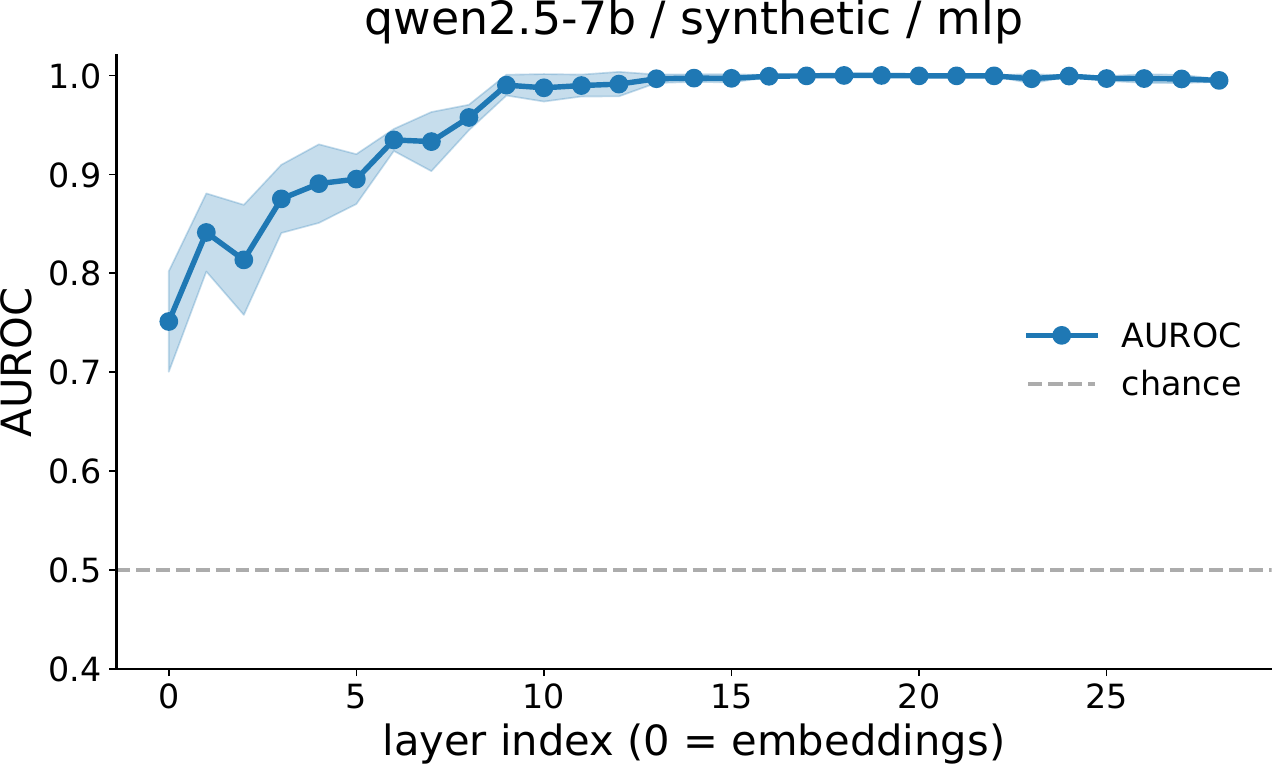}
\caption{Per-block MLP-probe AUROC for Qwen2.5-7B-Instruct ($28$ transformer blocks). \emph{Top-left}: TruthfulQA. \emph{Top-right}: HaluEval-QA. \emph{Bottom-left}: FEVER. \emph{Bottom-right}: synthetic. Shaded band: standard deviation over three seeds.}
  \label{fig:auroc-qwen-all}
\end{figure}

Figures~\ref{fig:auroc-llama-all}--\ref{fig:auroc-qwen-all} report the per-block MLP-probe AUROC curves for all $12$ model--dataset combinations. Within each model the qualitative shape is preserved across datasets: a monotonic rise through the early blocks, a high-AUROC plateau in the middle-to-late blocks, and a small decay at the output. The location and height of the plateau vary modestly across datasets (TruthfulQA peaks lower at $\approx\!0.91$--$0.93$, while HaluEval-QA, FEVER and synthetic peak above $0.95$ and often saturate). The linear-probe counterparts are visually indistinguishable from the MLP curves shown here (see Appendix~\ref{sec:appendix-auroc-linear} for the complete set of linear-probe layer-AUROC trajectories across all $12$ cells), and the equality of linear and MLP curves is exactly the F2 finding of \S\ref{sec:experiments-best-layer}.

\section{Class Geometry Across All Cells}
\label{sec:appendix-sep-all}

\begin{figure}[htbp]
  \centering
  \includegraphics[width=1\textwidth]{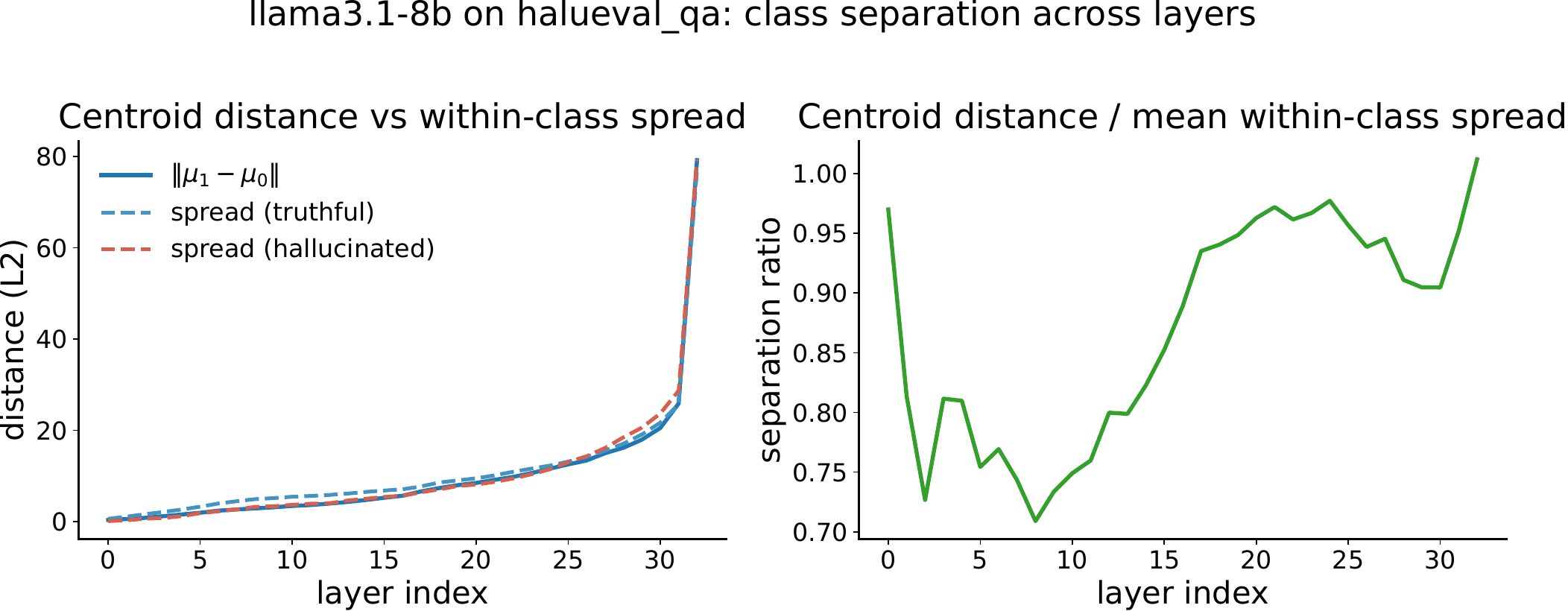}\\[3pt]
  \includegraphics[width=1\textwidth]{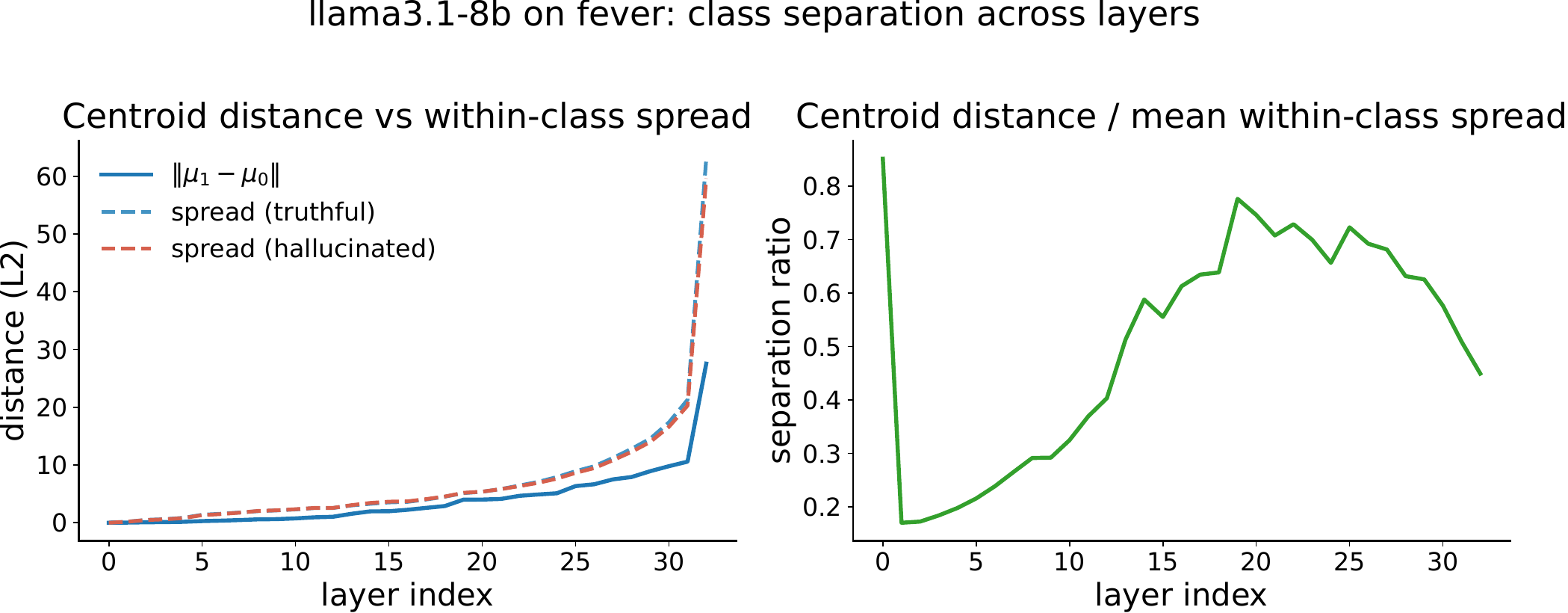}\\[3pt]
  \includegraphics[width=1\textwidth]{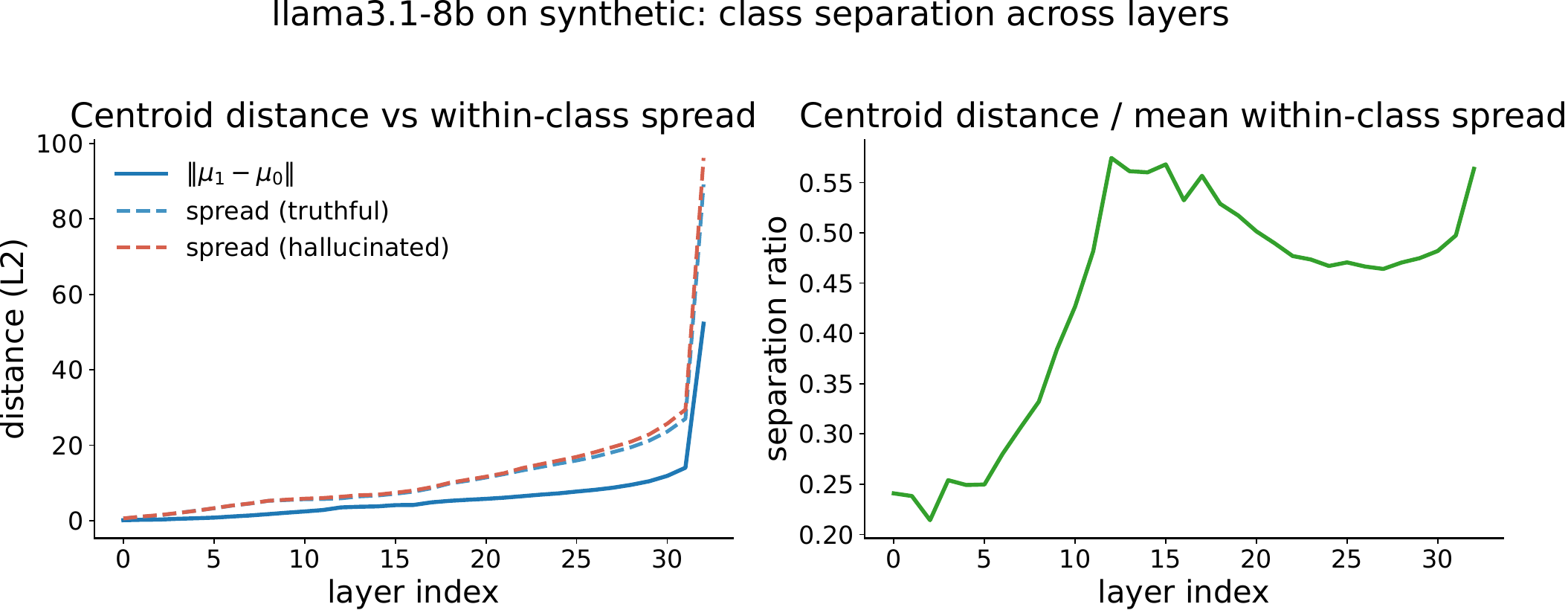}
\caption{Layer-wise class geometry for Llama-3.1-8B on HaluEval-QA (top), FEVER (middle), and the synthetic benchmark (bottom), showing centroid distance, within-class spread, and separation ratio. The separation ratio plateaus in mid-to-late blocks across all three datasets. Cf.\ Figure~\ref{fig:separation} for TruthfulQA.}
  \label{fig:sep-llama-all}
\end{figure}

\begin{figure}[htbp]
  \centering
  \includegraphics[width=1\textwidth]{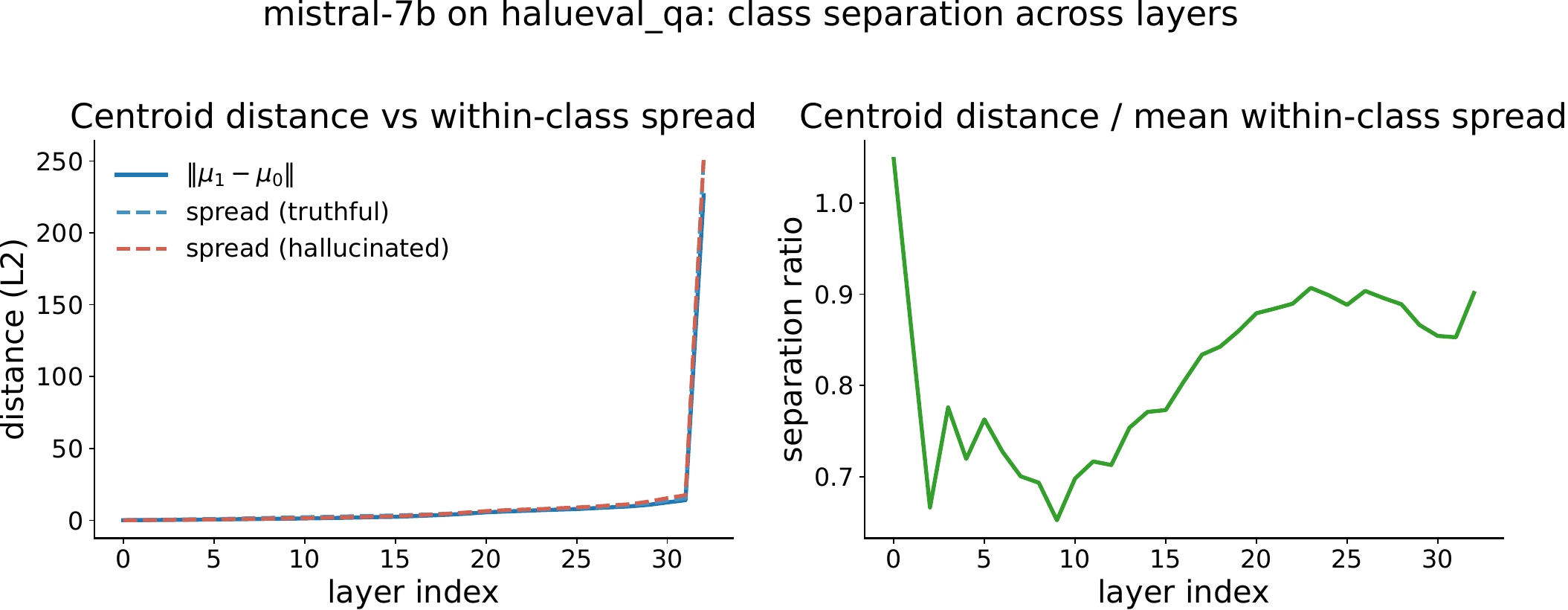}\\[3pt]
  \includegraphics[width=1\textwidth]{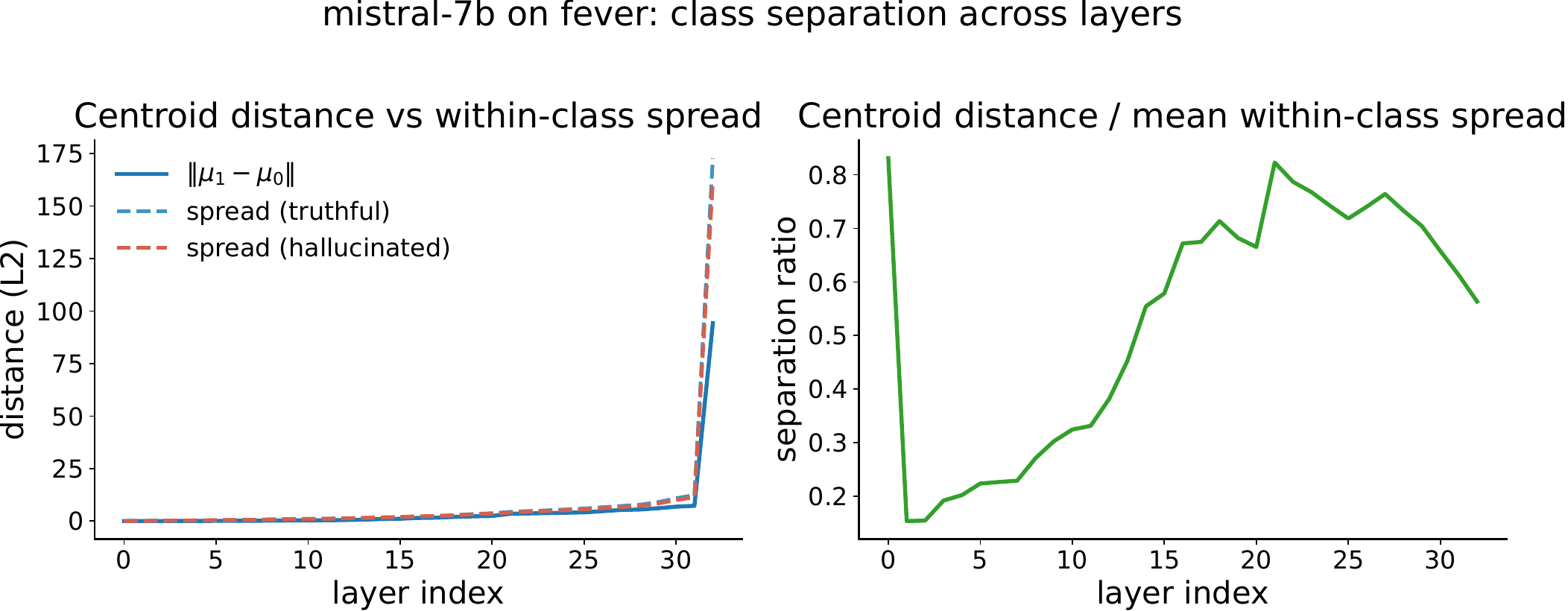}\\[3pt]
  \includegraphics[width=1\textwidth]{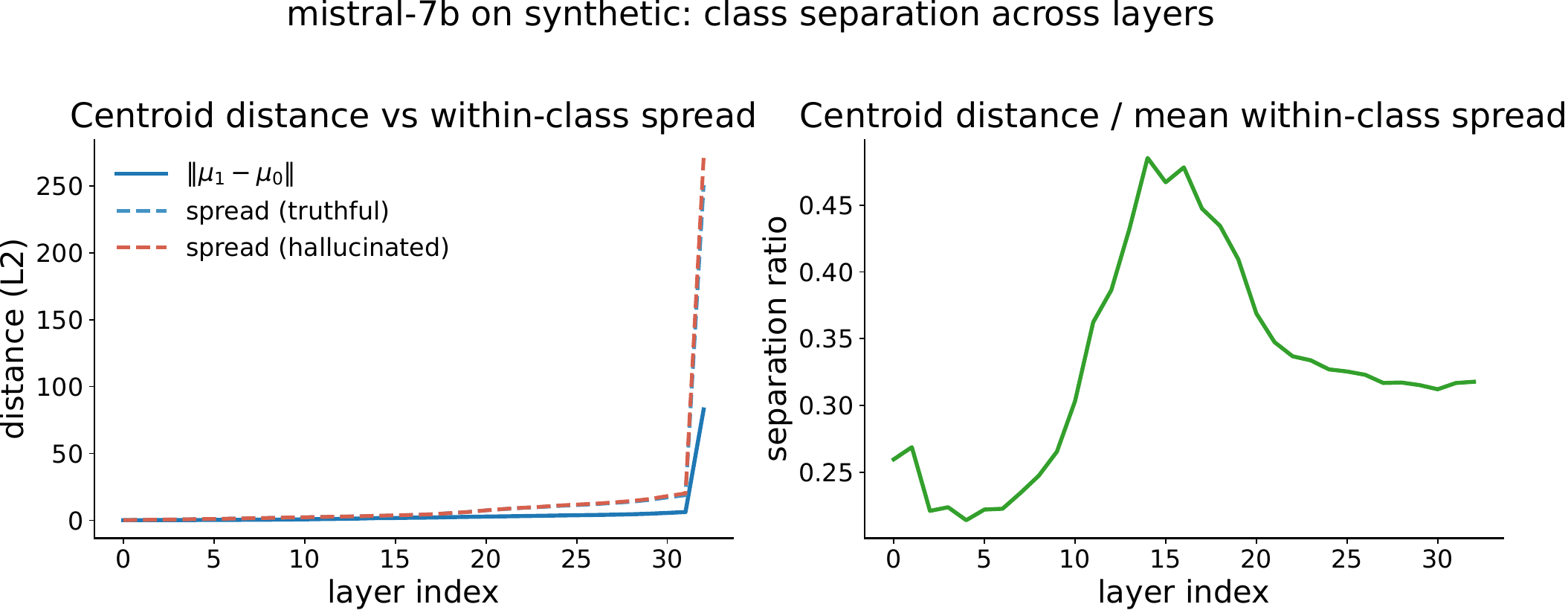}
\caption{Layer-wise class geometry for Mistral-7B on HaluEval-QA (top), FEVER (middle), and the synthetic benchmark (bottom). The same rise-and-plateau pattern holds across datasets. Cf.\ Figure~\ref{fig:sep-cross} (top) for TruthfulQA.}
  \label{fig:sep-mistral-all}
\end{figure}

\begin{figure}[htbp]
  \centering
  \includegraphics[width=1\textwidth]{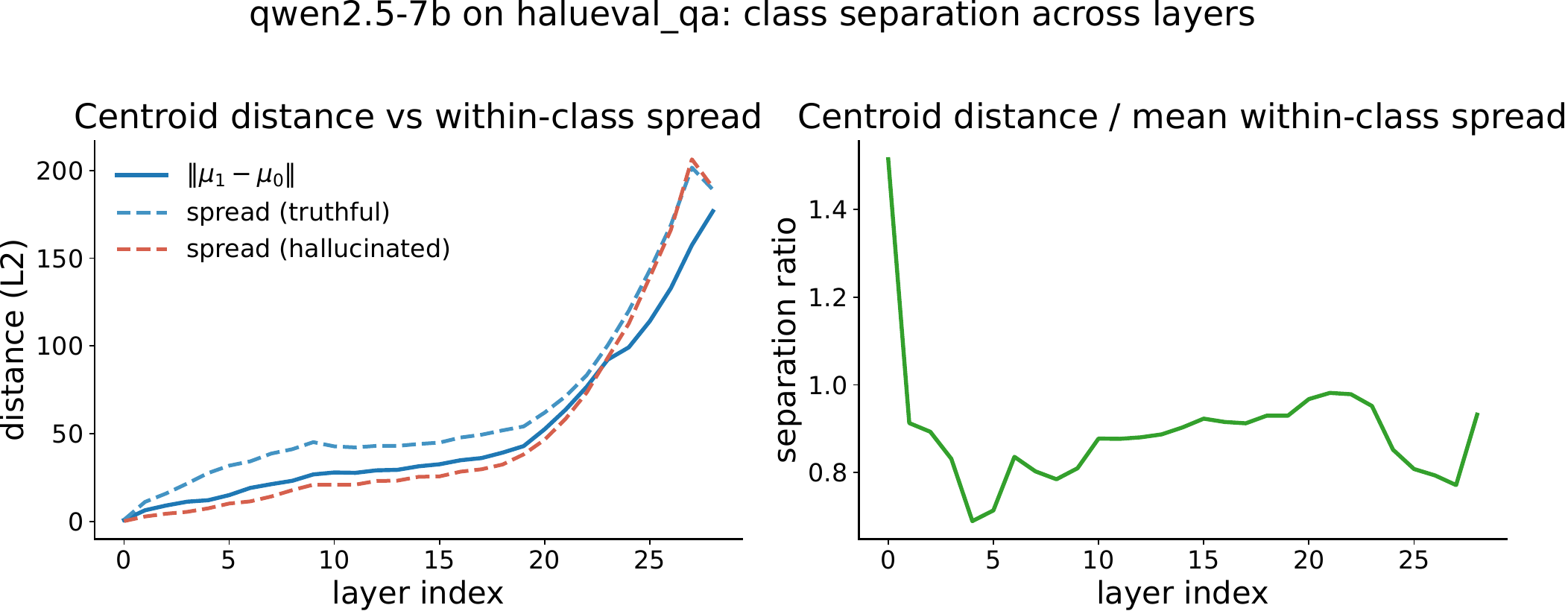}\\[3pt]
  \includegraphics[width=1\textwidth]{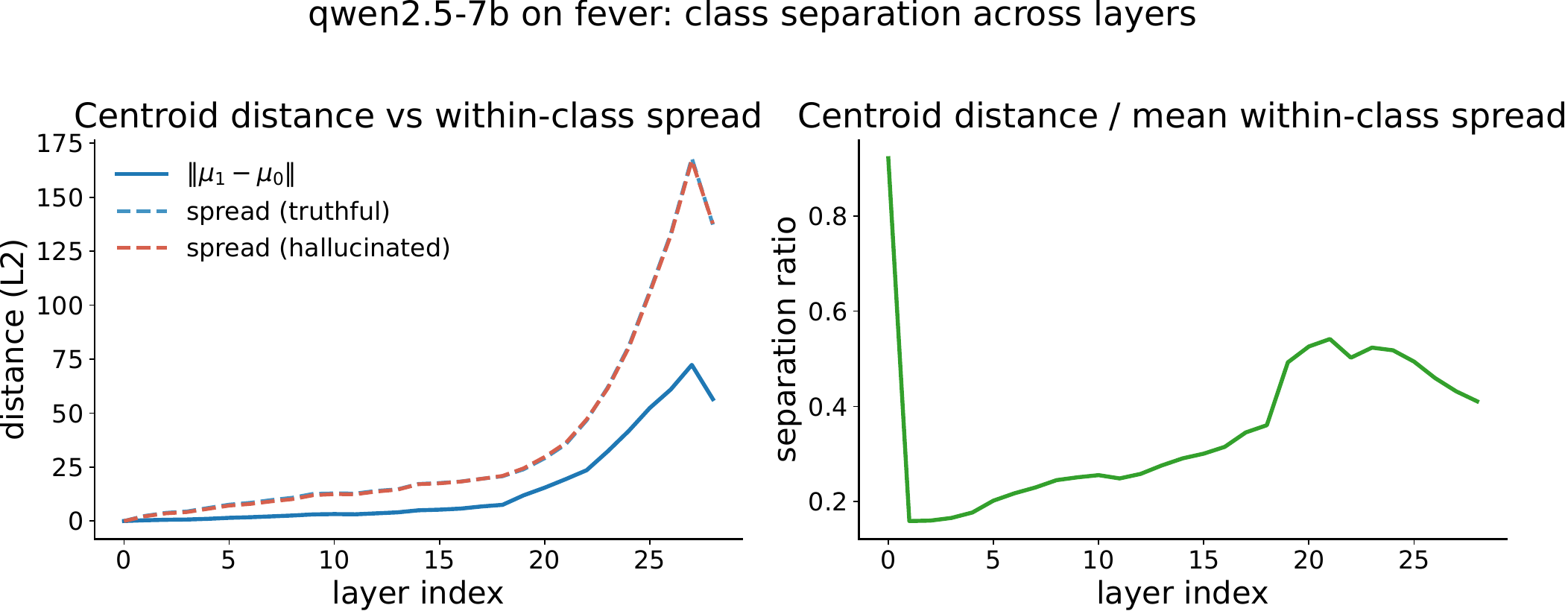}\\[3pt]
  \includegraphics[width=1\textwidth]{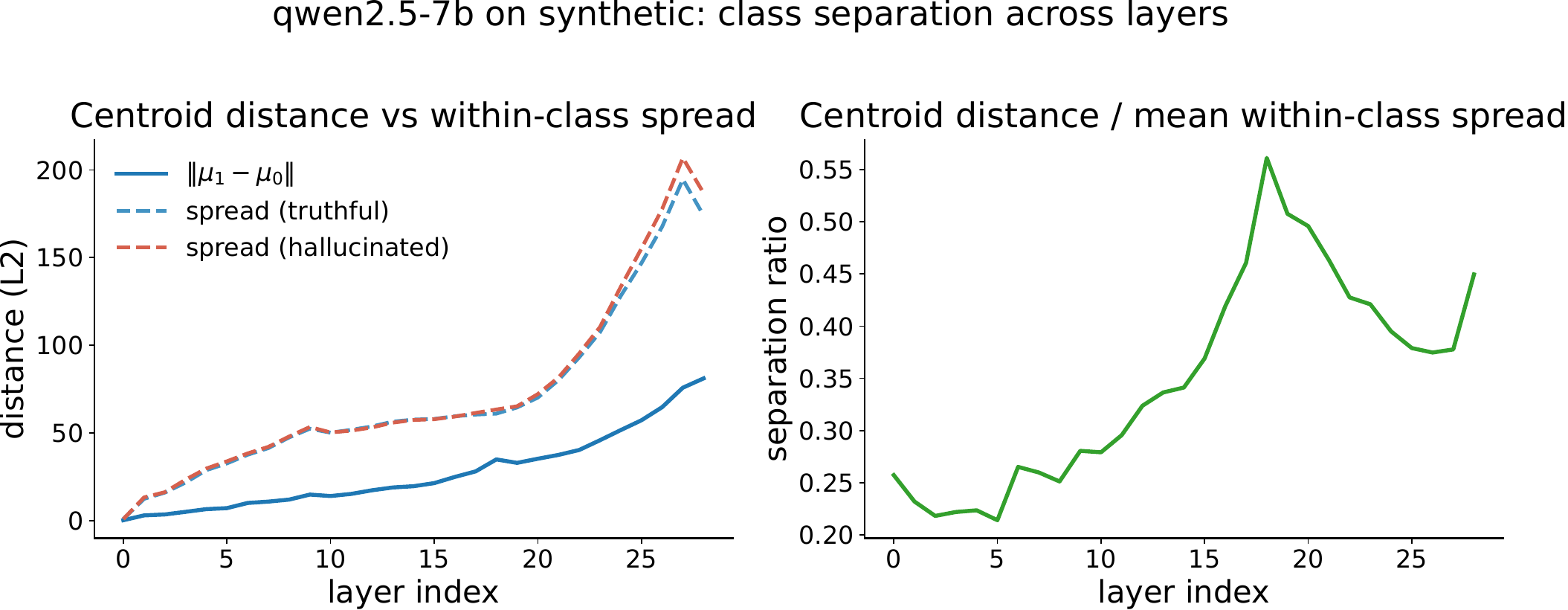}
\caption{Layer-wise class geometry for Qwen2.5-7B on HaluEval-QA (top), FEVER (middle), and the synthetic benchmark (bottom). The synthetic benchmark exhibits the steepest centroid drift, consistent with the ${\approx}1.000$ AUROC plateau. Cf.\ Figure~\ref{fig:sep-cross} (bottom) for TruthfulQA.}
  \label{fig:sep-qwen-all}
\end{figure}

Figures~\ref{fig:sep-llama-all}--\ref{fig:sep-qwen-all} extend the class-geometry diagnostic of Figure~\ref{fig:separation} to all $12$ configurations. The qualitative shape is preserved throughout: the centroid distance grows faster than the within-class spread, the separation ratio rises through the early blocks and plateaus in the mid-to-late blocks, and the depth at which it plateaus aligns with the probe-AUROC peak. The synthetic benchmark exhibits the steepest centroid drift, consistent with the $\approx\!1.000$ AUROC plateau.

\clearpage

\section{INSIDE EigenScore Distributions Across All Cells}
\label{sec:appendix-inside-all}

\begin{figure}[!ht]
  \centering
  \includegraphics[width=0.33\textwidth]{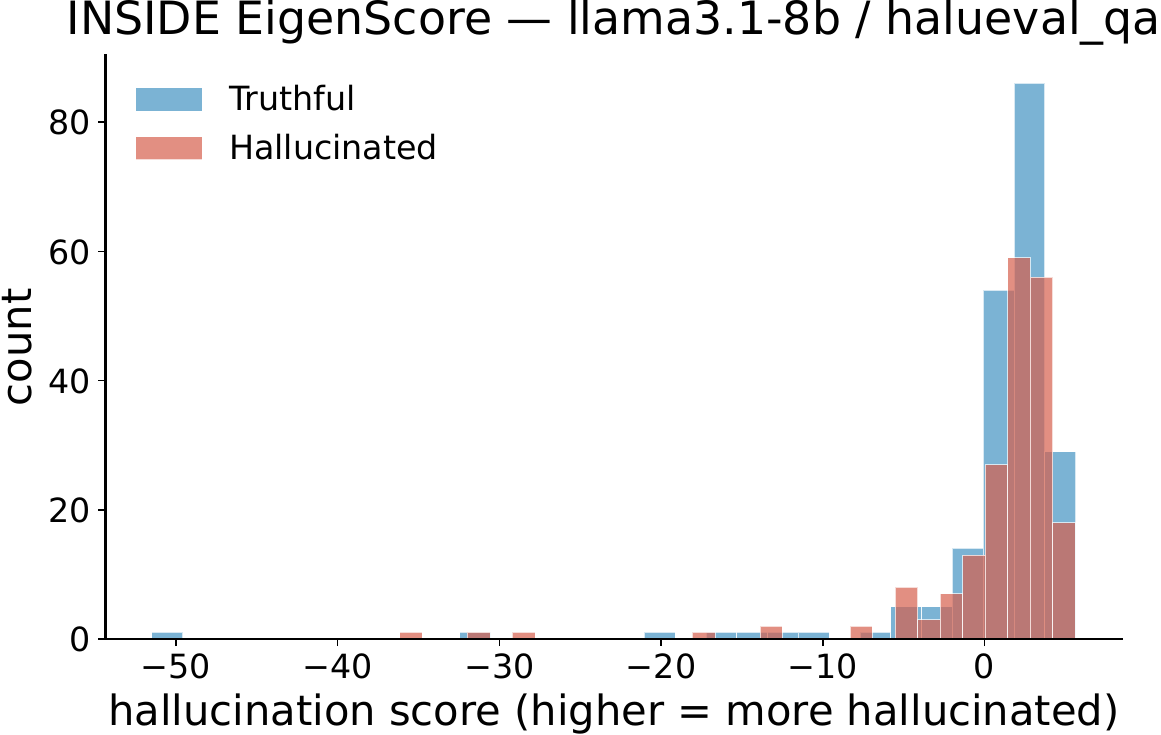}\hfill
  \includegraphics[width=0.33\textwidth]{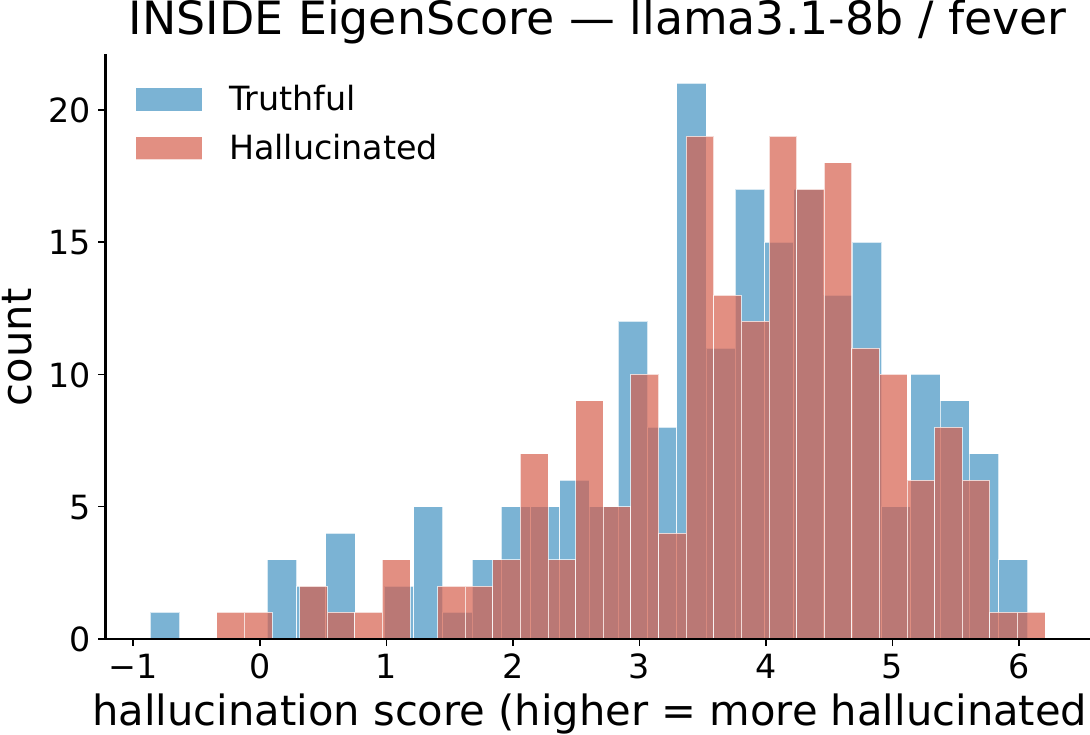}\hfill
  \includegraphics[width=0.33\textwidth]{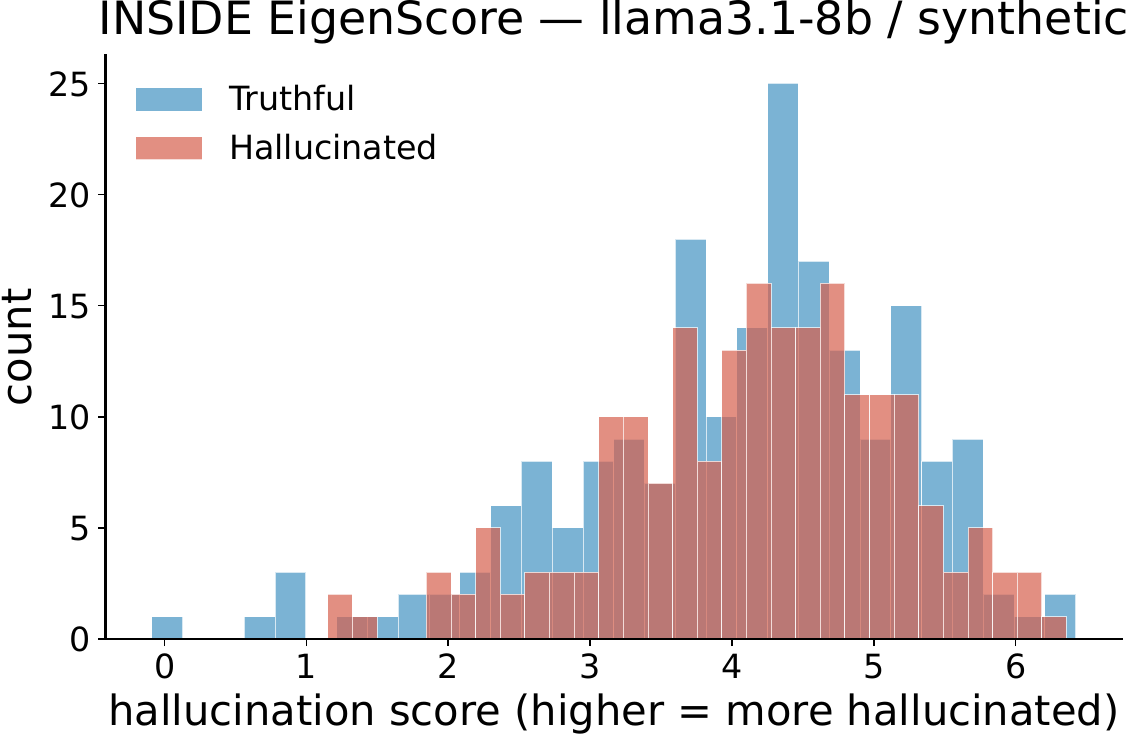}
\caption{INSIDE EigenScore distributions for Llama-3.1-8B on HaluEval-QA (top), FEVER (middle), and the synthetic benchmark (bottom), separated by ground-truth label. The two class distributions are nearly indistinguishable in every panel. Cf.\ Figure~\ref{fig:inside-dist} for TruthfulQA.}
  \label{fig:inside-llama-all}
\end{figure}

\begin{figure}[!ht]
  \centering
  \includegraphics[width=0.33\textwidth]{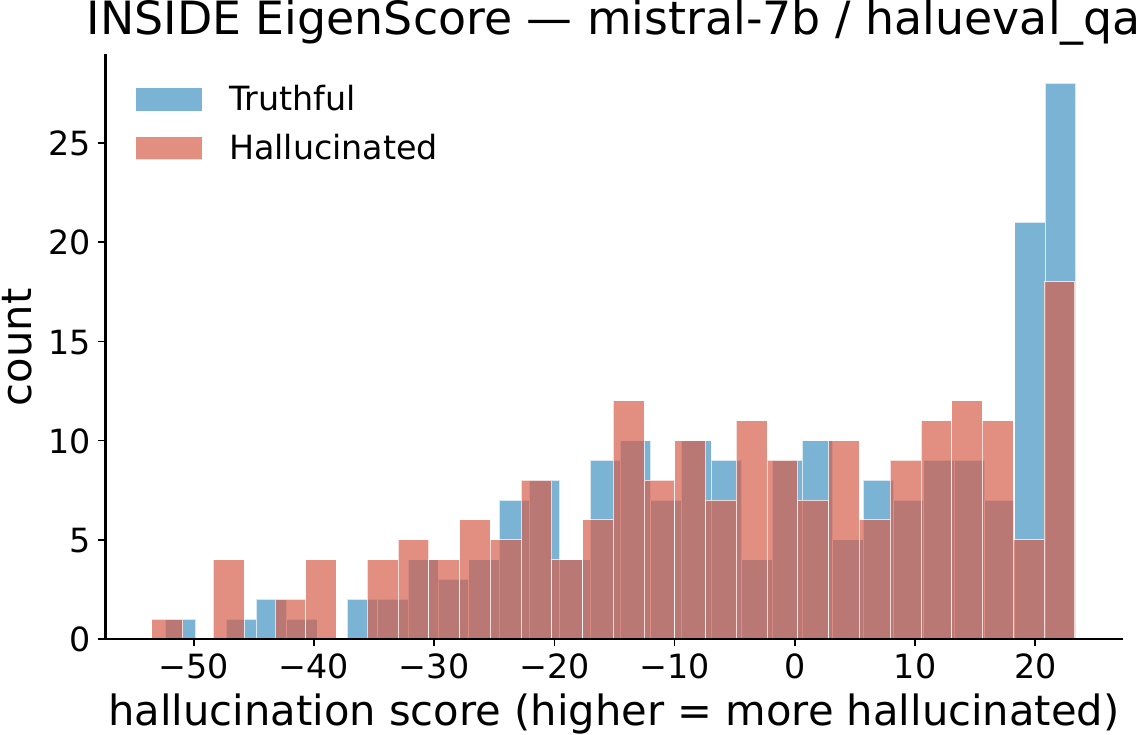}\hfill
  \includegraphics[width=0.33\textwidth]{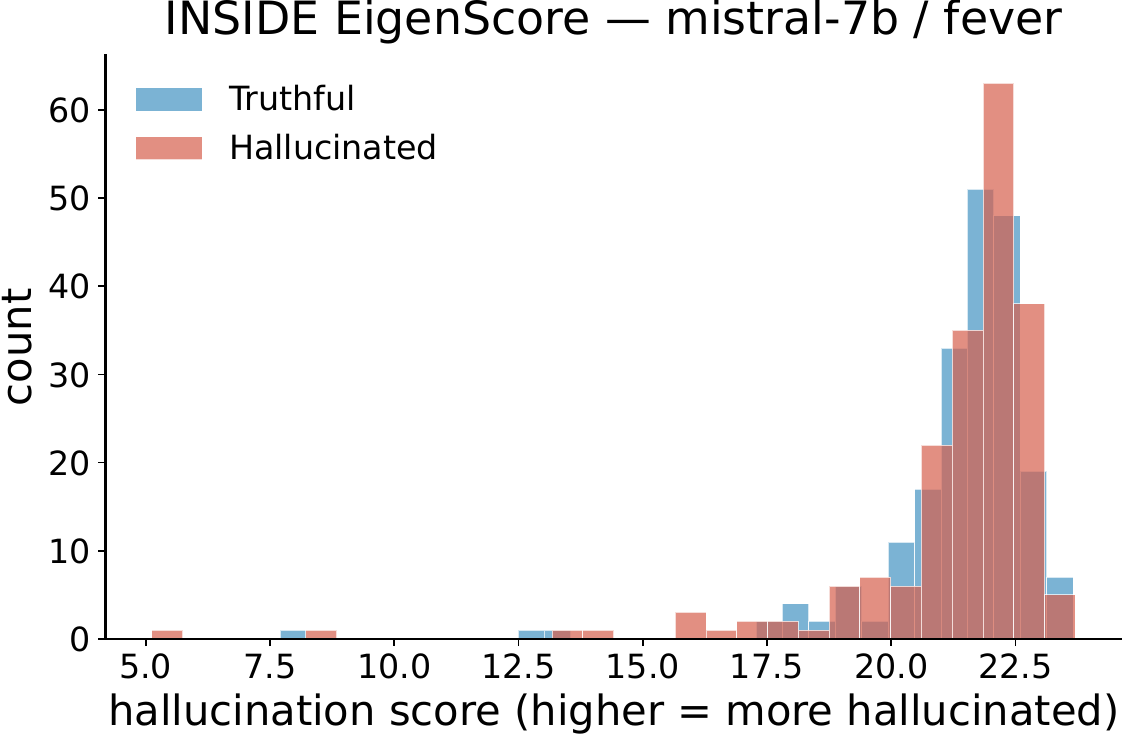}\hfill
  \includegraphics[width=0.33\textwidth]{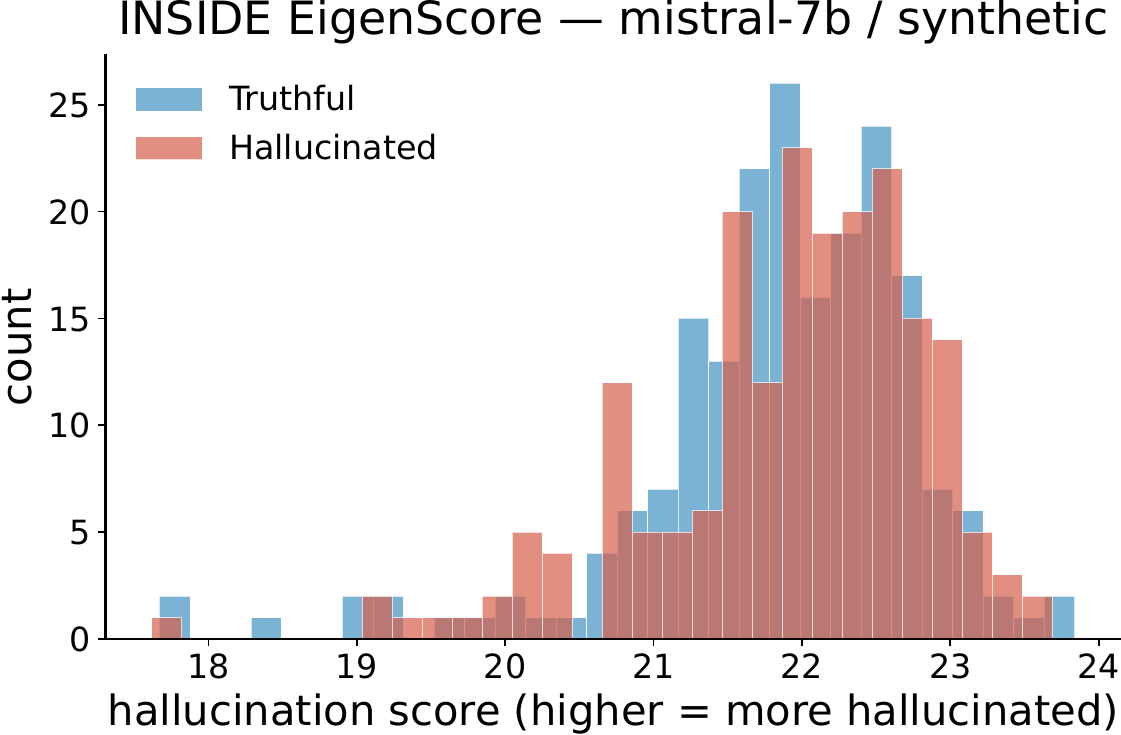}
\caption{INSIDE EigenScore distributions for Mistral-7B on HaluEval-QA (top), FEVER (middle), and the synthetic benchmark (bottom). The near-complete overlap persists across all datasets. Cf.\ Figure~\ref{fig:inside-cross} (top) for TruthfulQA.}
  \label{fig:inside-mistral-all}
\end{figure}

\begin{figure}[!ht]
  \centering
  \includegraphics[width=0.33\textwidth]{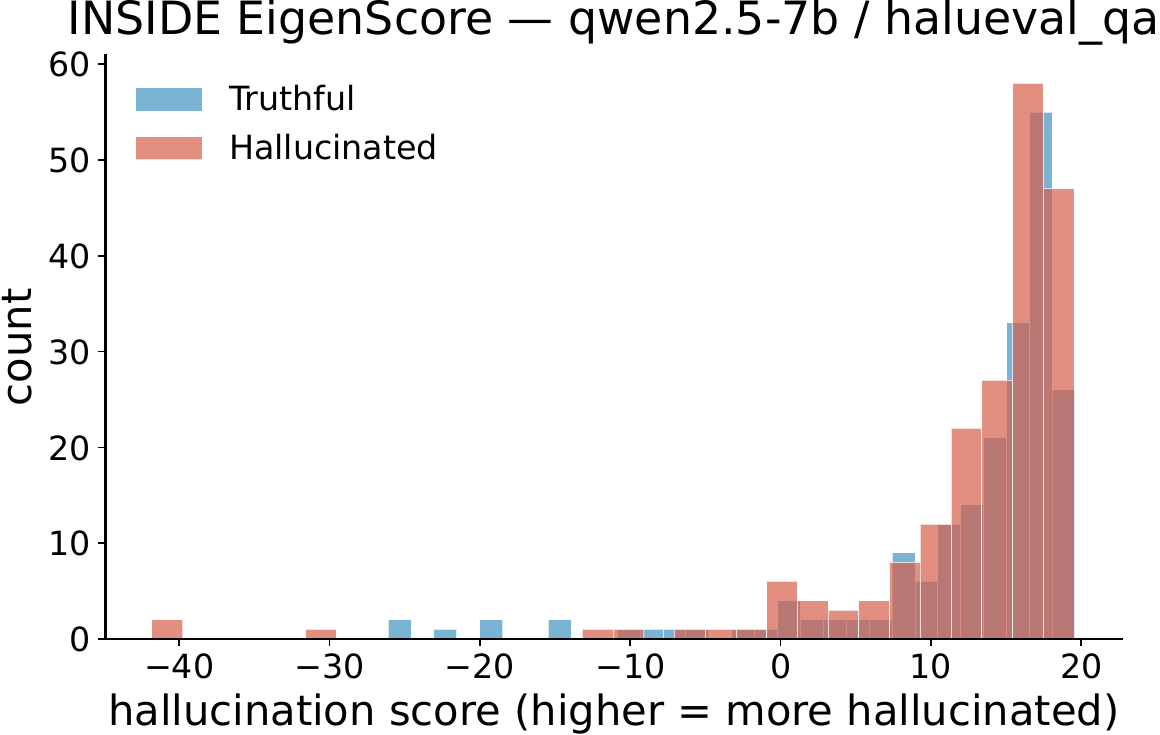}\hfill
  \includegraphics[width=0.33\textwidth]{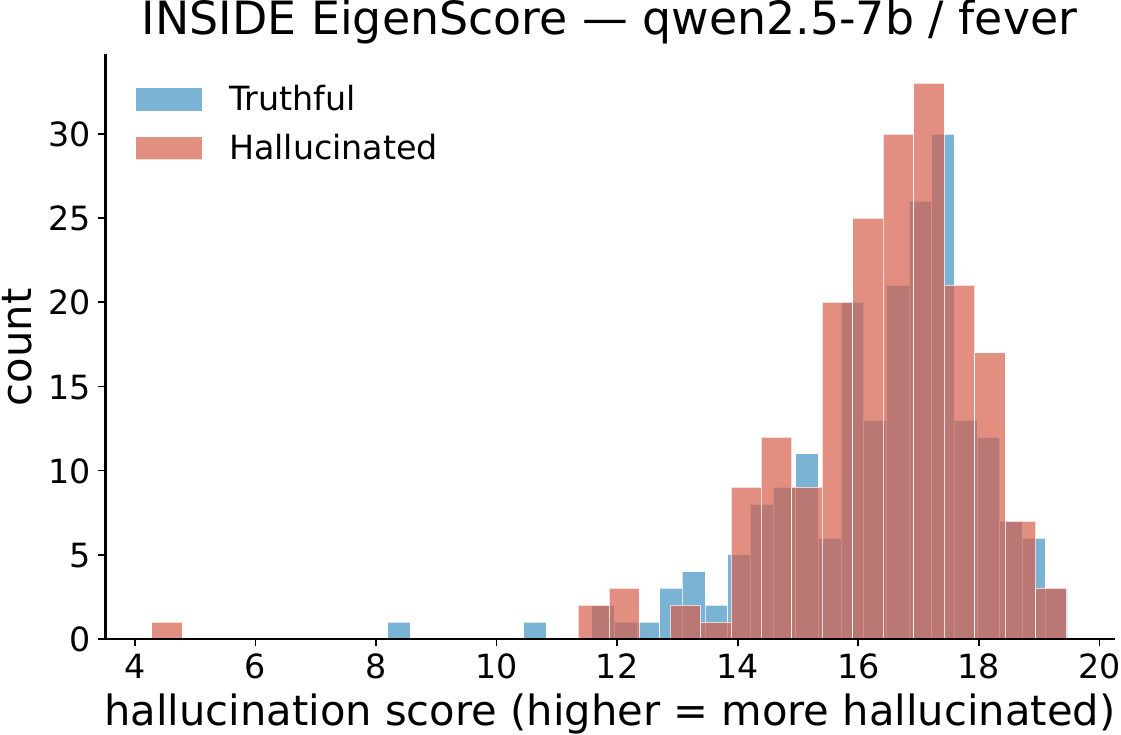}\hfill
  \includegraphics[width=0.33\textwidth]{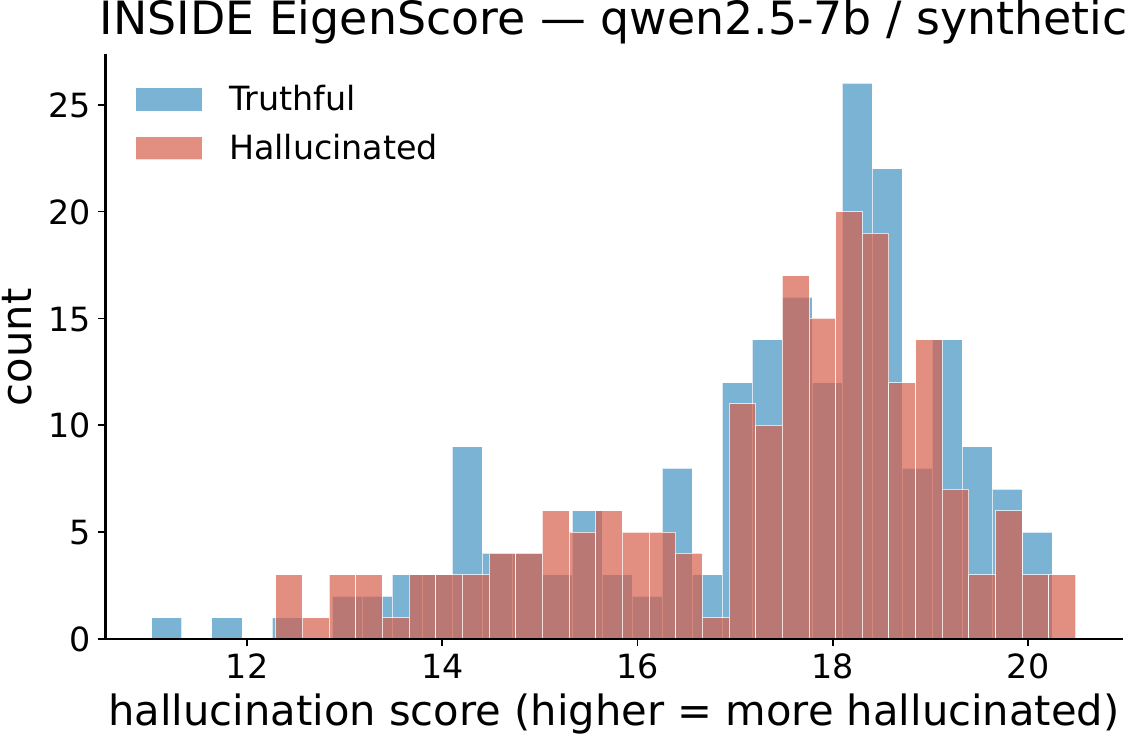}
\caption{INSIDE EigenScore distributions for Qwen2.5-7B on HaluEval-QA (left), FEVER (middle), and the synthetic benchmark (right). The uniform overlap across all $12$ configurations confirms that the chance-level INSIDE AUROC ($0.433$--$0.529$) is inherent to the evaluation protocol. Cf.\ Figure~\ref{fig:inside-cross} (bottom) for TruthfulQA.}
  \label{fig:inside-qwen-all}
\end{figure}

Figures~\ref{fig:inside-llama-all}--\ref{fig:inside-qwen-all} reproduce the per-class INSIDE EigenScore histograms across all $12$ configurations, complementing Figure~\ref{fig:inside-dist} (Llama / TruthfulQA) and Figure~\ref{fig:inside-cross} (Mistral, Qwen on TruthfulQA). In every panel the truthful and hallucinated distributions are nearly indistinguishable. This is a uniform finding: the INSIDE AUROC remains in the narrow band $[0.433, 0.529]$ across all $12$ entries in Table~\ref{tab:headline}.

\vspace{-1em}

\vspace{-3em}

\vspace{-3em}

\clearpage

\section{Attention Entropy on Non-Knowledge-Grounded Datasets}
\label{sec:appendix-attn-all}

\begin{figure}[!ht]
  \centering
  \includegraphics[width=0.88\textwidth]{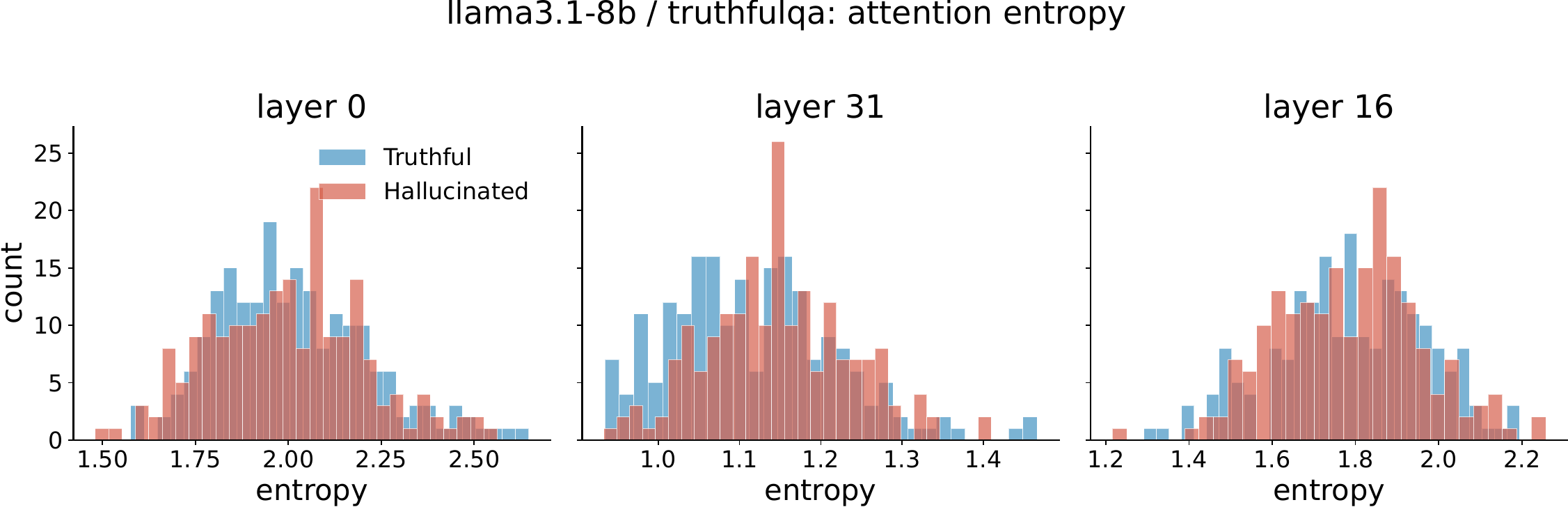}\\[3pt]
  \includegraphics[width=0.88\textwidth]{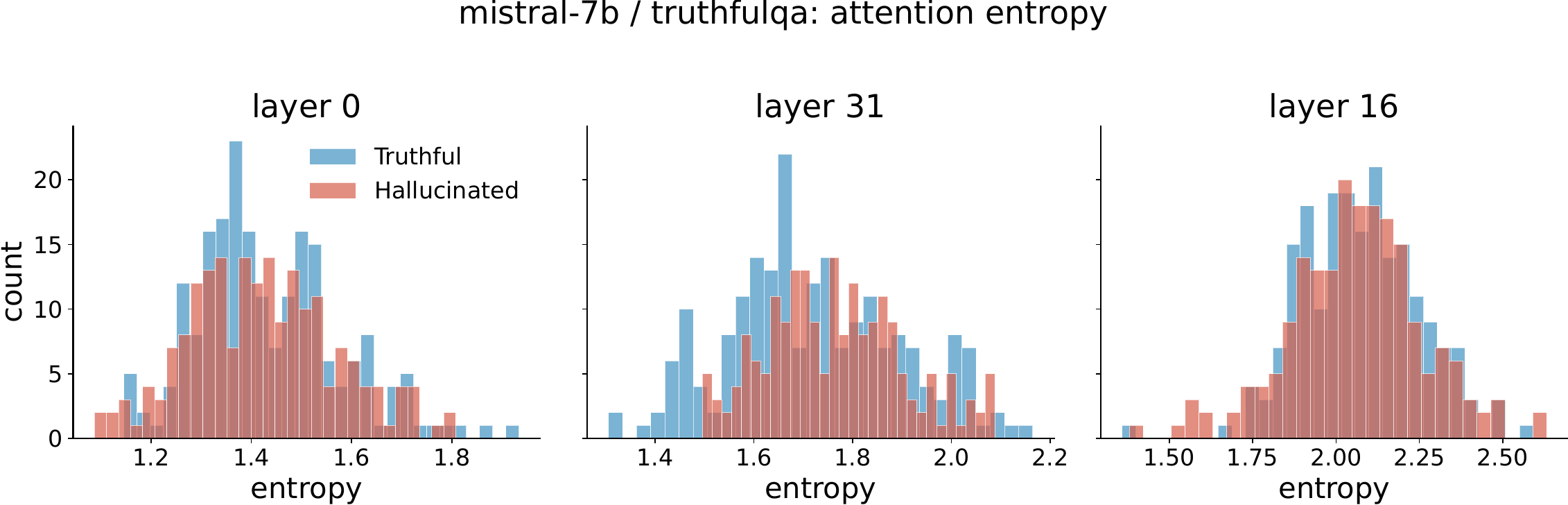}\\[3pt]
  \includegraphics[width=0.88\textwidth]{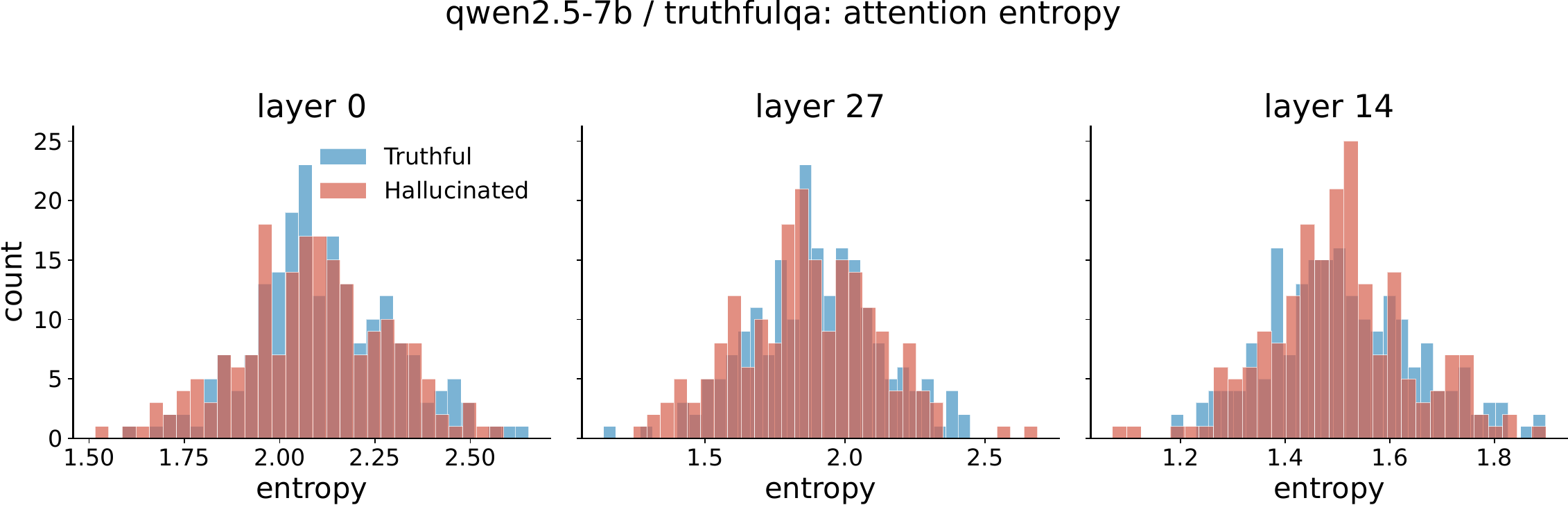}
  \vspace{-2mm}
  \caption{Last-token attention entropy on TruthfulQA for Llama-3.1-8B (top), Mistral-7B (middle), and Qwen2.5-7B (bottom) at three captured transformer blocks. The truthful and hallucinated histograms overlap substantially in every panel, with the best per-model AUROC remaining below $0.61$.}
  \label{fig:attn-truthfulqa-all}
\end{figure}

\begin{figure}[ht]
  \centering
  \includegraphics[width=1\textwidth]{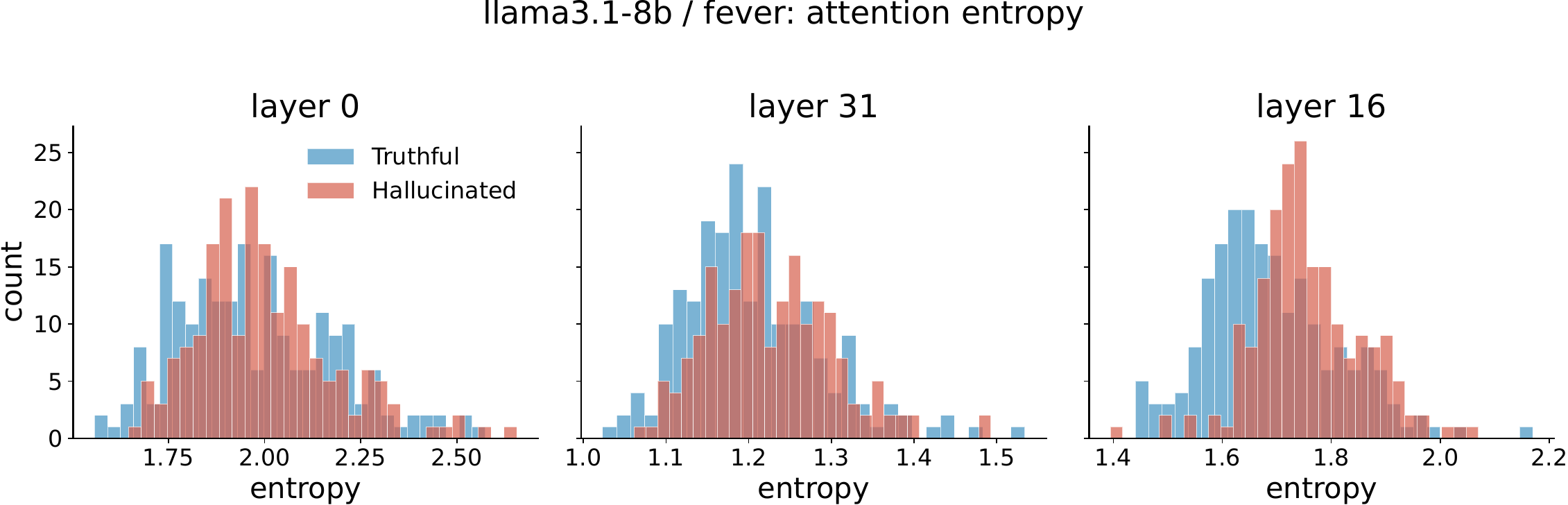}\\[3pt]
  \includegraphics[width=1\textwidth]{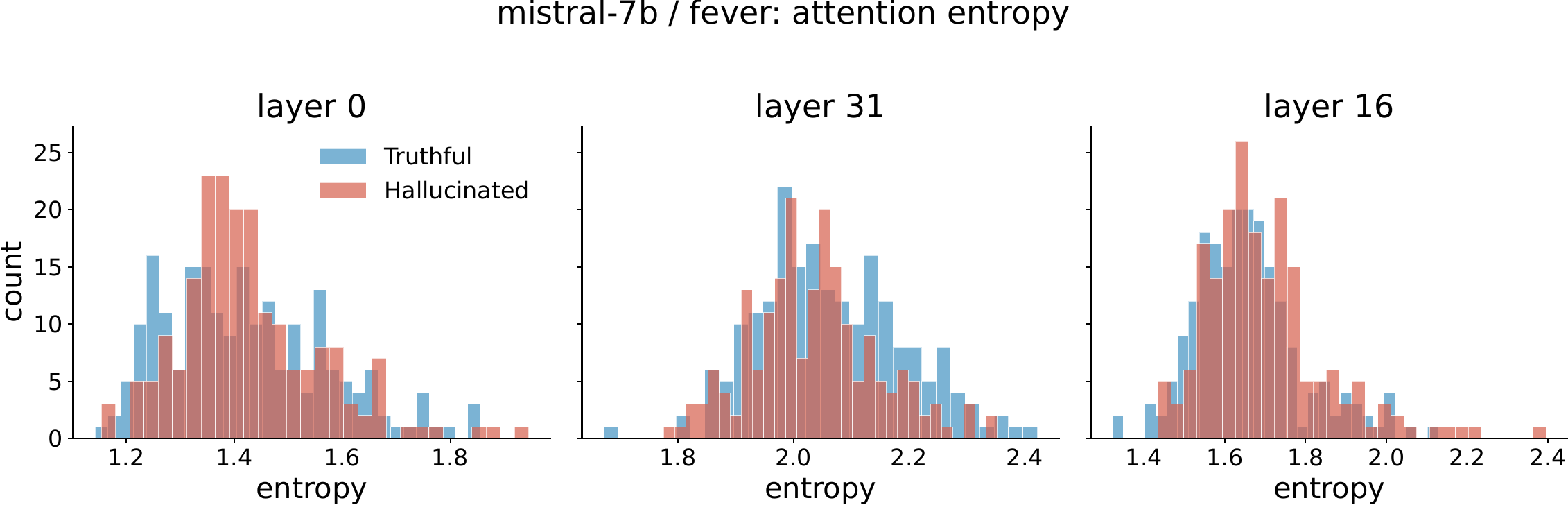}\\[3pt]
  \includegraphics[width=1\textwidth]{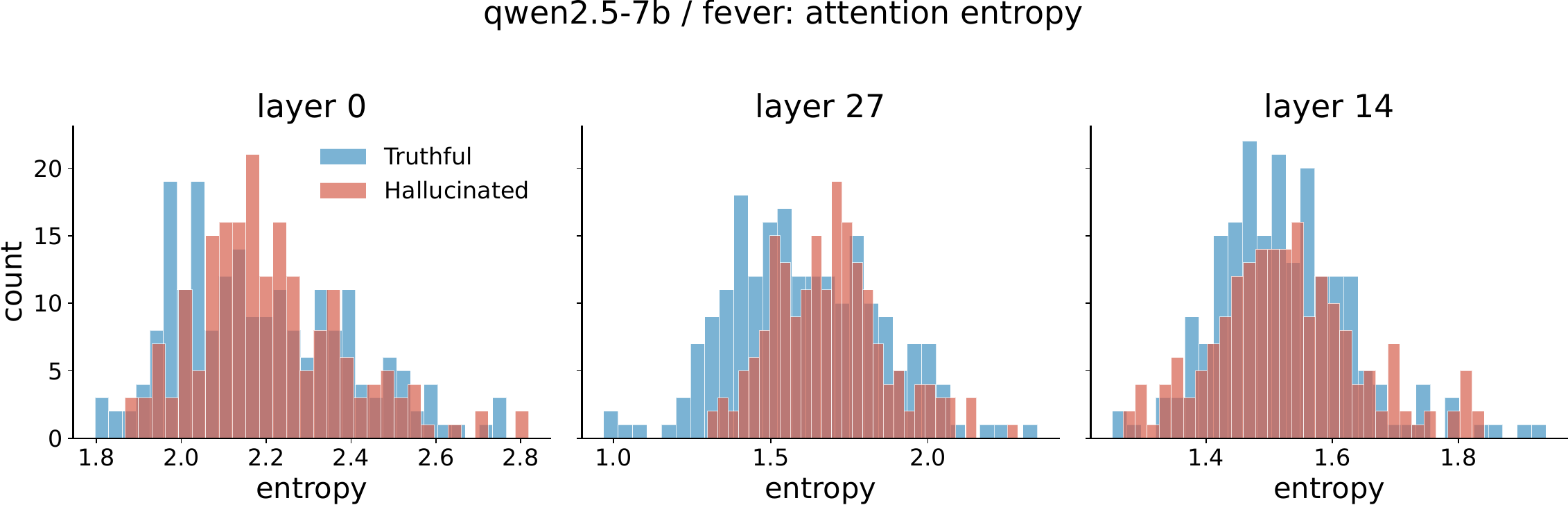}
\caption{Last-token attention entropy on FEVER for Llama-3.1-8B (top), Mistral-7B (middle), and Qwen2.5-7B (bottom). Without an explicit supporting-knowledge channel, the per-class entropy histograms are no longer cleanly separable, contrasting with the strong first-block signal on HaluEval-QA (Figure~\ref{fig:attn-cross}).}
  \label{fig:attn-fever-all}
\end{figure}

\begin{figure}[htbp]
  \centering
  \includegraphics[width=1\textwidth]{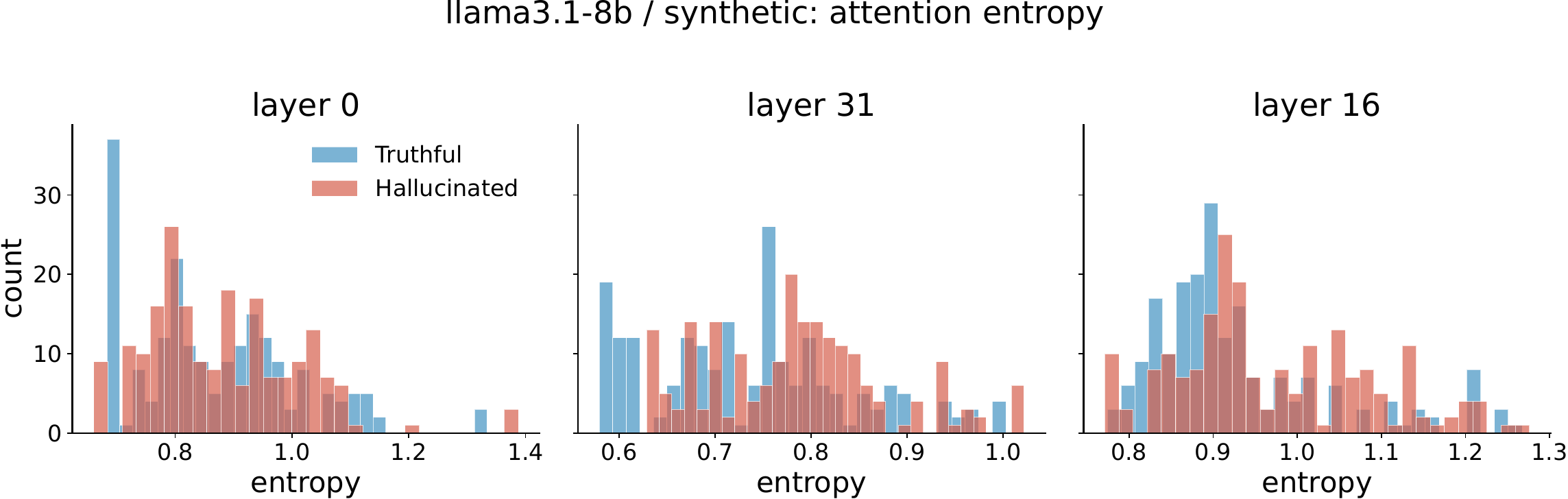}\\[3pt]
  \includegraphics[width=1\textwidth]{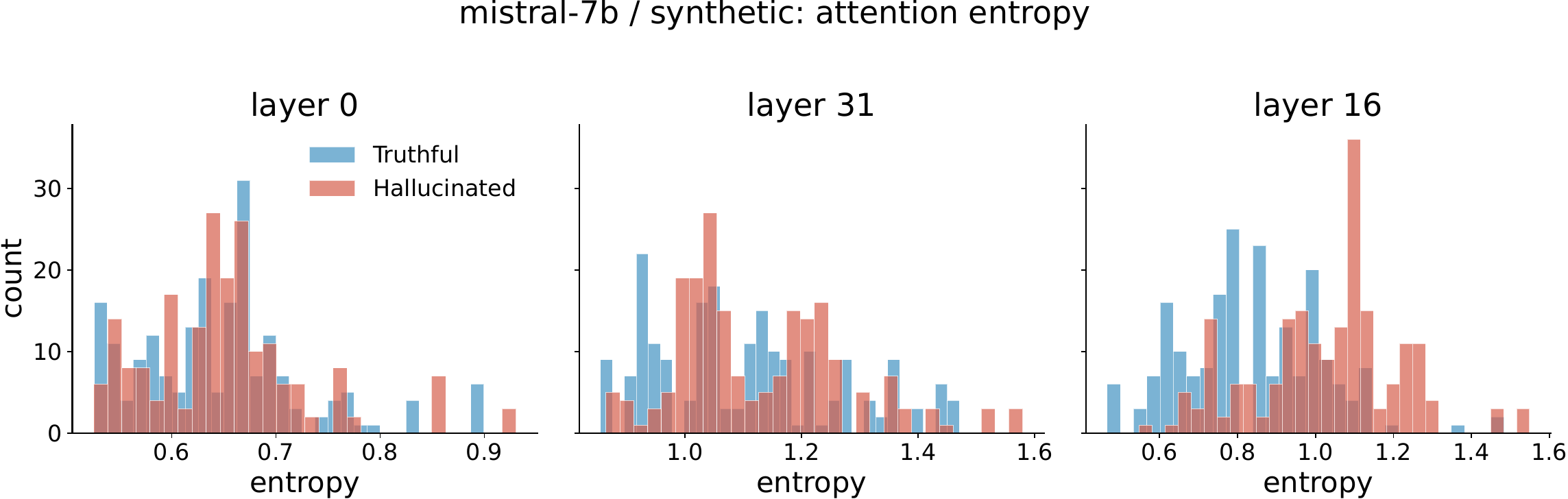}\\[3pt]
  \includegraphics[width=1\textwidth]{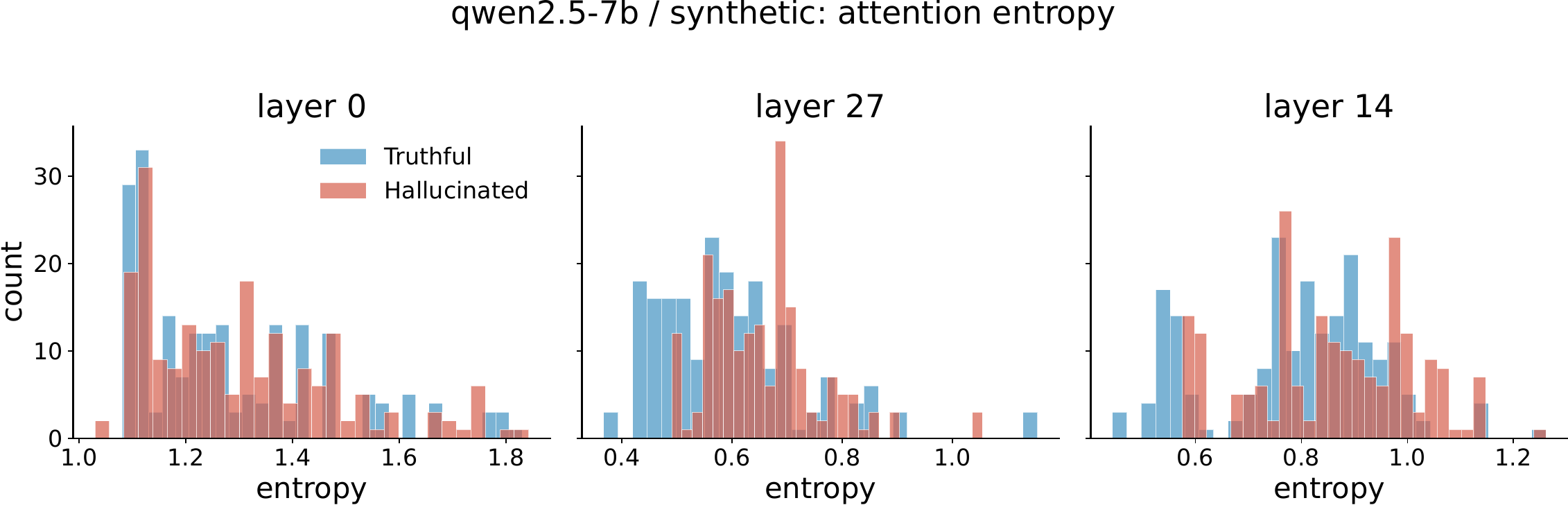}
\caption{Last-token attention entropy on the synthetic benchmark for Llama-3.1-8B (top), Mistral-7B (middle), and Qwen2.5-7B (bottom). The middle block performs best for Mistral (AUROC $0.774$), but the signal remains far weaker than probe-based detectors across all three models.}
  \label{fig:attn-synthetic-all}
\end{figure}

Figures~\ref{fig:attn-truthfulqa-all}--\ref{fig:attn-synthetic-all} report the head-averaged last-token attention entropy on TruthfulQA, FEVER, and the synthetic benchmark for all three models, substantiating that the first-block attention-entropy signal does \emph{not} generalize beyond HaluEval-QA. In each panel the three captured blocks (first, middle, last) are shown side by side. The truthful and hallucinated entropy histograms overlap substantially in all three non-HaluEval datasets, with the best attention-entropy AUROC ranging from $0.489$ (Qwen / TruthfulQA; Table~\ref{tab:headline}) to $0.774$ (Mistral / synthetic). This contrasts with HaluEval-QA, where first-block attention entropy reaches $0.866$--$0.941$ AUROC. The disparity is expected: only HaluEval-QA prepends an explicit evidence passage, giving the model a concrete span to attend to when the answer is supported.


\vspace{-3em}

\clearpage

\section{2-D Projections Across All Cells}
\label{sec:appendix-proj-all}

\begin{figure}[ht]
  \centering
  \includegraphics[width=0.48\columnwidth]{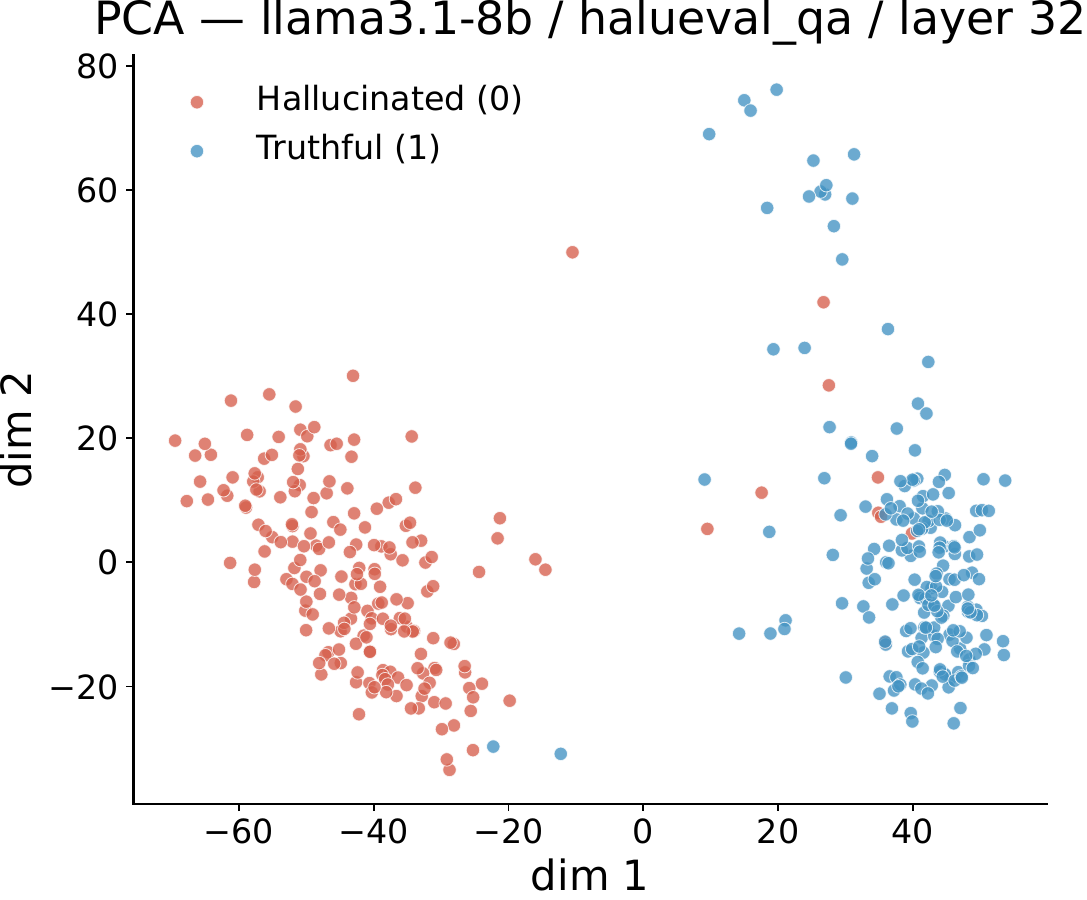}\hfill
  \includegraphics[width=0.48\columnwidth]{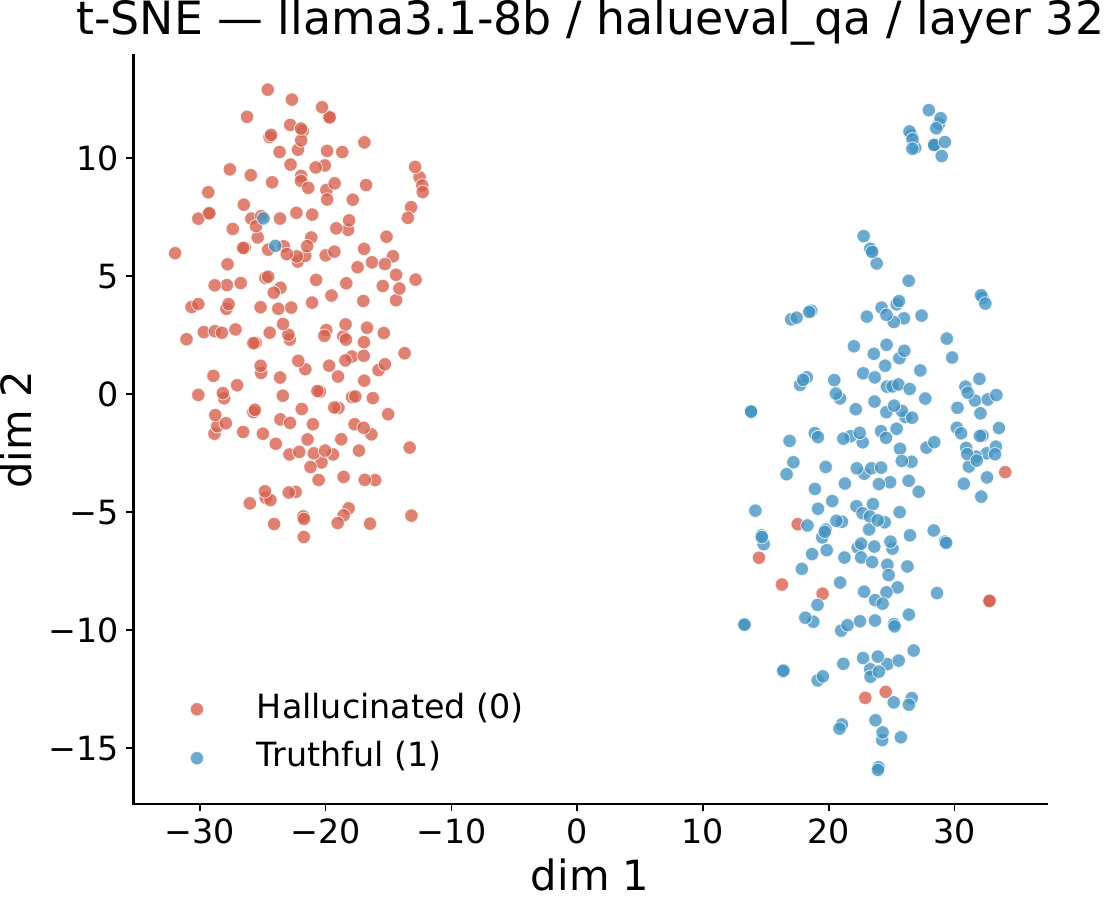}\\[2pt]
  \includegraphics[width=0.48\columnwidth]{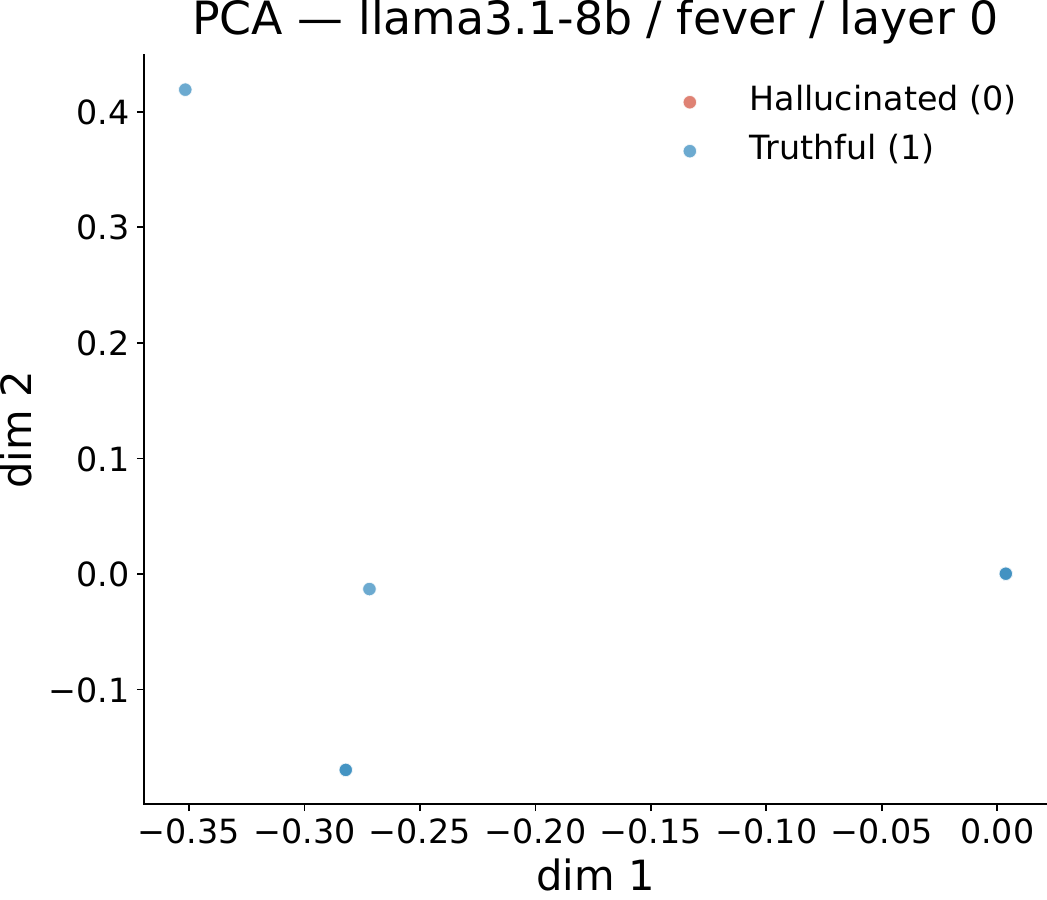}\hfill
  \includegraphics[width=0.48\columnwidth]{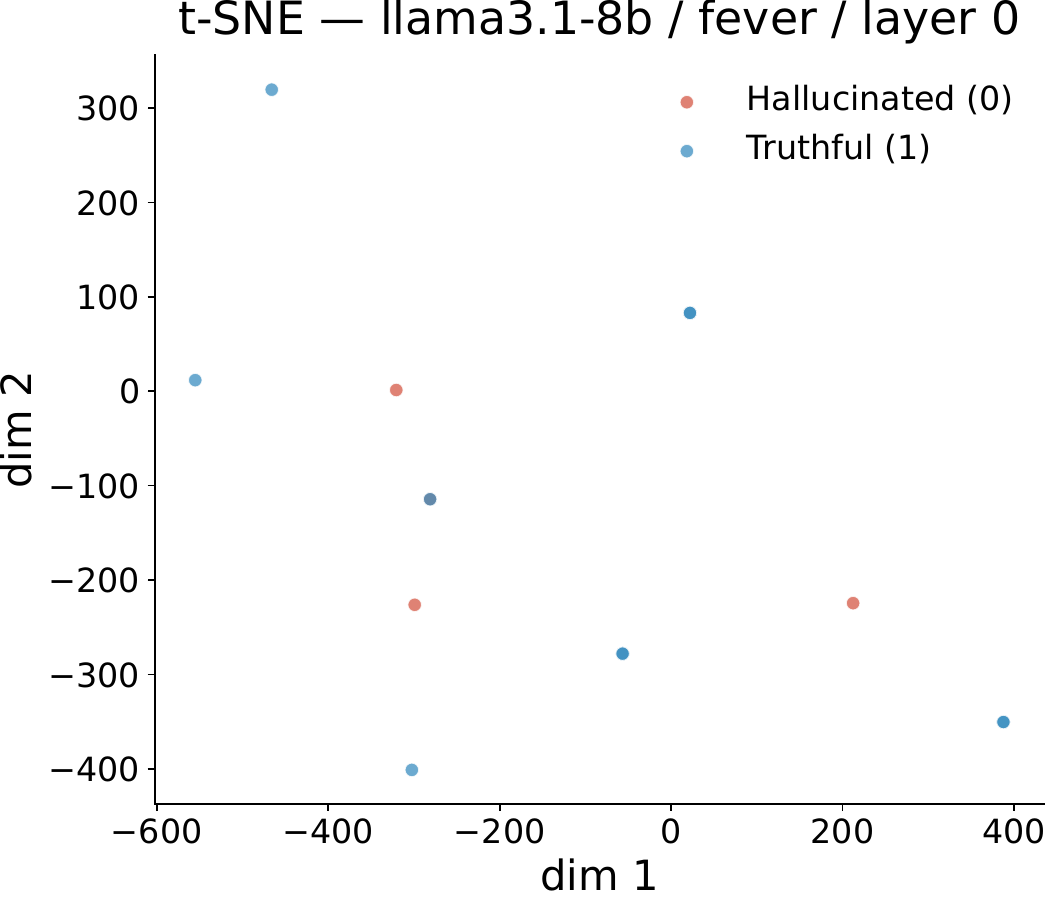}\\[2pt]
  \includegraphics[width=0.48\columnwidth]{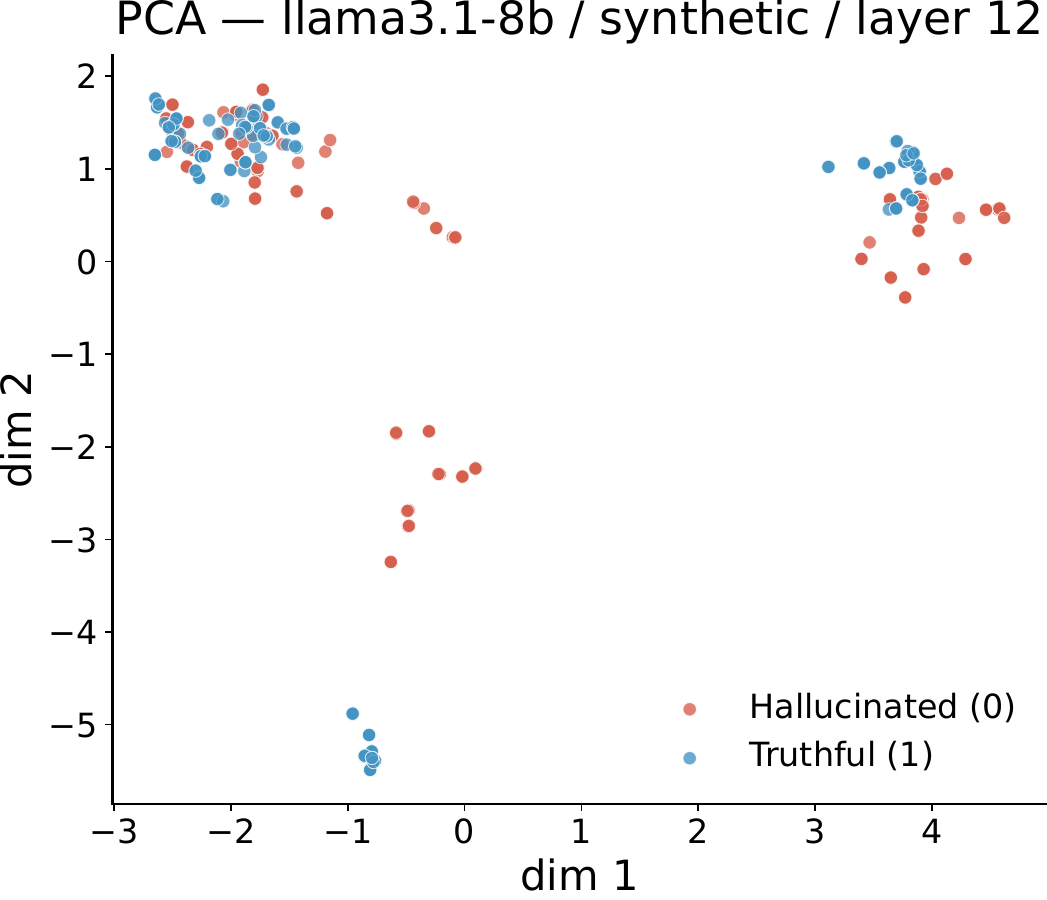}\hfill
  \includegraphics[width=0.48\columnwidth]{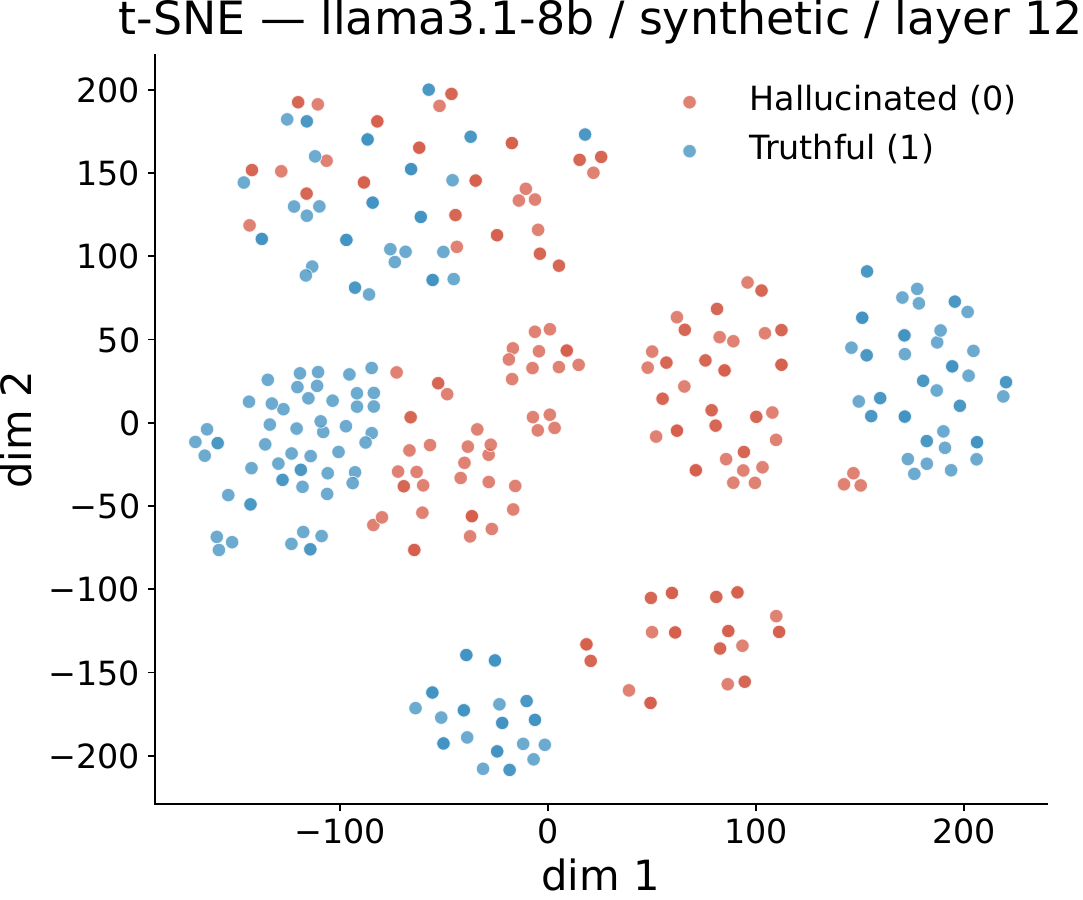}
\caption{PCA (left) and t-SNE (right) projections of Llama-3.1-8B hidden states at the block maximizing the classifier-free separation ratio: HaluEval-QA (top, block~$32$), FEVER (middle, block~$0$), and synthetic (bottom, block~$12$). Panels at block~$0$ show heavily overlapping clouds because the separation ratio peaks early due to lexical-identity effects rather than truthfulness geometry. Cf.\ Figures~\ref{fig:tsne},~\ref{fig:pca} for the TruthfulQA probing-peak projection.}
  \label{fig:proj-llama-all}
\end{figure}

\begin{figure}[ht]
  \centering
  \footnotesize
  \includegraphics[width=0.399\columnwidth]{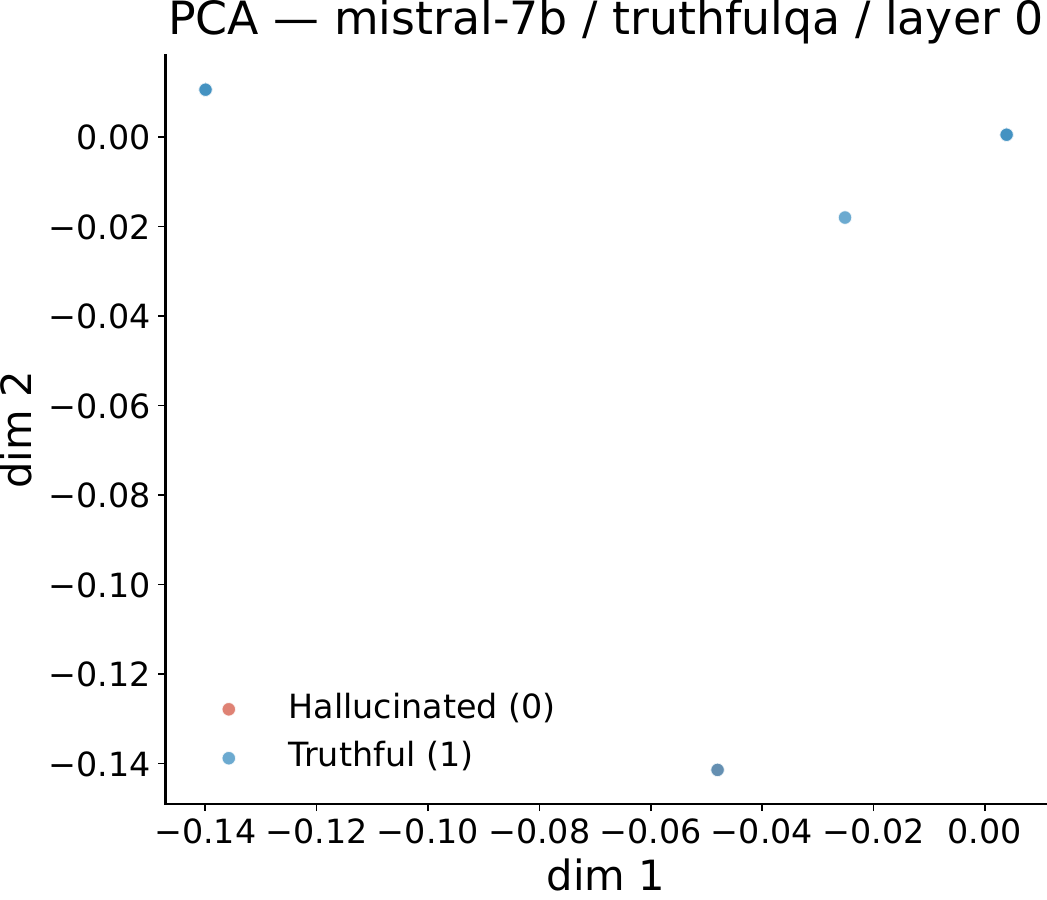}\hfill
  \includegraphics[width=0.399\columnwidth]{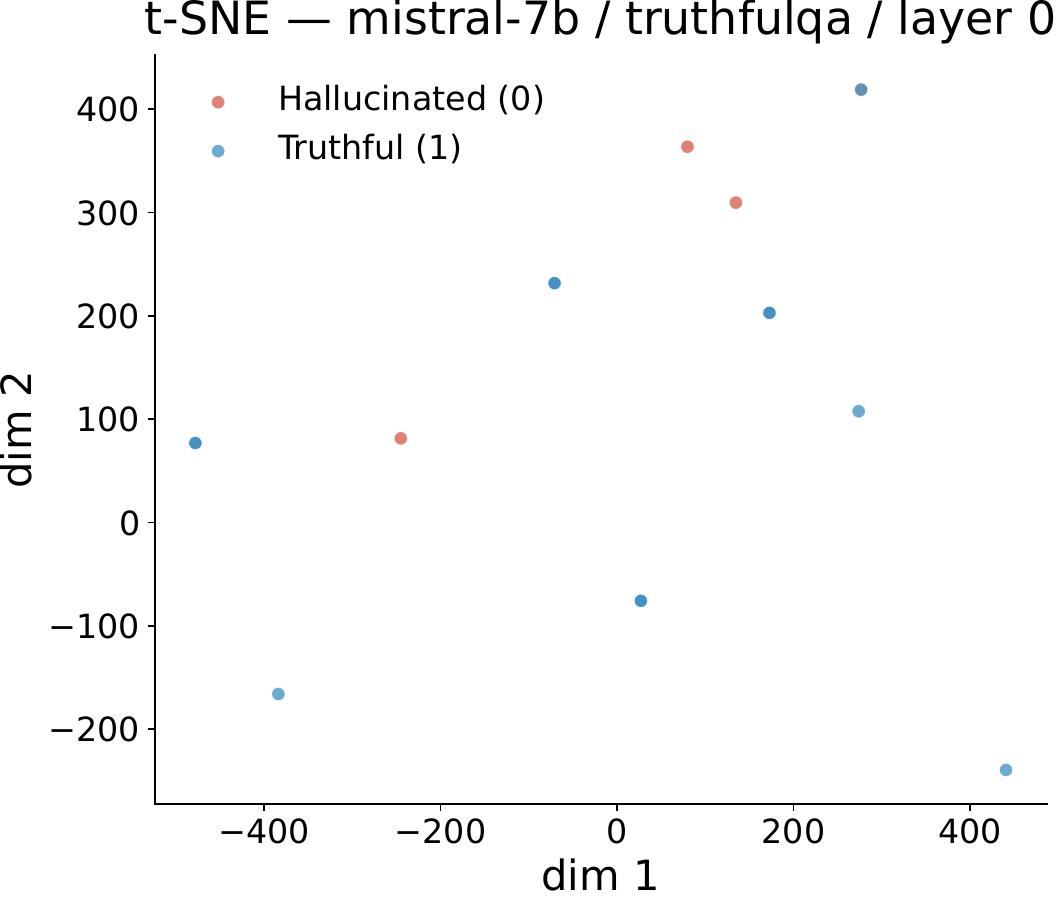}\\[2pt]
  \includegraphics[width=0.399\columnwidth]{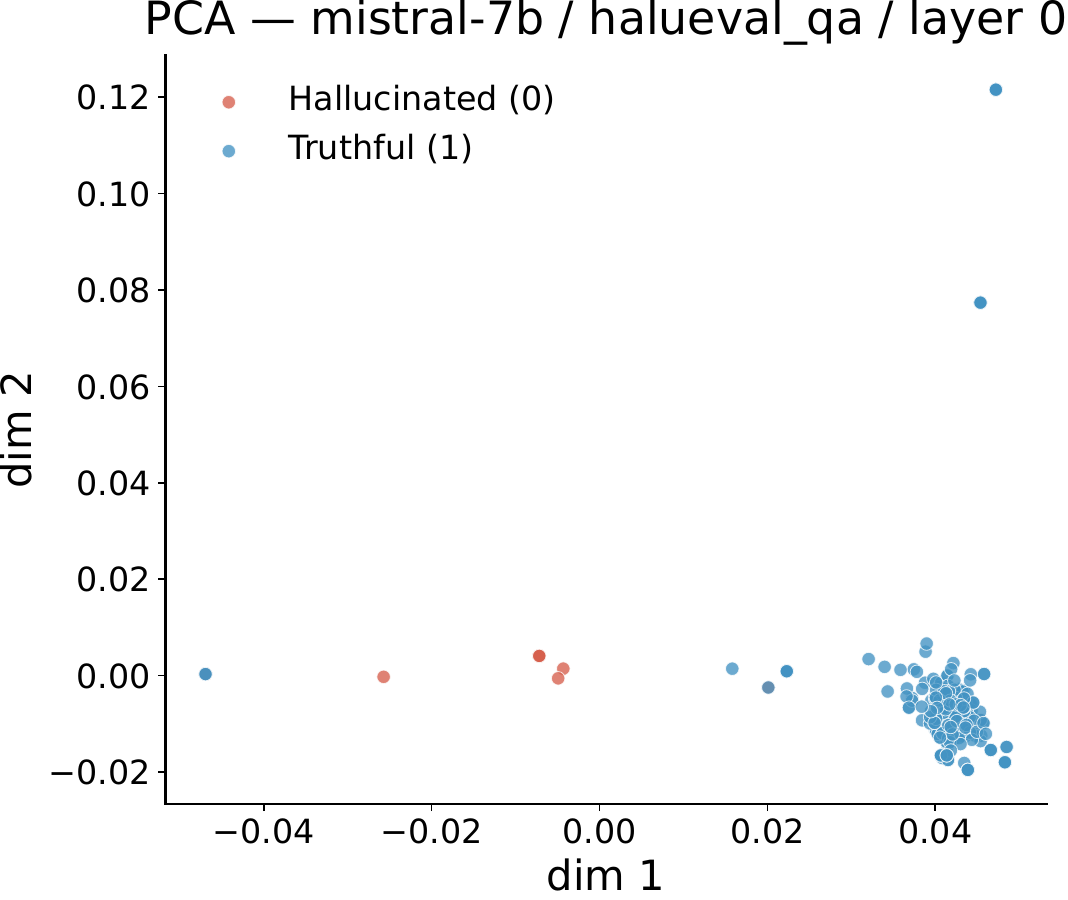}\hfill
  \includegraphics[width=0.399\columnwidth]{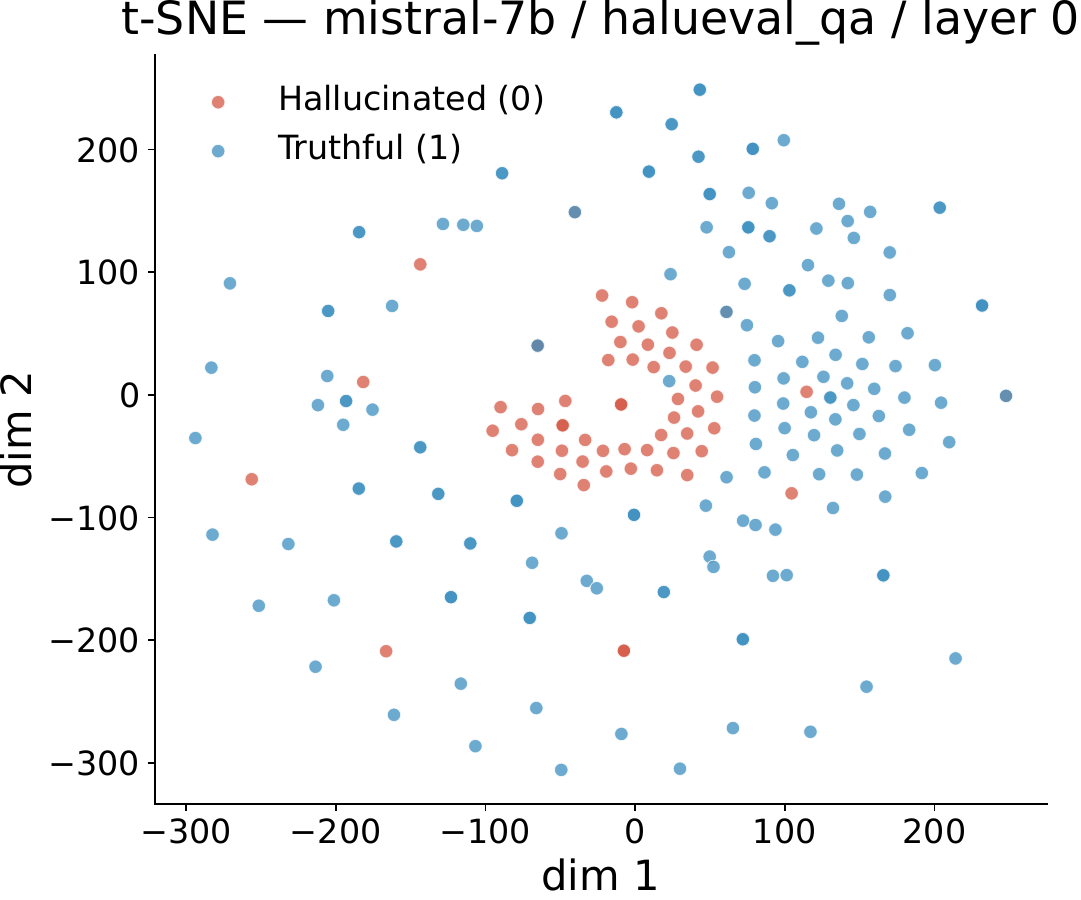}\\[2pt]
  \includegraphics[width=0.399\columnwidth]{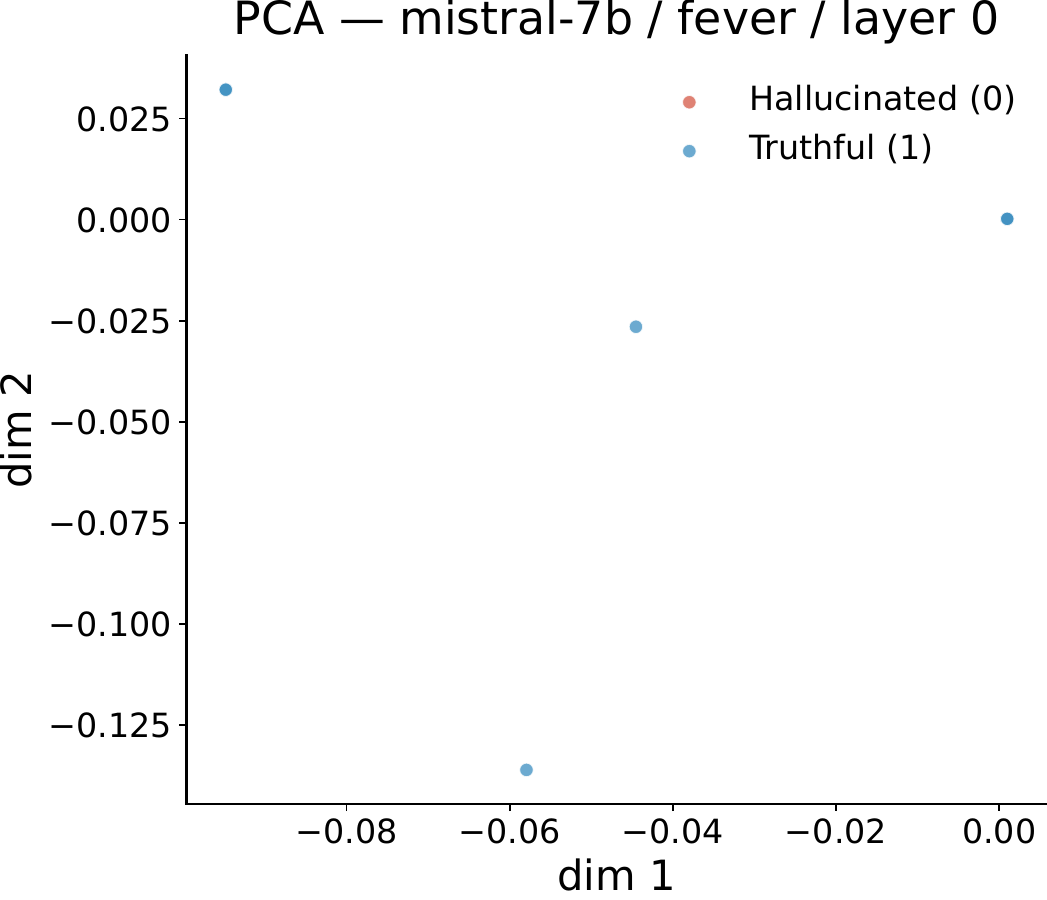}\hfill
  \includegraphics[width=0.399\columnwidth]{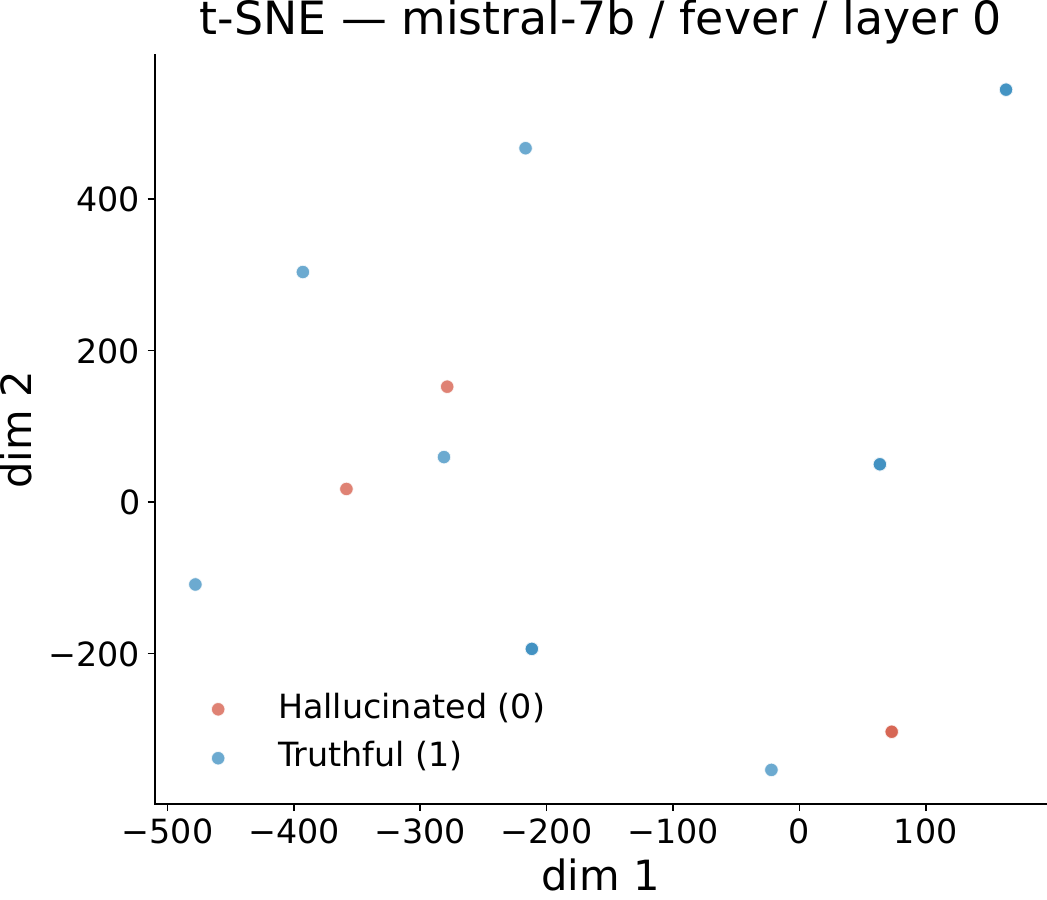}\\[2pt]
  \includegraphics[width=0.399\columnwidth]{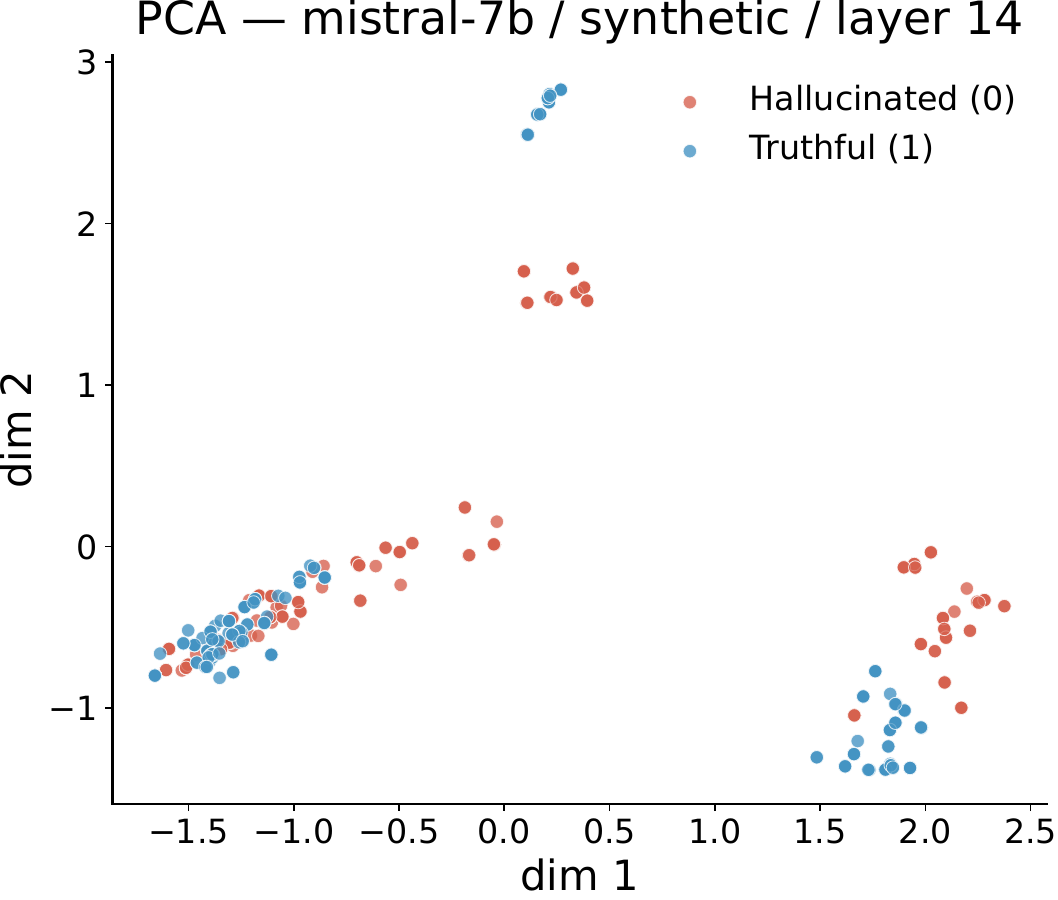}\hfill
  \includegraphics[width=0.399\columnwidth]{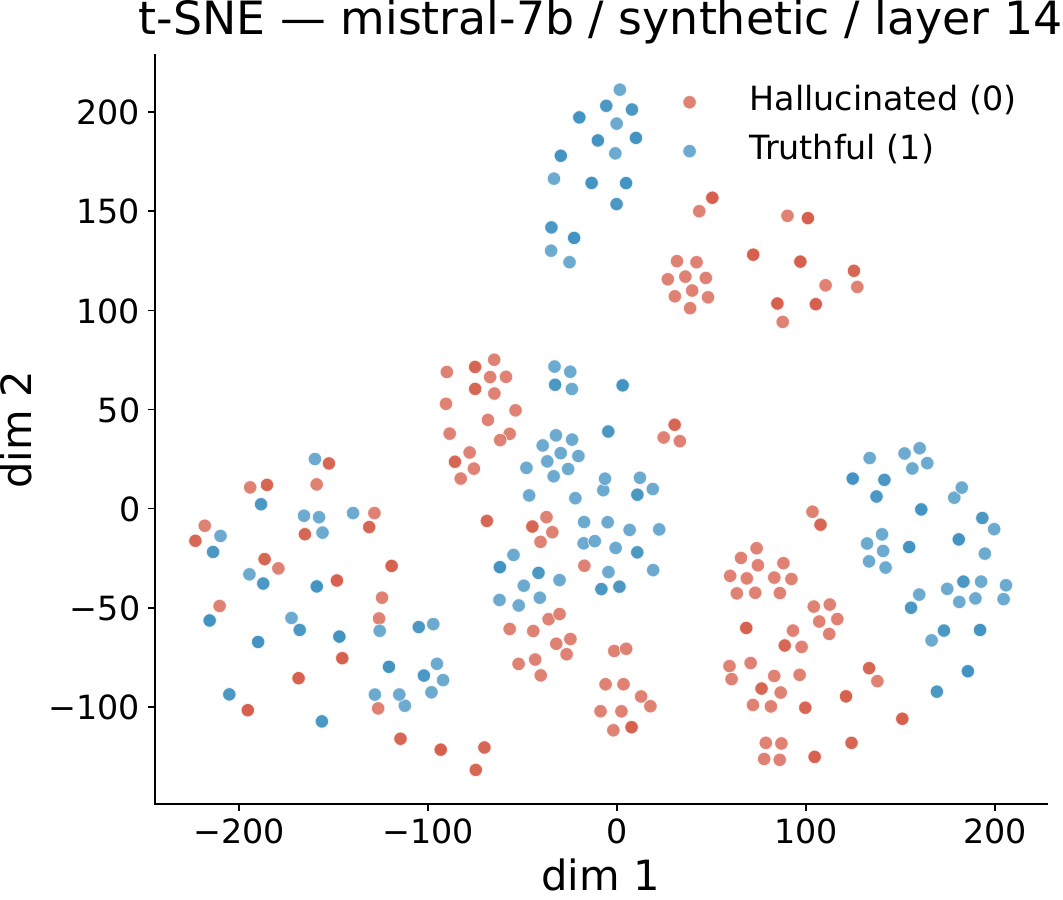}
\caption{PCA (left) and t-SNE (right) projections of Mistral-7B hidden states at the separation-ratio argmax block. Top to bottom: TruthfulQA (block~$0$), HaluEval-QA (block~$0$), FEVER (block~$0$), and synthetic (block~$14$). The three block-$0$ panels show overlapping clusters; the synthetic block-$14$ panel shows clearer separation, consistent with its high probe AUROC.}
  \label{fig:proj-mistral-all}
\end{figure}

\begin{figure}[htbp]
  \centering
  \includegraphics[width=0.48\columnwidth]{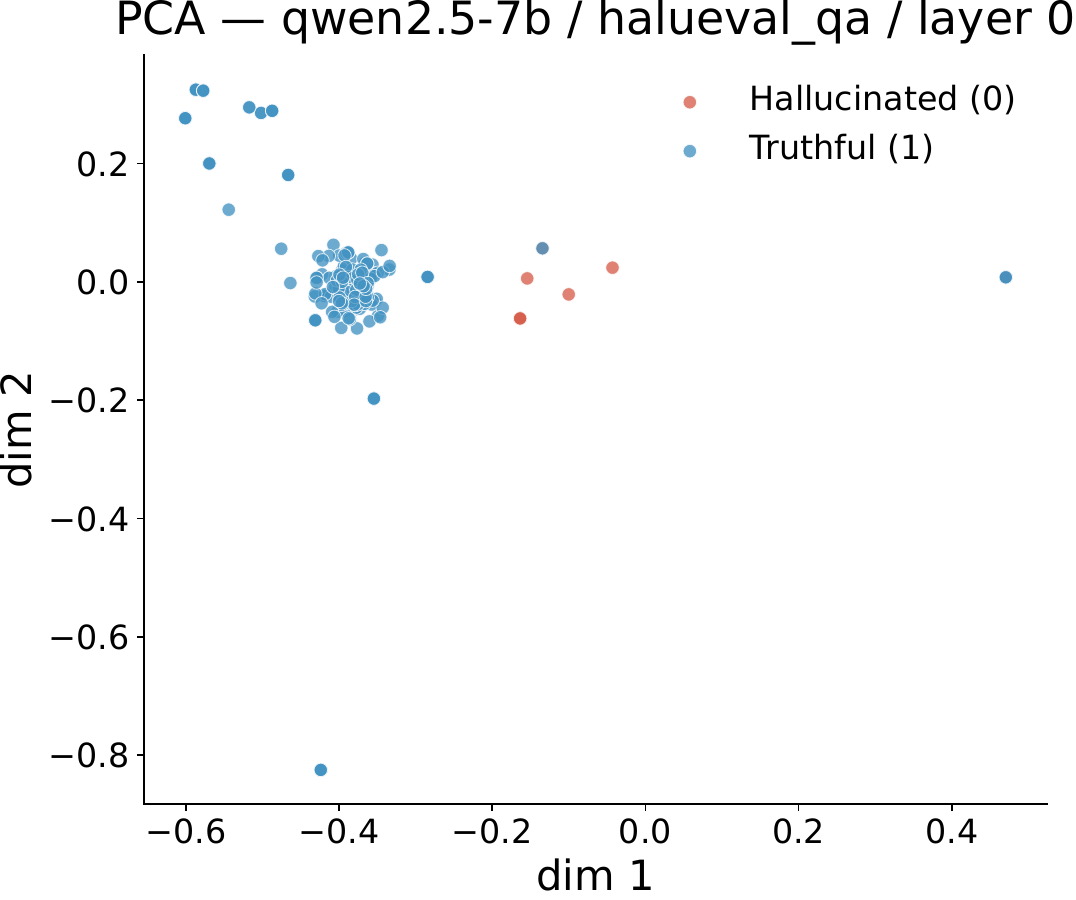}\hfill
  \includegraphics[width=0.48\columnwidth]{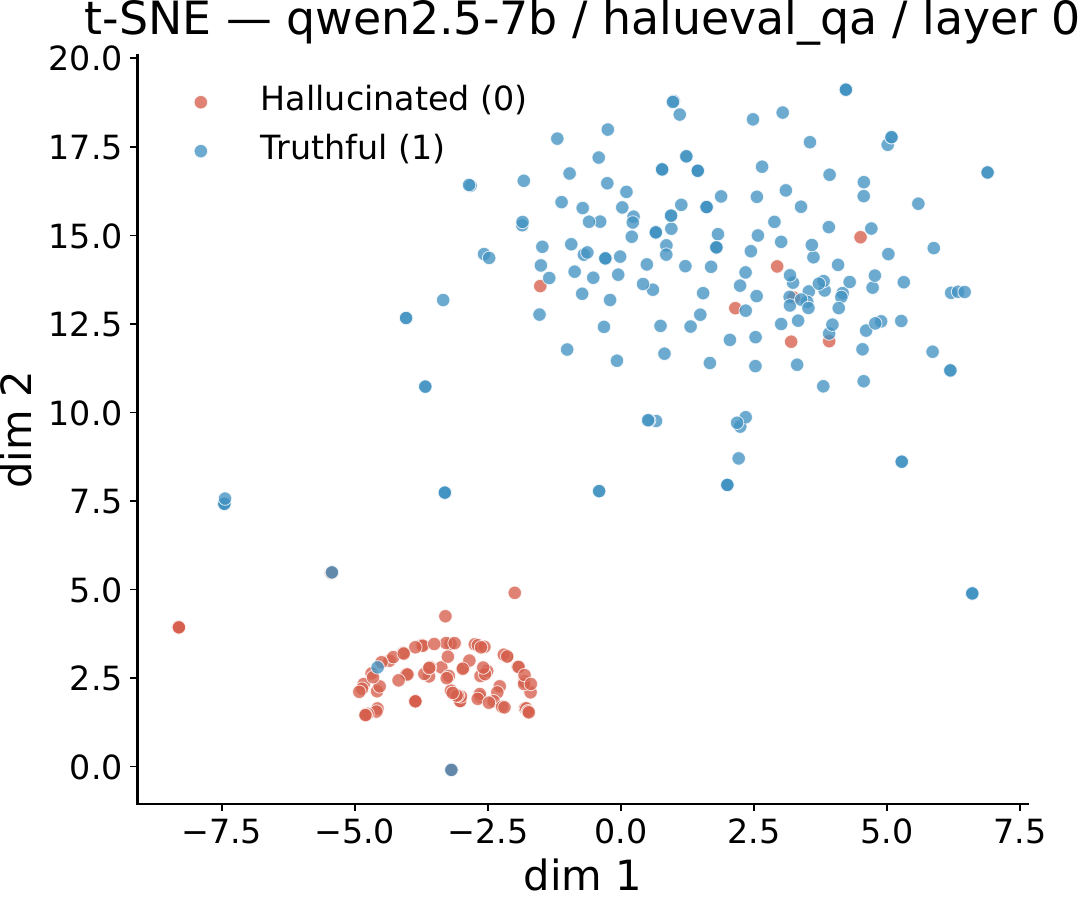}\\[2pt]
  \includegraphics[width=0.48\columnwidth]{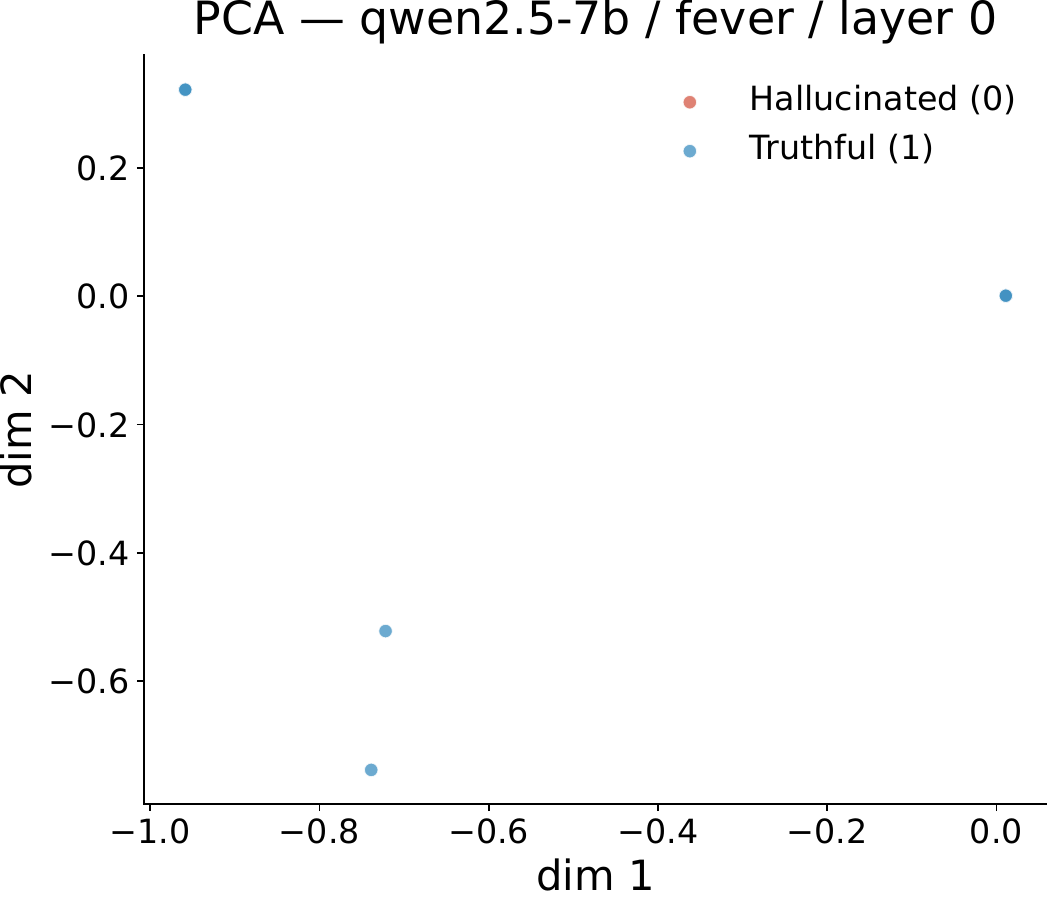}\hfill
  \includegraphics[width=0.48\columnwidth]{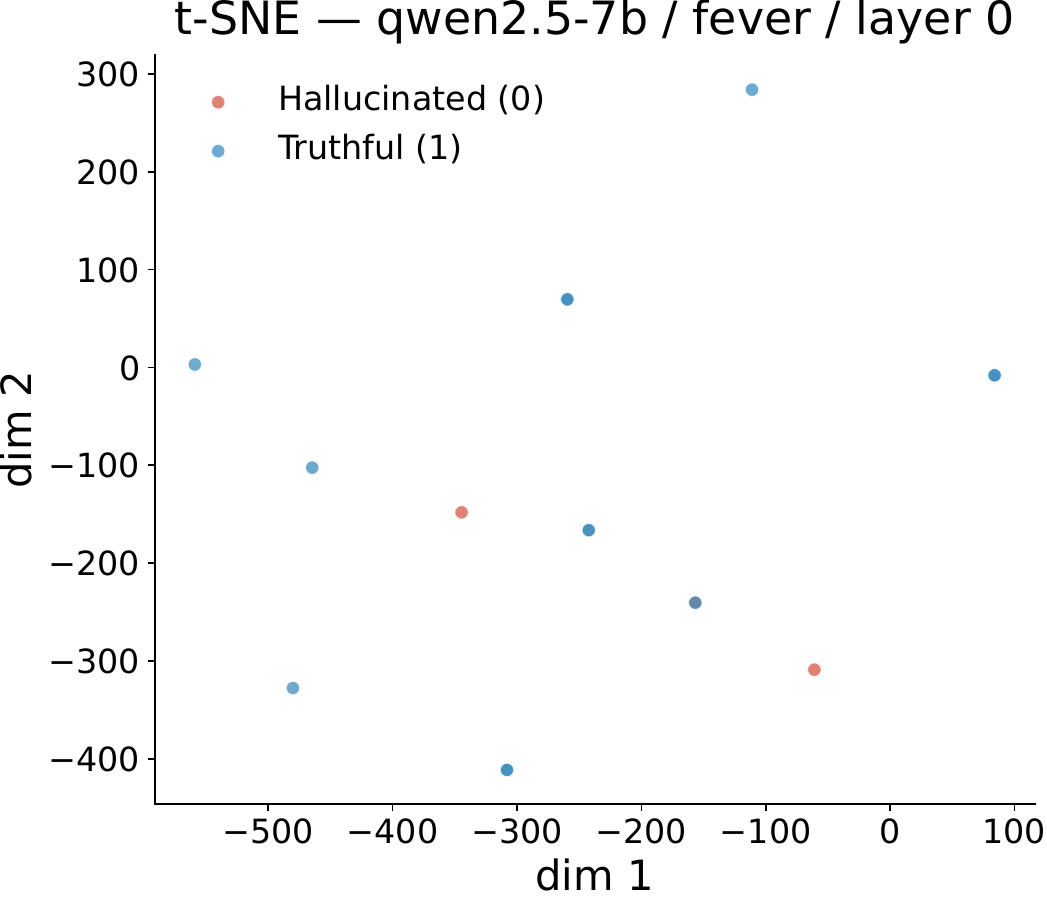}\\[2pt]
  \includegraphics[width=0.48\columnwidth]{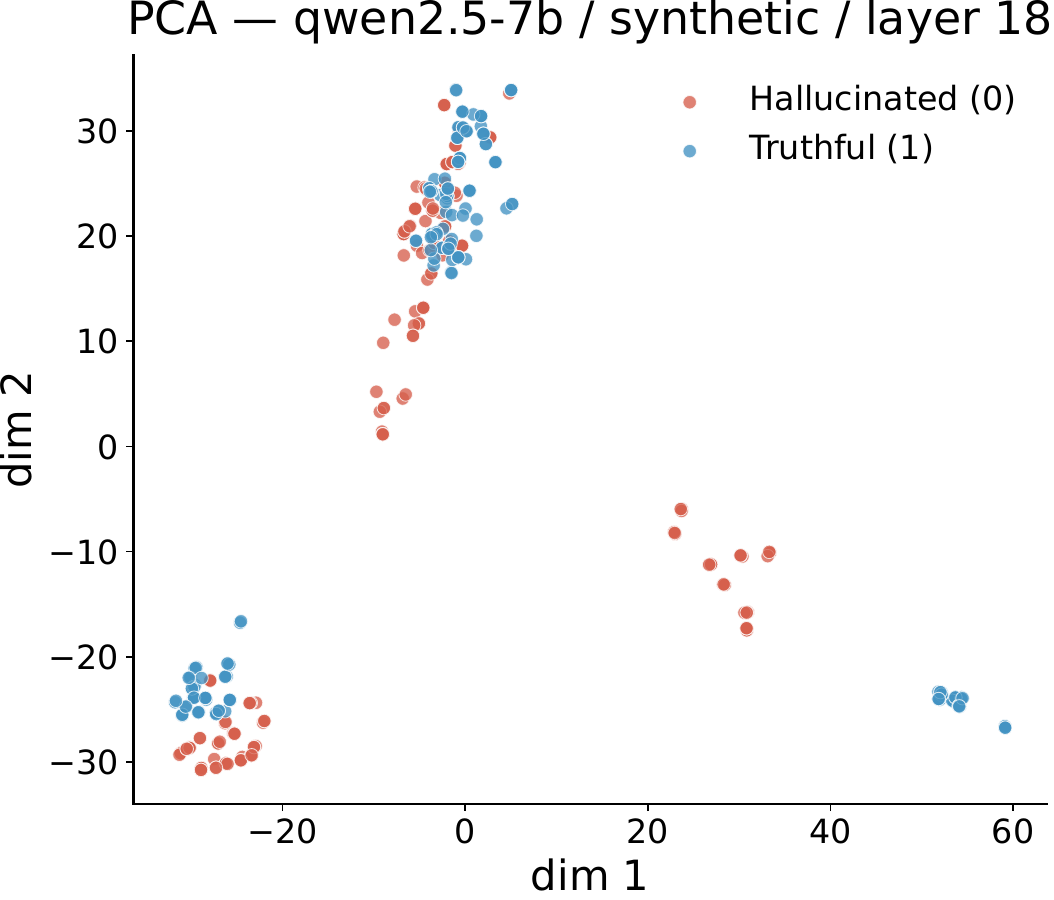}\hfill
  \includegraphics[width=0.48\columnwidth]{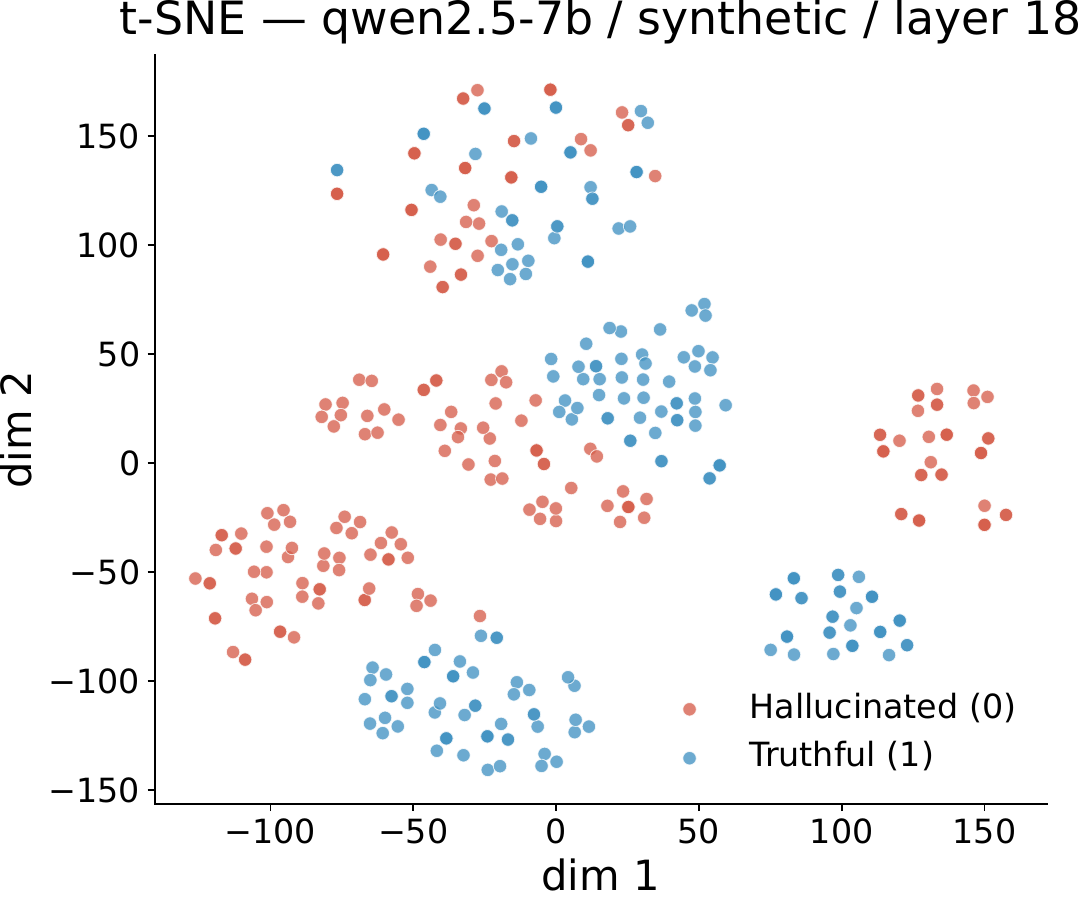}
\caption{PCA (left) and t-SNE (right) projections of Qwen2.5-7B hidden states at the separation-ratio argmax block. Top to bottom: HaluEval-QA (block~$0$), FEVER (block~$0$), and synthetic (block~$18$). The block-$18$ synthetic panel shows clear class clusters; the block-$0$ panels reflect lexical rather than truthfulness separation. Cf.\ Figures~\ref{fig:pca-qwen},~\ref{fig:tsne-qwen} for the TruthfulQA probing-peak projection.}
  \label{fig:proj-qwen-all}
\end{figure}

Figures~\ref{fig:proj-llama-all}--\ref{fig:proj-qwen-all} provide PCA and t-SNE projections of each model's hidden states at one selected block per configuration. The block shown is the one that maximizes the \emph{separation ratio} (centroid distance divided by mean within-class spread; see \S\ref{sec:method-metrics}), chosen independently of any trained probe. For Llama and Qwen on TruthfulQA, this block coincides with the probing peak (blocks~$14$ and~$19$ respectively). For several other configurations the separation ratio peaks at very early blocks, where surface-level token identity dominates the centroid geometry while truthful/hallucinated separation remains weak (visible as heavily overlapping point clouds). The projections at the probing peak (main text and Appendix~\ref{sec:appendix-pca}) remain the most informative visualization of the linearly decodable signal reported in Table~\ref{tab:headline}.

\clearpage

\section{Accuracy and $F_1$ at Best Block}
\label{sec:appendix-acc-f1}

Tables~\ref{tab:accuracy} and \ref{tab:f1} in the main text (\S\ref{sec:experiments-best-layer}) report accuracy and $F_1$ at the best block for every model, dataset, and method. Probes use a fixed threshold of $0.5$; the remaining methods use the Youden-$J$ optimal threshold (maximizing sensitivity plus specificity minus one). The tables confirm the same qualitative pattern as the AUROC results: probes achieve accuracy $\geq\!0.800$ and $F_1\!\geq\!0.807$ on all natural-language benchmarks, while the sampling-based methods stay near chance (attention entropy again being the exception on HaluEval-QA). On the synthetic benchmark, MLP probes reach $0.975$--$0.992$ accuracy and comparable $F_1$, consistent with the $\geq\!0.998$ AUROC plateau. Exact-match self-consistency $F_1$ values are unstable because stochastic completions rarely collide as exact strings at $K{=}5$, causing the method to assign identical scores to nearly all samples.


\section{Linear-Probe Layer-Wise AUROC Trajectories}
\label{sec:appendix-auroc-linear}

\begin{figure}[htbp]
  \centering
  \includegraphics[width=0.49\textwidth]{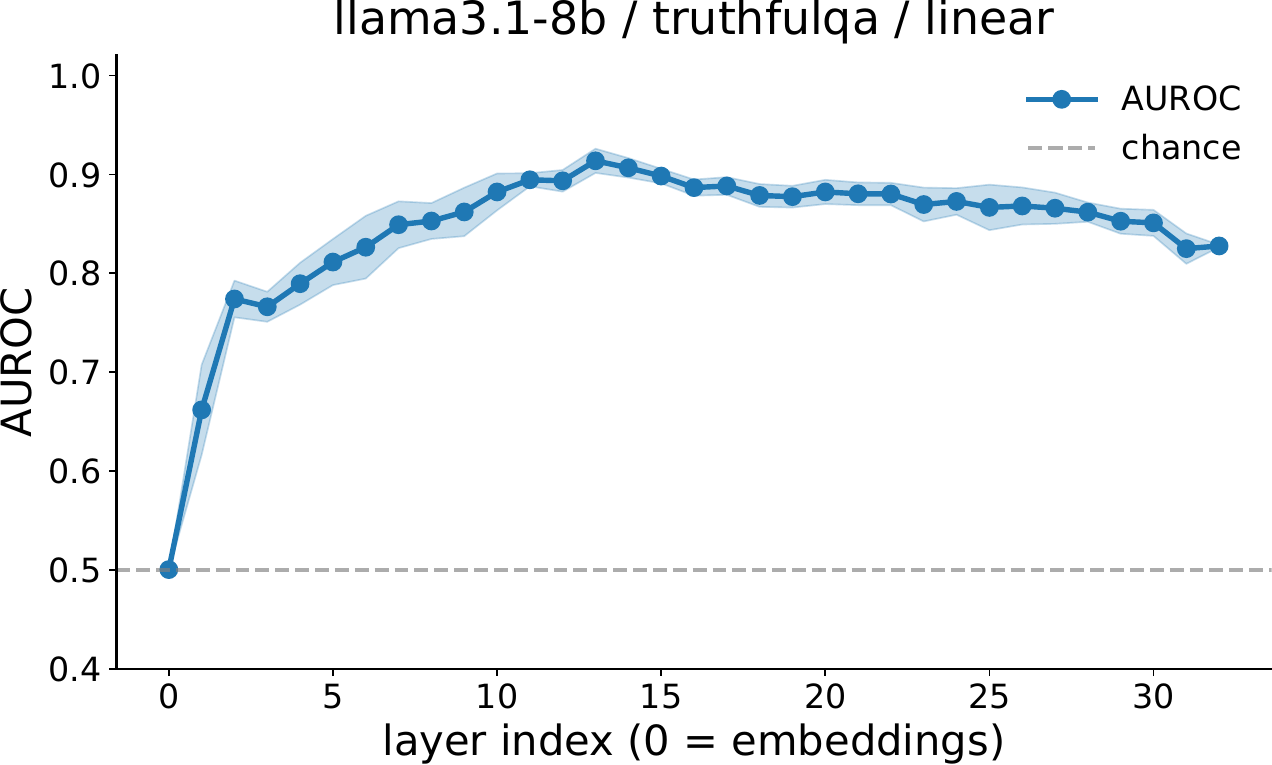}\hfill
  \includegraphics[width=0.49\textwidth]{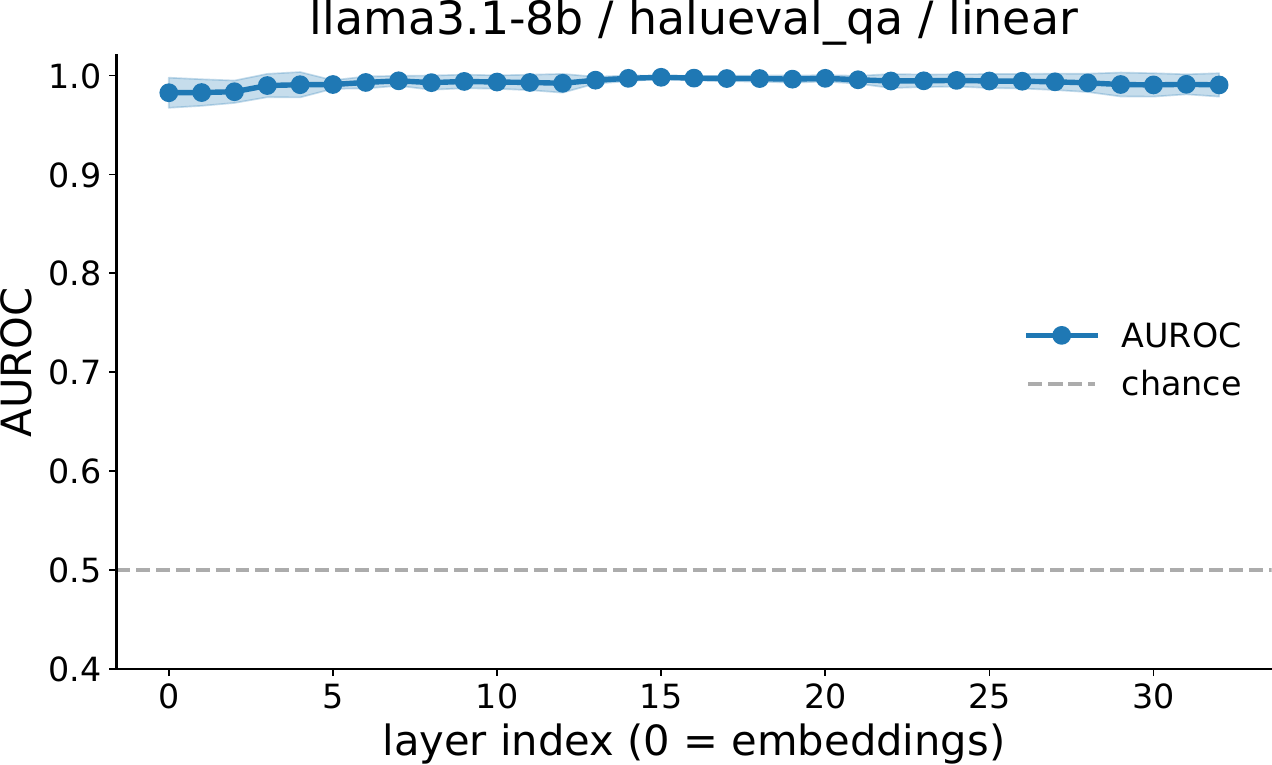}\\[2pt]
  \includegraphics[width=0.49\textwidth]{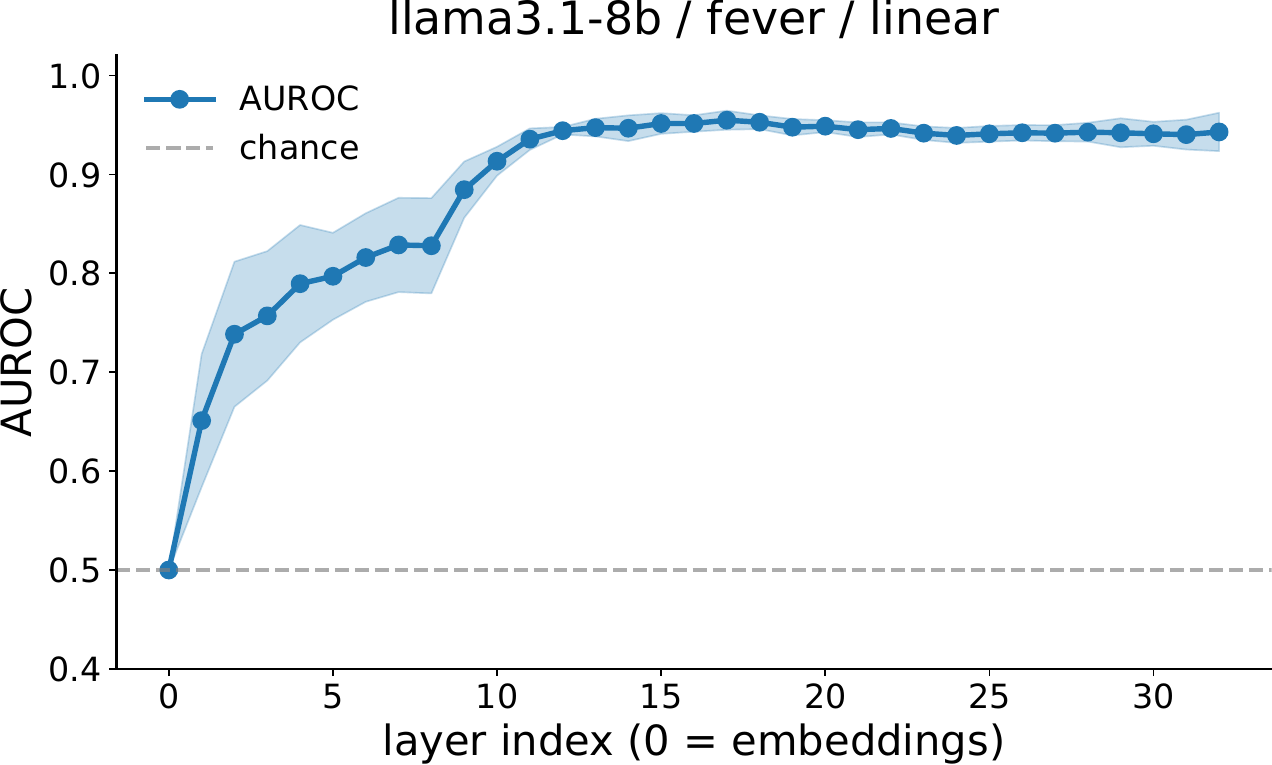}\hfill
  \includegraphics[width=0.49\textwidth]{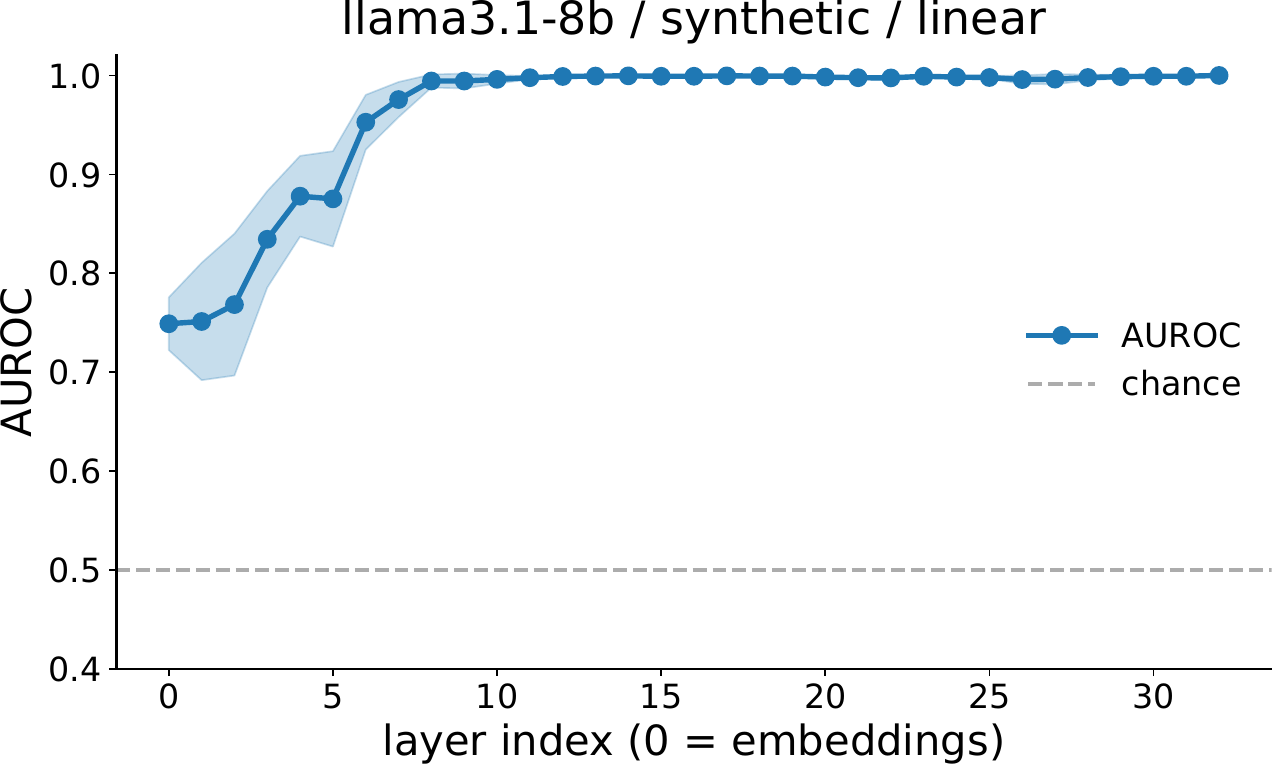}
\caption{Per-block \emph{linear}-probe AUROC for Llama-3.1-8B. \emph{Top-left}: TruthfulQA. \emph{Top-right}: HaluEval-QA. \emph{Bottom-left}: FEVER. \emph{Bottom-right}: synthetic. Shaded band: standard deviation over three seeds. Cf.\ Figure~\ref{fig:auroc-llama-all} for the MLP counterpart.}
  \label{fig:auroc-llama-linear}
\end{figure}

\begin{figure}[!ht]
  \centering
  \includegraphics[width=0.49\textwidth]{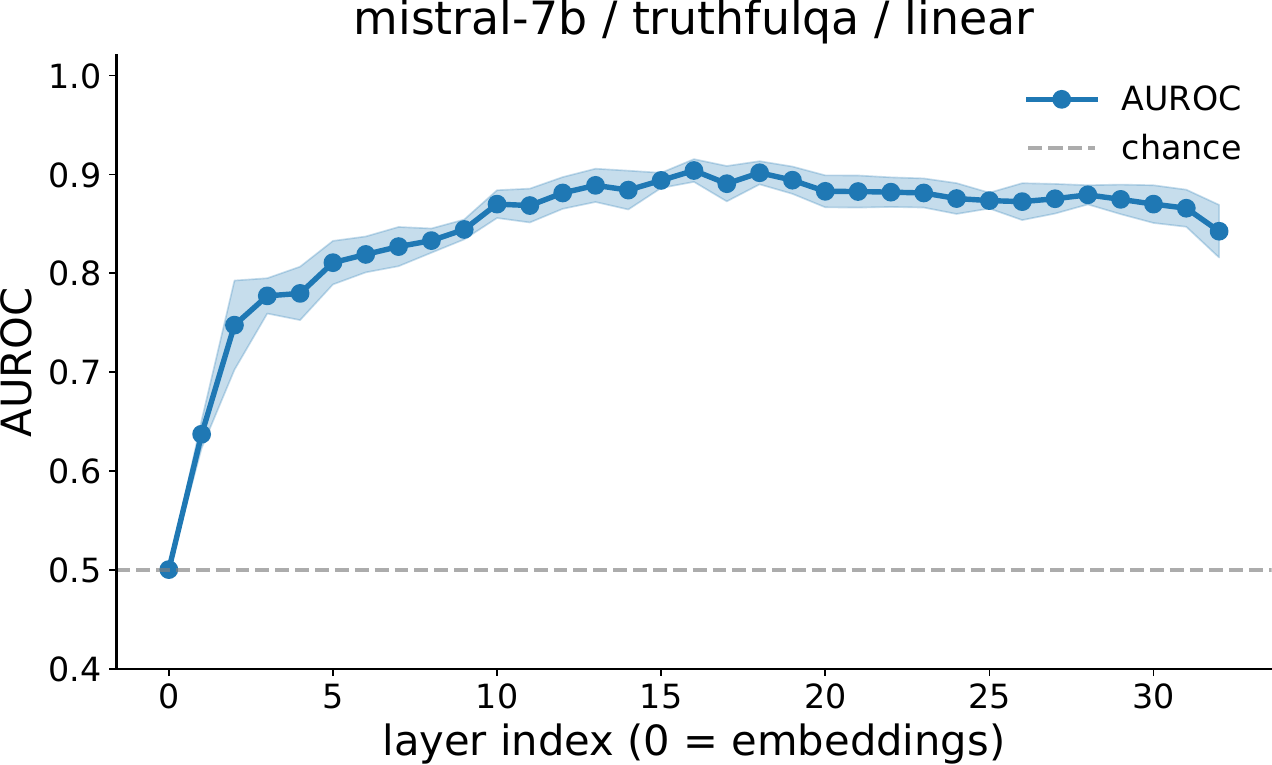}\hfill
  \includegraphics[width=0.49\textwidth]{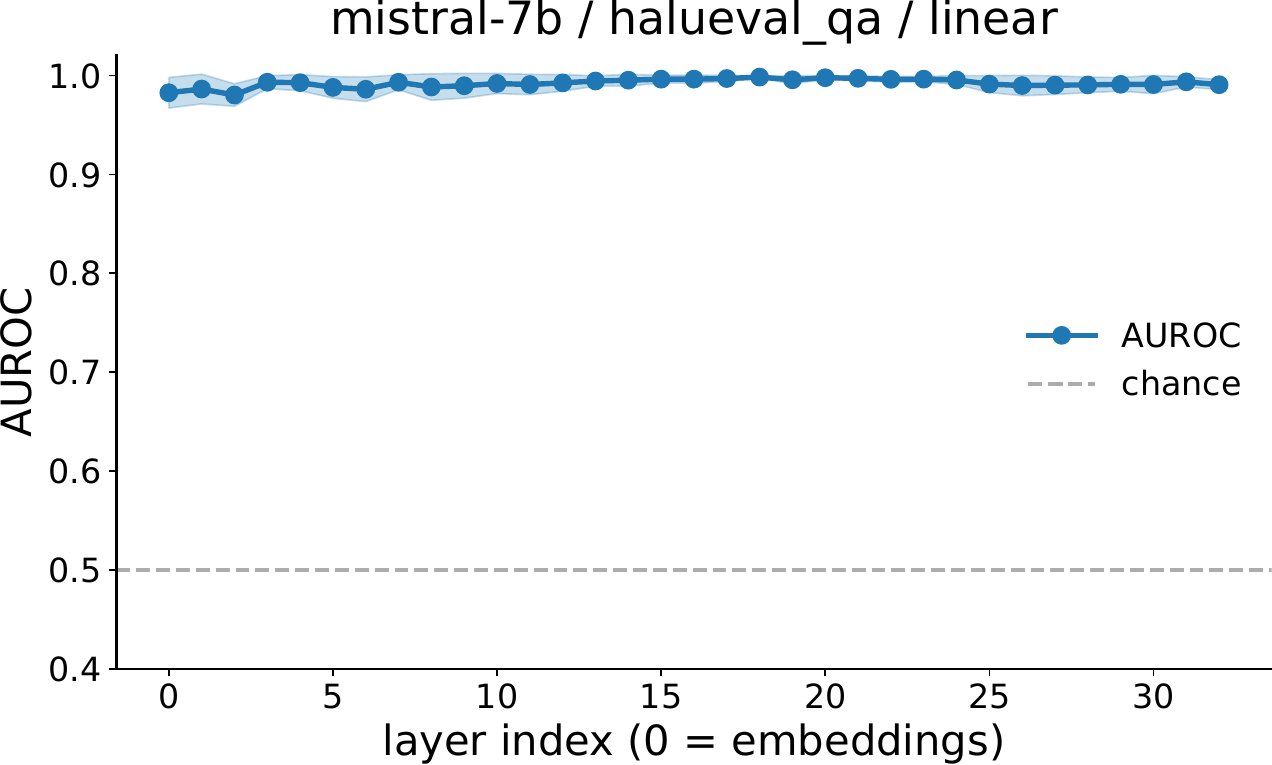}\\[2pt]
  \includegraphics[width=0.49\textwidth]{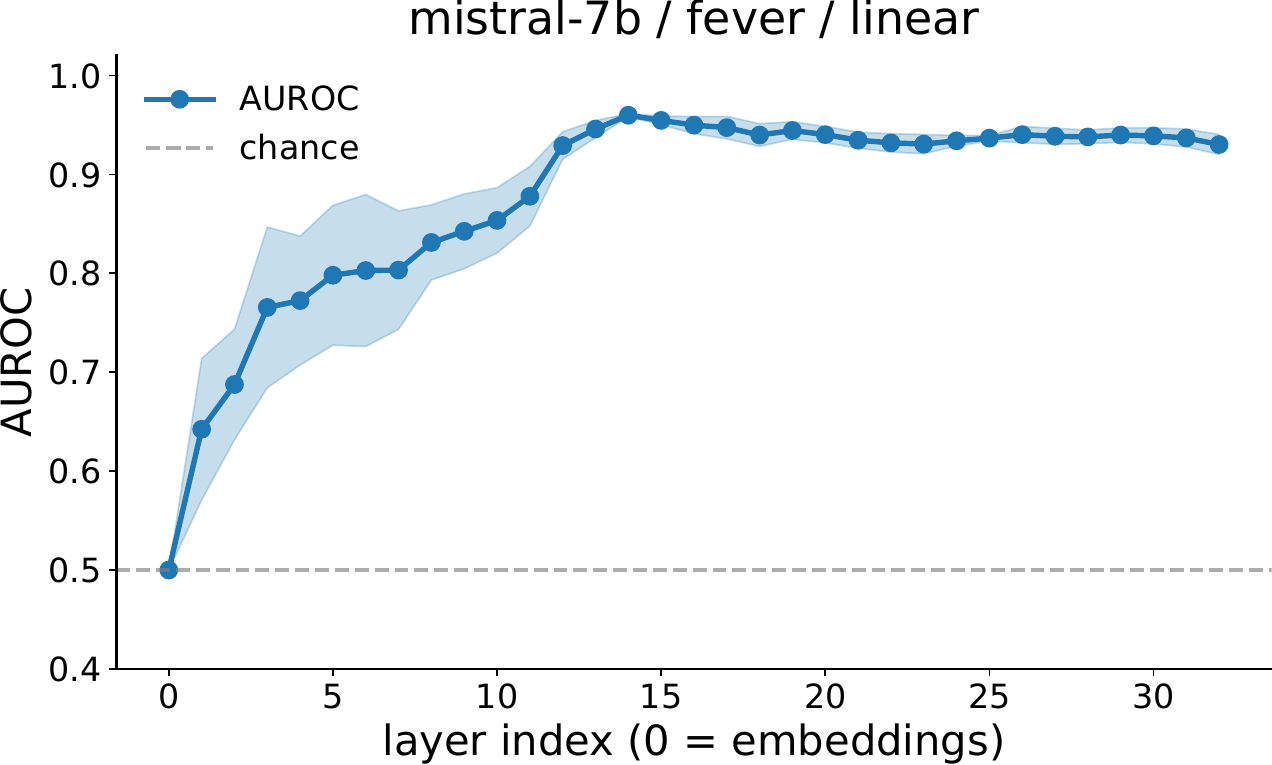}\hfill
  \includegraphics[width=0.49\textwidth]{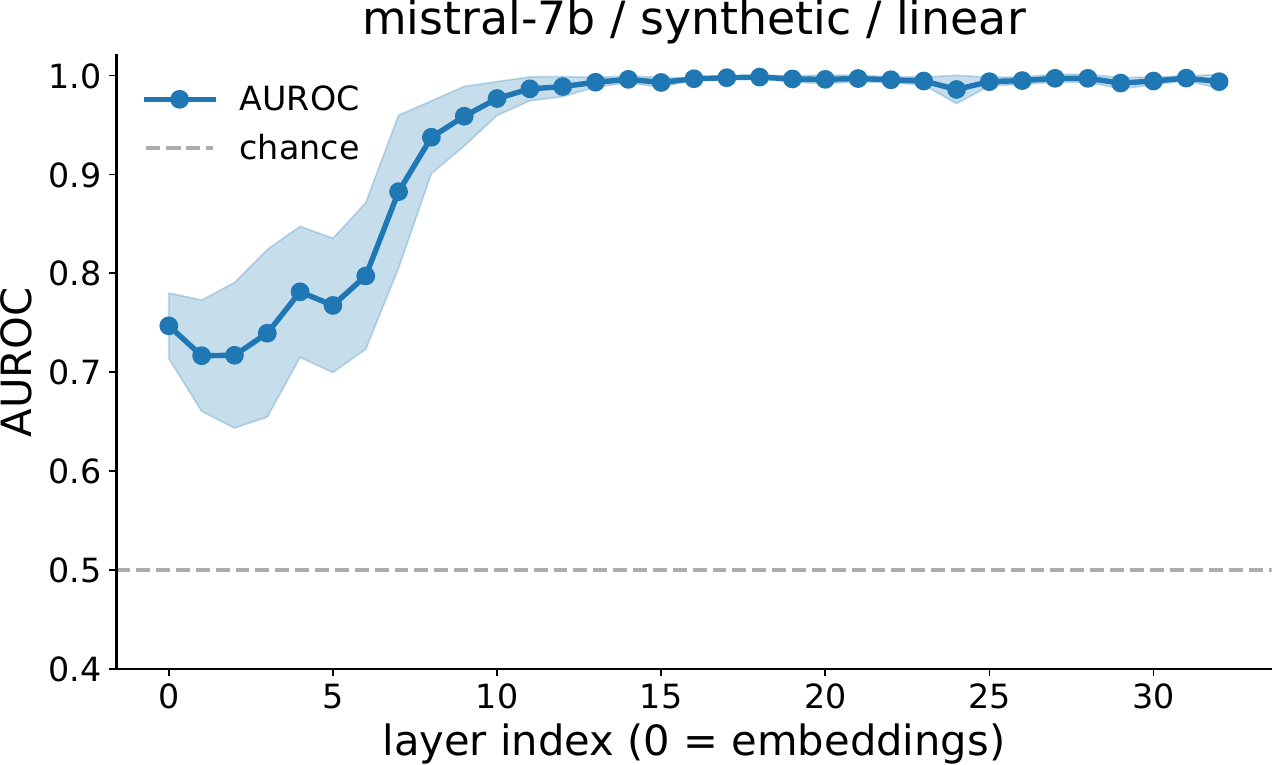}
  \vspace{-2mm}
  \caption{Per-block \emph{linear}-probe AUROC for Mistral-7B-Instruct-v0.3. \emph{Top-left}: TruthfulQA. \emph{Top-right}: HaluEval-QA. \emph{Bottom-left}: FEVER. \emph{Bottom-right}: synthetic. Shaded band: standard deviation over three seeds. Cf.\ Figure~\ref{fig:auroc-mistral-all} for the MLP counterpart.}
  \label{fig:auroc-mistral-linear}
\end{figure}

\begin{figure}[!ht]
  \centering
  \includegraphics[width=0.49\textwidth]{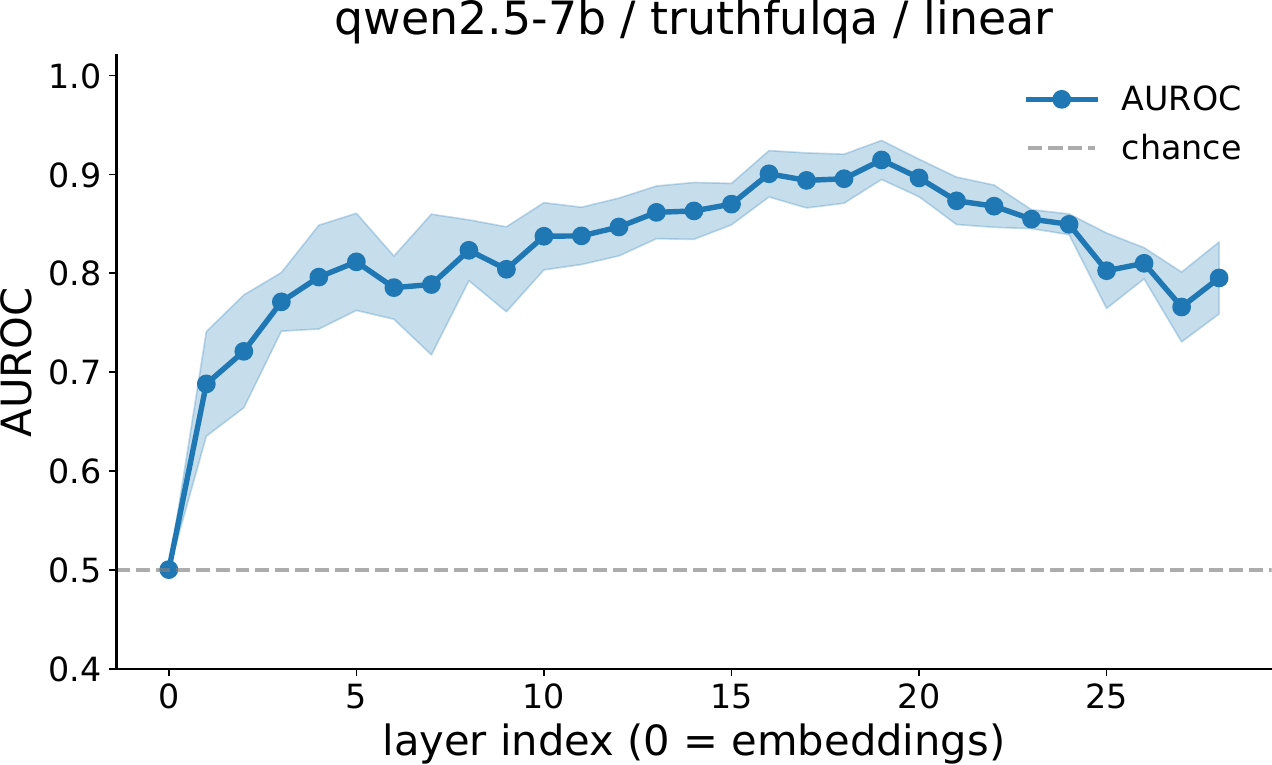}\hfill
  \includegraphics[width=0.49\textwidth]{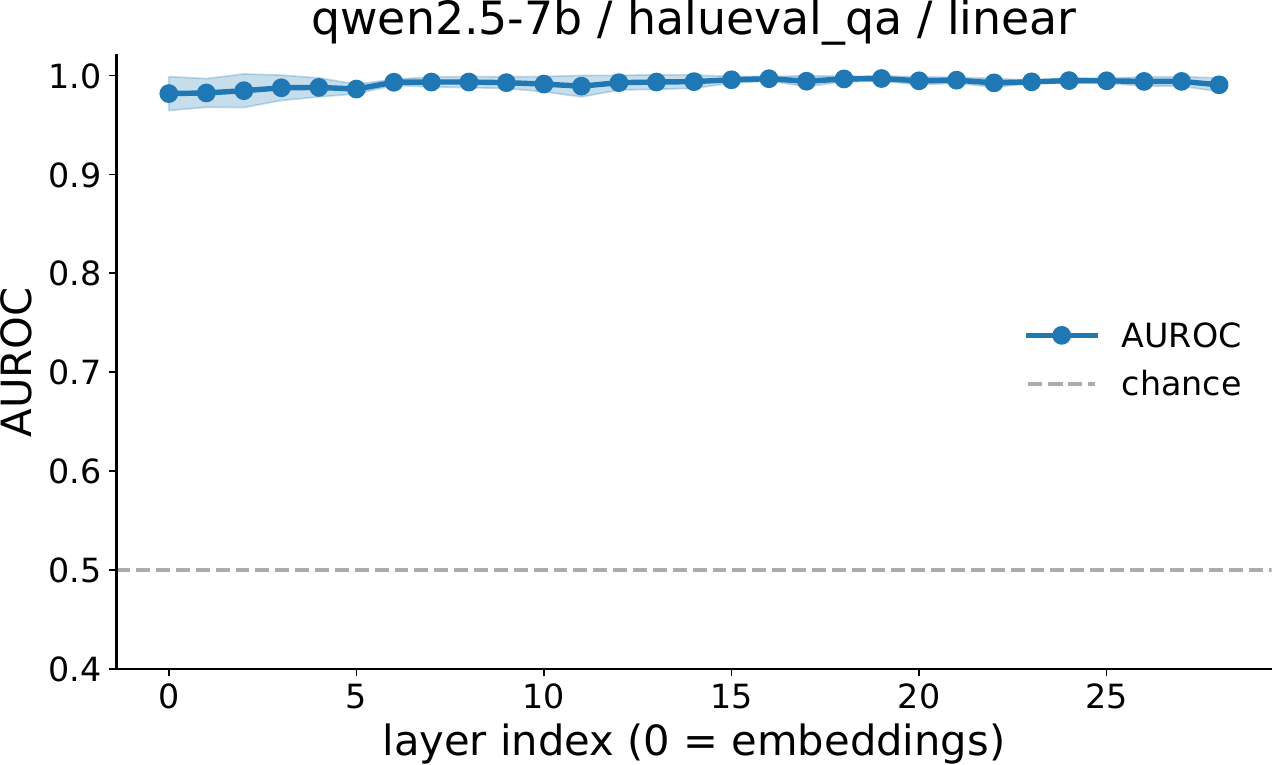}\\[2pt]
  \includegraphics[width=0.49\textwidth]{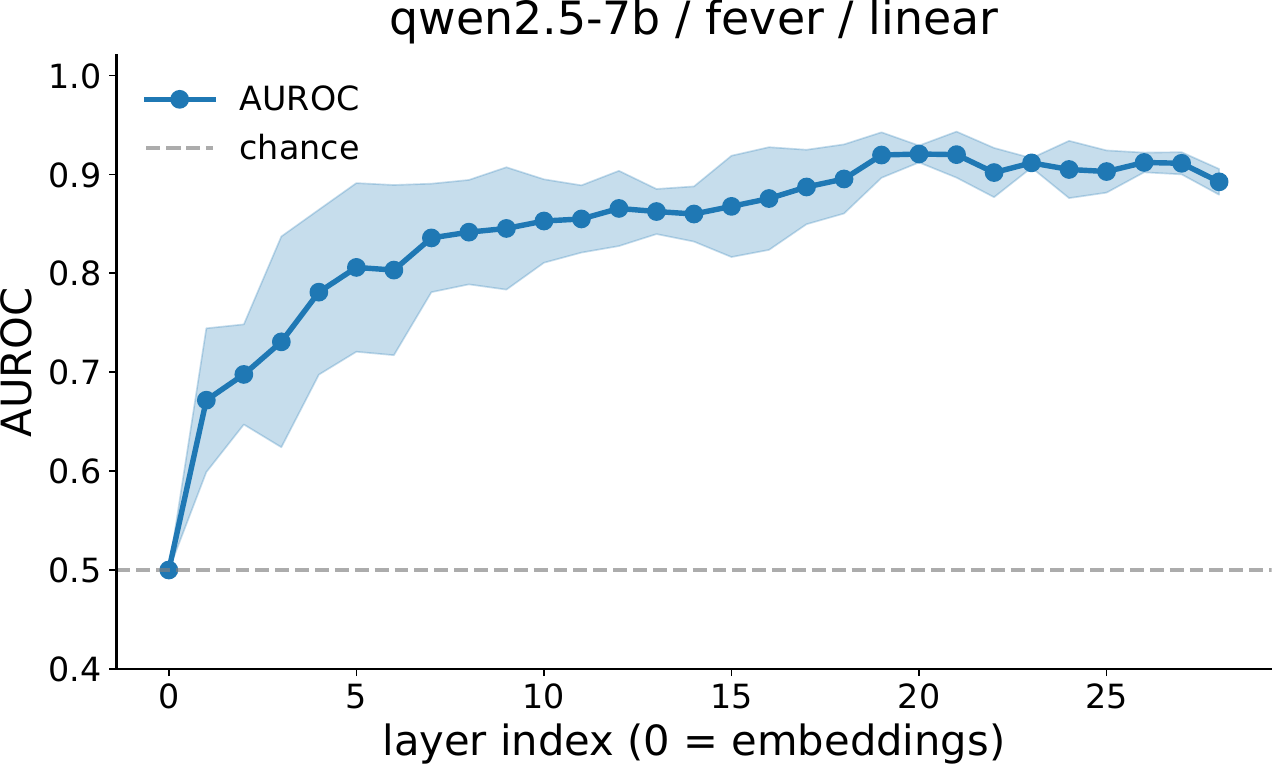}\hfill
  \includegraphics[width=0.49\textwidth]{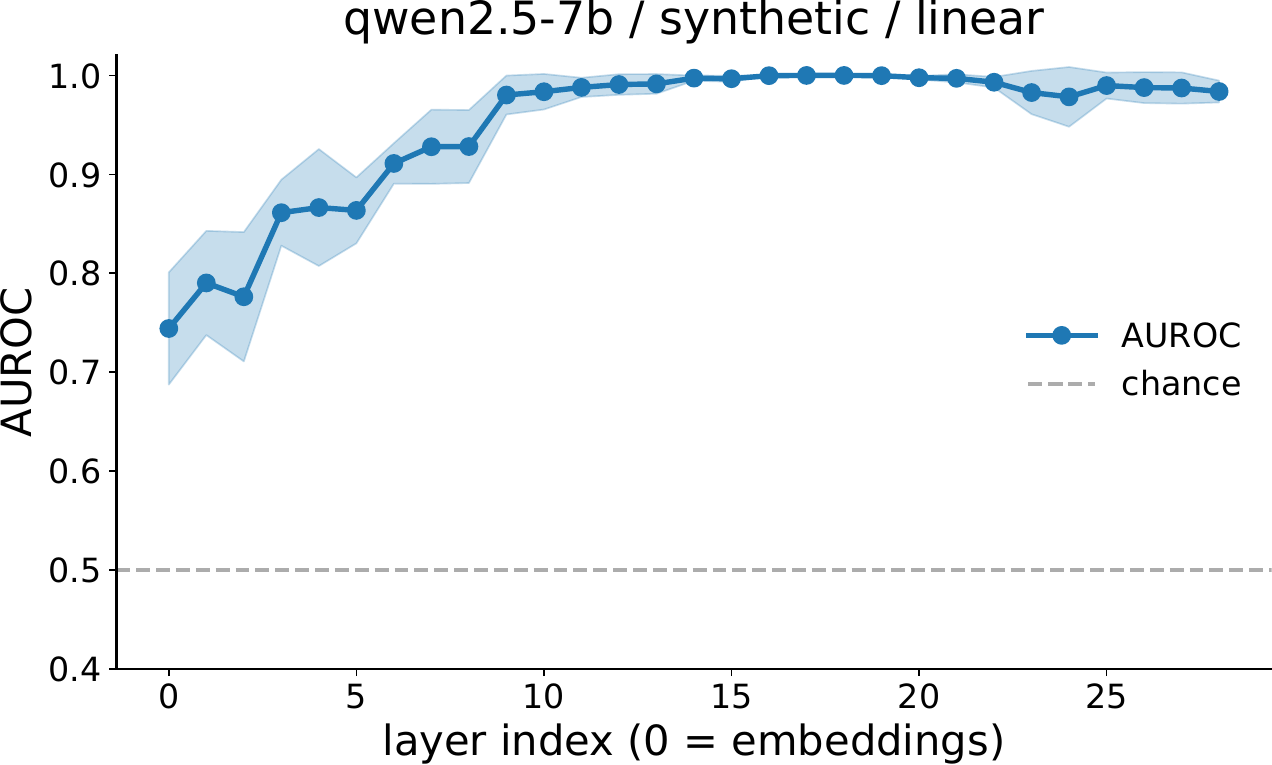}
  \vspace{-2mm}
  \caption{Per-block \emph{linear}-probe AUROC for Qwen2.5-7B-Instruct ($28$ transformer blocks). \emph{Top-left}: TruthfulQA. \emph{Top-right}: HaluEval-QA. \emph{Bottom-left}: FEVER. \emph{Bottom-right}: synthetic. Shaded band: standard deviation over three seeds. Cf.\ Figure~\ref{fig:auroc-qwen-all} for the MLP counterpart.}
  \label{fig:auroc-qwen-linear}
\end{figure}

Figures~\ref{fig:auroc-llama-linear}--\ref{fig:auroc-qwen-linear} report the per-block \textbf{linear}-probe AUROC curves for all $12$ configurations. As noted in \S\ref{sec:experiments-best-layer} (F2), these trajectories are visually near-identical to their MLP counterparts (Figures~\ref{fig:auroc-llama-all}--\ref{fig:auroc-qwen-all}), providing direct visual confirmation that the truthfulness signal is approximately linear across all three model families and all four datasets.

\vspace{3mm}

\section{Hyperparameters}
\label{sec:appendix-hyperparams}

Table~\ref{tab:hyper} consolidates all hyperparameters used in the experiments; these values are also included in the released code for full reproducibility.

\begin{table}[ht]
\centering
\small
\setlength{\tabcolsep}{4pt}
\caption{Hyperparameters used in all experiments. All values are provided in the released code for reproducibility.}
\label{tab:hyper}
\begin{tabular}{lr}
\toprule
\textbf{Hyperparameter} & \textbf{Value} \\
\midrule
Seed                             & $42$ \\
Items per dataset                & $400$ \\
Quantization                     & $4$-bit NF4 \\
Compute dtype                    & bfloat16 \\
Attention impl.                  & eager \\
Generation temperature           & $0.7$ \\
Top-$p$                          & $0.95$ \\
Max new tokens                   & $64$ \\
INSIDE samples $K$               & $10$ \\
INSIDE regularizer $\alpha$      & $10^{-3}$ \\
INSIDE block index               & $-1$ (final) \\
Self-consistency samples $K$     & $5$ \\
Self-consistency embedding model & all-MiniLM-L6-v2 \\
Probe optimizer                  & AdamW \\
Probe learning rate              & $10^{-3}$ \\
Probe weight decay               & $10^{-4}$ \\
Probe batch size                 & $128$ \\
Probe epochs                     & $30$ \\
MLP hidden dims                  & $[256, 64]$ \\
MLP dropout                      & $0.2$ \\
Train / val / test               & $70 / 10 / 20$ \\
Probe seeds                      & $3$ \\
Attention blocks captured        & first, middle, last \\
Extraction batch size            & $4$ \\
Max input length                 & $512$ tokens \\
Pooling                          & last answer token \\
t-SNE perplexity                 & $30$ \\
\bottomrule
\end{tabular}
\end{table}

\end{document}